\documentclass[letterpaper]{article} 
\usepackage{aaai2026}  
\usepackage{times}  
\usepackage{helvet}  
\usepackage{courier}  
\usepackage[hyphens]{url}  
\usepackage{graphicx} 
\usepackage{natbib}  
\usepackage{caption} 
\usepackage{algorithm}
\usepackage{algorithmic}

\usepackage{newfloat}
\usepackage{listings}
\DeclareCaptionStyle{ruled}{labelfont=normalfont,labelsep=colon,strut=off} 
\floatstyle{ruled}
\newfloat{listing}{tb}{lst}{}
\floatname{listing}{Listing}
\usepackage{algorithm}
\usepackage{algorithmic}
\usepackage{float}
\usepackage{placeins} 
\usepackage{caption}  

\usepackage{url}            
\usepackage{booktabs}       
\usepackage{amsfonts}       
\usepackage{nicefrac}       
\usepackage{microtype}      
\usepackage{xcolor}         
\usepackage{booktabs}
\usepackage{amsmath}
\usepackage{graphicx}
\usepackage{subcaption}
\usepackage{colortbl}      
\usepackage{xcolor}        
\usepackage{amssymb} 
\usepackage{adjustbox} 
\usepackage{makecell}
\usepackage{multirow}
\usepackage{comment}
\usepackage{booktabs}
\usepackage{tabularx}

\usepackage{newfloat}
\usepackage{listings}
\DeclareCaptionStyle{ruled}{labelfont=normalfont,labelsep=colon,strut=off} 
\floatstyle{ruled}
\newfloat{listing}{tb}{lst}{}
\floatname{listing}{Listing}
\title{Segment and Matte Anything in a Unified Model}
\author{
   Zezhong Fan\equalcontrib, Xiaohan Li\equalcontrib, Topojoy Biswas, Kaushiki Nag, Kannan Achan
}
\affiliations{
    Personalization Team, Walmart Global Tech\\

    Sunnyvale, CA, USA\\
    \{zezhong.fan, xiaohan.li, topojoy.biswas, kaushiki.nag, kannan.achan\}@walmart.com
}

\usepackage{bibentry}

\begin{document}

\maketitle

\begin{abstract}

Segment Anything (SAM) has recently pushed the boundaries of segmentation by demonstrating zero-shot generalization and flexible prompting after training on over one billion masks. Despite this, its mask prediction accuracy often falls short of the precision required in real-world applications. While several refinement modules have been proposed to boost SAM’s segmentation quality, achieving highly accurate object delineation within a single, unified framework remains an open challenge. Furthermore, interactive image matting—which aims to generate fine-grained alpha mattes guided by diverse user hints—has not yet been explored in the context of SAM. Insights from recent studies highlight strong correlations between segmentation and matting, suggesting the feasibility of a unified model capable of both tasks.

In this paper, we introduce Segment And Matte Anything (SAMA), a lightweight extension of SAM that delivers high-quality interactive image segmentation and matting with minimal extra parameters. Our Multi-View Localization Encoder (MVLE) captures detailed features from local views, while the Localization Adapter (Local-Adapter) refines mask outputs by recovering subtle boundary details. We also incorporate two prediction heads for each task into the architecture to generate segmentation and matting masks, simultaneously. Trained on a diverse dataset aggregated from publicly available sources, SAMA achieves state-of-the-art performance across multiple segmentation and matting benchmarks, showcasing its adaptability and effectiveness in a wide range of downstream tasks.
\end{abstract}

\section{Introduction}

Precise object segmentation lies at the heart of computer-vision applications, from photo editing and augmented reality to autonomous driving and medical analysis. Two complementary problems dominate this landscape. Semantic/instance segmentation assigns a class label to every pixel, while natural image matting predicts a continuous alpha matte that captures fine, semi-transparent boundaries such as hair or glass. Segment Anything (SAM)~\cite{kirillov2023segment} represents a milestone in segmentation research: trained on over one billion masks, it exhibits remarkable zero-shot generalization and supports diverse prompting modalities (points, boxes, text). Nevertheless, SAM’s raw masks often lack the tight boundaries, sub-pixel accuracy and detail preservation as discussed in~\cite{ke2023segment, liu2023promoting, liu2024segment, fan2024prompt}.

Recently, researchers improve SAM with dedicated refinement modules. For example, HQ-SAM~\cite{ke2023segment} extends the original SAM by introducing a learnable High-Quality (HQ) output token into the mask decoder to enhance the quality of mask prediction. Several other approaches, such as DIS-SAM~\cite{liu2023promoting}, SAMRefiner~\cite{lin2025samrefiner} and Pi-SAM~\cite{liu2024segment}, have attempted to address this limitation. However, these methods typically require additional post-processing models or rely on extra human interactions to refine the input prompts, thereby increasing model complexity and reducing practicality. We identify two primary challenges that hinder these SAM-based models from achieving accurate segmentation. First, interactive segmentation models like SAM struggle to capture detailed structures of target objects due to limited fine-grained perception. Second, it remains difficult to integrate high-resolution detail into the decoding process without compromising SAM's strong zero-shot generalization ability. Addressing these challenges is critical for advancing fine-grained, high-quality segmentation in complex scenes. 

Interactive matting~\cite{li2024matting, yao2024matte}, in contrast, focuses on estimating accurate alpha mattes under sparse user guidance (e.g., trimaps, scribbles, or clicks). While classical matting networks achieve remarkable boundary detail, they struggle with object-level reasoning and cannot generalize across categories without extensive retraining. Importantly, recent studies reveal strong structural correlations between segmentation and matting~\cite{wang2005iterative, zheng2024bilateral}: segmentation offers global object cues, whereas matting supplies local boundary precision. Leveraging these synergies within a unified model promises both practical simplicity and performance gains, yet remains largely unexplored.

To address these challenges, we present Segment And Matte Anything (SAMA), a lightweight extension of SAM that unifies high-accuracy segmentation and interactive matting in a single framework. It includes three key components. First, the Multi-View Localization Encoder (MVLE) enhances spatial precision by aggregating localized details from the multiple local views, capturing fine structures that the coarse global encoder may overlook.
Second, the Localization Adapter (Local-Adapter) refines mask predictions by injecting local features into the decoding process to integrate fine-grained local features into the decoding process. Furthermore, we extend our framework on both image segmentation and matting tasks with two prediction heads for each task. Specifically, we introduce a lightweight up-sampling module, enabling SAMA to produce both high-quality segmentation and matting masks simultaneously. This unified design allows seamless task transfer without architectural modification of the encoder and decoder. Importantly, all SAM parameters are kept frozen during training, and only the proposed modules are fine-tuned. This strategy ensures that our approach remains both data-efficient and computationally lightweight. Collectively, these modules add only 1.8\% to SAM’s parameter count and impose only marginal latency.

We utilize publicly open-sourced datasets including segmentation masks with high-quality alpha mattes and train SAMA end-to-end on both tasks. Comprehensive experiments on standard segmentation suites and matting benchmarks demonstrate that SAMA outperforms prior interactive segmentation and matting networks, while retaining SAM’s advantage of prompting flexibility.

Our contributions can be summarized as follows:
\begin{list}{$\square$}{\leftmargin=1em \itemindent=0em}
\item Unified framework: We propose the first SAM-based model that jointly performs interactive segmentation and matting with minimal overhead.

\item Architectural advances: We design a Multi-View Localization Encoder, Localization Adapter, and Prediction Heads to bridge object-level context and boundary-level detail.

\item State-of-the-art results: SAMA achieves new performance records across diverse segmentation and matting benchmarks without sacrificing inference speed or prompting versatility.
\end{list}

\section{Related Works}
\subsection{Interactive Segmentation and Matting}
Interactive segmentation allows users to steer the extraction of target regions through prompts such as bounding boxes~\cite{kirillov2023segment, ke2023segment, liu2024segment}, points~\cite{kirillov2023segment, ke2023segment, yao2025towards}, or natural-language descriptions~\cite{zou2023segment, nguyen2024calico, fan2025layoutagent}. Recent work embeds these prompts directly into the network to condition its predictions, with the Segment Anything Model (SAM)~\cite{kirillov2023segment}—pre-trained on over one billion masks—emerging as the de-facto benchmark. Some methods refine SAM to boost either accuracy, such as HQ-SAM~\cite{ke2023segment}, SAMRefiner~\cite{lin2025samrefiner} and DIS-SAM~\cite{liu2023promoting}. Meanwhile, there are also some papers to extend the functionality of SAM in semantic segmentation~\cite{zou2023segment, li2024segment}, iterative click-based refinement~\cite{liu2024segment}, cross-modal inputs~\cite{xiao2024segment} and segmentation with large vision–language models~\cite{nguyen2024calico}.


Variants of SAM have also been adapted for image matting. MAM~\cite{li2024matting} transforms SAM features into alpha mattes using a lightweight mask-to-matte (M2M) head, while MatAny~\cite{yao2024matte} generates a trimap with SAM and feeds it to VitMatte~\cite{yao2024vitmatte} for high-quality results. Despite their effectiveness, these approaches still depend on additional heavy models~\cite{yao2024vitmatte} or cascaded modules.

Recognizing the strong synergy between segmentation and matting~\cite{wang2005iterative, zheng2024bilateral}, we present a unified framework that augments SAM with a lightweight matting head, delivering precise segmentation masks and high-fidelity alpha mattes with minimal computational overhead.

\subsection{High-Quality Segmentation and Matting}

\textbf{Segmentation:}
Accurate delineation of fine-grained, complex objects underpins numerous sub-tasks, including dichotomous image segmentation (DIS)~\cite{qin2022highly,yu2024multi,zheng2024bilateral}, semantic segmentation~\cite{long2015fully,zhao2017pyramid,cheng2021per}, instance segmentation~\cite{he2017mask,dai2016instance}, and panoptic segmentation~\cite{kirillov2019panoptic,cheng2020panoptic}. Classic CNN-based approaches~\cite{he2017mask,qin2022highly,long2015fully,zhao2017pyramid,dai2016instance,yu2024multi} design sophisticated multi-scale modules to fuse low-level texture with high-level semantics via diverse receptive fields. Transformer-based models push this further with self-attention windows to capture local details while retaining global context~\cite{kirillov2023segment,zheng2024bilateral,cheng2021per,cheng2022masked}.

\textbf{Matting:}
Image matting methods fall into two streams.  
(1) \emph{Trimap-based matting} supplies a foreground/background/unknown trimap to networks, enabling deep models to produce precise alpha mattes~\cite{xu2017deep,lutz2018alphagan,tang2019learning,lu2019indices,hou2019context,li2020natural,yao2024vitmatte,park2022matteformer}.  
(2) \emph{Trimap-free matting} predicts the matte directly. Although more convenient, these methods still need auxiliary cues such as segmentation masks~\cite{yu2021mask,huynh2024maggie}, motion information~\cite{sengupta2020background}, or prompt signals~\cite{li2024matting,yao2024matte}.

As segmentation and matting are inherently complementary~\cite{wang2005iterative,zheng2024bilateral}, we propose a single unified network that includes lightweight prediction heads with a high-quality interactive segmentation backbone based on SAM, enabling both tasks to share features and mutually enhance each other, while incurring only minimal computational overhead.

\section{Methodology}
We propose a unified model Segment And Matte Anything (SAMA) to leverage SAM achieving both highly accurate segmentation and interactive image matting. 

\subsection{Preliminary}
SAM ~\cite{kirillov2023segment} consists of three components: An image encoder which is a VIT ~\cite{dosovitskiy2020image} backbone produces a $64\times64$ spatial feature map for the input image. A prompt encoder embeds user interactions (points, boxes, or masks) as positional tokens. Then combining image features with prompt tokens, a mask decoder, which is a two-layer transformer, is used to predict the final segmentation mask. To acquire the zero-shot transfer capability, SAM is trained on SA‑1B, a dataset containing 11 million images and over 1 billion automatically generated masks. 

Image matting is to estimate alpha matte $\alpha$ given only image $I$ as the input. Formally, given an image $I$, which can be viewed as a combination of foreground image $F$ and background image $B$ with coefficient $\alpha$,
\begin{equation}
    I = \alpha F + (1- \alpha) B
\end{equation}

\begin{figure*}[ht]
    \centering
        \includegraphics[width=1\textwidth]{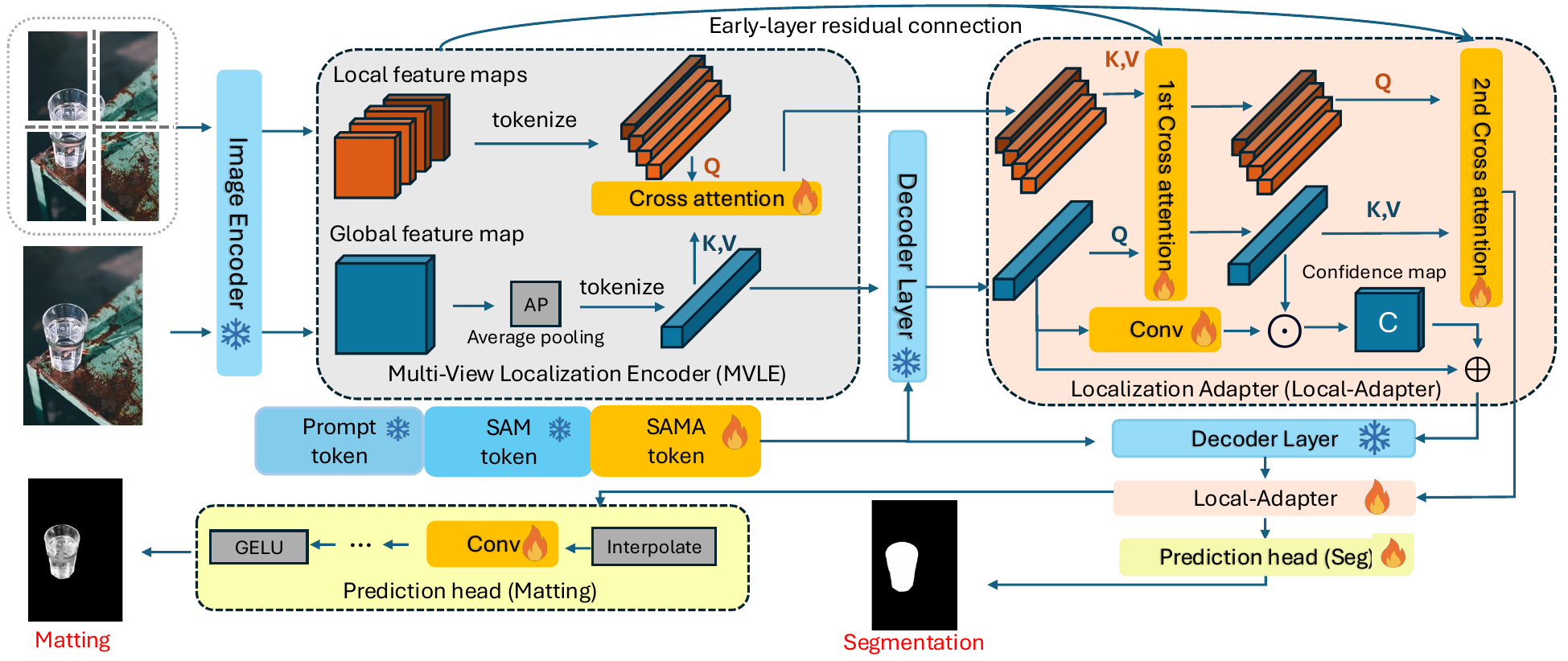}
        \caption{SAMA Overall Framework.}
        \label{fig:model}
\end{figure*}

\subsection{SAMA}
SAMA is a unified model that enables both highly accurate image segmentation and matting while preserving the zero-shot generalization capabilities of SAM. Unlike conventional single-image inputs, SAMA introduces a multi-view input strategy by treating the original image as a global view and incorporating additional local views to capture fine-grained object details. To effectively extract and fuse high-resolution features, we design three key components: Multi-view Localization Encoder (MVLE), Localization Adapter (Local-Adapter) and Matting Module, as illustrated in Figure~\ref{fig:model}. 

\subsubsection{Multi-view Localization Encoder (MVLE)}

In the SAM framework, an input image $I \in \mathbb{R}^{3 \times H \times W}$ is passed through an pre-trained image encoder $\mathcal{E}$ to produce a global feature map $F^{I} \in \mathbb{R}^{C \times \frac{H}{16} \times \frac{W}{16}}$. However, relying solely on this single global representation limits the model’s ability to capture fine-grained visual details. To address this limitation, we introduce the Multi-View Localization Encoder (MVLE), which enhances object localization through the use of high-resolution local views.

Inspired by human vision, we divide high-resolution inputs into distant-view global contexts and close-view local details, following MVANet~\cite{yu2024multi}, to promote comprehensive scene understanding. Specifically, we evenly crop the input image $I$ into four non-overlapping local patches $\{L_m\}_{m=1}^{4} \in \mathbb{R}^{3 \times h \times w}$, such that $(H, W) = (2h, 2w)$ as ~\cite{yu2024multi}. Each cropped patch is then up-sampled back to the original resolution and passed through the same encoder $\mathcal{E}$, yielding $m$ high-resolution local feature maps $F^{L_m} \in \mathbb{R}^{B \times C \times \frac{H}{16} \times \frac{W}{16}}$. The resulting local feature maps are stacked to form a $m$-layer (m = 4 here) local feature map $F^{L} \in \mathbb{R}^{B \times 4 \times C \times \frac{H}{16} \times \frac{W}{16}}$.

To effectively align local features with their global context, we apply a cross-attention mechanism between local features $F^{L_m}$ and the global feature $F^{I}$. First, we apply average pooling with multiple receptive fields (e.g., 4, 8, 16) to $F^{I}$ to obtain a multi-scale context representation $F^{I}_{\text{pool}}$. We then partition $F^{I}_{\text{pool}}$ into four spatial regions $\{I_m\}_{m=1}^{4}$, each corresponding to the position of a local patch. Within each region, we perform multi-head cross-attention, treating the local features as queries and the pooled global features as keys and values:

\begin{equation}
\scalebox{0.85}{$
\mathbf{Q}_m = F^{L_m} \mathbf{W}^{Q_m}, 
\mathbf{K}_m = F_{pool}^{I_m} \mathbf{W}^{K_m}, 
\mathbf{V}_m = F_{pool}^{I_m} \mathbf{W}^{V_m}$}
\end{equation}

\begin{equation}
F'^{P_m} = \text{Cross-Attn}(\mathbf{Q}_m, \mathbf{K}_m, \mathbf{V}_m),
\end{equation}

where $\mathbf{W}^{Q}_m, \mathbf{W}^{K}_m, \mathbf{W}^{V}_m$ are learnable projection matrices for the $m$-th patch. The output of the cross-attention layer in MVLE is the updated local features $F'^{P_m}$. These refined features are then passed to decoders to support precise mask prediction.

\subsubsection{Localization Adapter (Local-Adapter)}
Local-Adapter is a dedicated module designed to inject fine-grained visual information from high-resolution local features into the SAM decoder via a specialized local attention mechanism. Specifically, after processing the local features through one layer of the SAM decoder, we divert the output to Local-Adapter instead of directly passing it to the next decoder layer. 

SAM original decoder is a two-way Transformer. The input of the decoder layer is composed of two parts. The first part is the global feature map $F^{I_{\text{token}}}$ to serve as image tokens to the decoder layer. The other one is the concatenation of the prompt token, SAM token provided by SAM and our proposed SAMA token initialized in the model. Specifically, To enable the model to perform segmentation and matting simultaneously, we replace SAM’s original output token with learnable SAMA tokens—two tokens dedicated to segmentation and matting, respectively. These SAMA tokens are concatenated with the original SAM tokens and fed jointly into the decoder layers. The prompt token and SAM token are frozen while only SAMA token is trainable. Please note that the SAMA token is distinct in segmentation and matting tasks, i.e. two SAMA tokens are used separately when training and inference on both tasks. 

After each decoder layer, our proposed Local-Adapter is followed to enhance the local fine-grained features. There are three steps in Local-Adapter:

1) First cross-attention layer. We use the output of the cross-attention layer in MVLE as the input in this layer. To further enhance boundary sensitivity, we extract early-layer features from the image encoder as mentioned in~\cite{ke2023segment}, denoted as $F_{\text{early}}$, and then they are fused with local features via a residual connection. The fused tokens are served as the keys and values in this layer, while the decoder outputs $F_{\text{out}}$ provide the queries, allowing local and global information to be integrated seamlessly.

\begin{align}
    \mathbf{Q_A} &= F_{\text{out}} \mathbf{W}^{A}, \\
\mathbf{K_A}_m &= (F'^{P_m} + F_{\text{early}})\mathbf{W}^{K_A}, \\
\mathbf{V_A}_m &= (F'^{P_m} + F_{\text{early}}) \mathbf{W}^{V_A}, \\
F''^{P_m} &= \text{Cross-Attn}(\mathbf{Q_A}, \mathbf{K_A}_m, \mathbf{V_A}_m),
\end{align}

2) Second cross-attention layer. Inspired by GLIP\cite{li2022grounded} and GroundingDINO~\cite{liu2024grounding}, we introduce a feature fusion with global-local and local-global cross-attention modules by swapping keys and values in the previous layer, enabling bidirectional interaction between global context and local details. This allows the Local-Adapter to become both globally and locally aware within the decoders. Similar to the first layer, we swap roles: the keys and values produced in the previous layer become the queries for this layer, while the previous queries now serve as keys and values.

3) Generation of output features. There are two output features from Local-Adapter. The first is the tokens from the output of the second cross-attention layer, which will be used as the input of the second Local-Adapter in the SAMA model. The second feature is the updated global feature map $F'_{\text{out}}$, which is used as the input of the second decoder layer. It is the combination of a confidence map $C$ and the global feature from the output of the decoder layer $F_{\text{out}}$. The confidence map is to maintain SAM’s strong zero-shot generalization and mitigate the risks of overfitting or catastrophic forgetting. A confidence map $C$ is generated using a $1 \times 1$ convolution followed on the $F_{\text{out}}$ and activated by a sigmoid function, and multiple the output of the first cross attention layer $F''^{P_m}$. The second output feature is calculated from
\begin{equation}
C = \sigma\left(\text{Conv}(F_{\text{out}})\right)\odot F''^{P_m},
\end{equation}
\begin{equation}
F'_{\text{out}} = F_{\text{out}} + C ,
\end{equation}

where $\odot$ denotes element-wise product. This formulation allows the model to adaptively blend detailed information from the local features with the original decoder output, thereby achieving a better balance between precision and generalization. After the Local-Adapter, there is another set of decoder layer and Local-Adapter to increase the depth of the model and improve the performance.




\subsection{Prediction Heads}

Here we introduce how SAMA predicts final segmentation and alpha matting masks. First, following~\cite{ke2023segment}, we introduce two trainable output tokens-a segmentation token and a matting token (all shown as SAMA token in Figure~\ref{fig:model})—designed to generate high-quality outputs for their respective tasks. These tokens are processed by the SAMA decoder, which provides semantic features as global priors for final prediction heads. To enhance fine-grained predictions, we employ two lightweight task-specific prediction heads for segmentation and matting. Each head integrates an interpolation operation for upsampling with convolutional layers that collaboratively reconstruct and enhance details. The convolutional layers, along with batch normalization and GeLU activation function, are to generate fine-grained feature maps from the output of Local-Adapter. This design enables SAMA to produce high-resolution segmentation and matting masks simultaneously, achieving both semantic coherence and boundary-level precision. 

\subsection{Training of SAMA}
\textbf{Training Data Construction}
To enable efficient and effective training of SAMA, we opt for high-quality segmentation datasets with exceptionally accurate mask annotations- DIS-5K~\cite{qin2022highly} and ThinObject-5K~\cite{liew2021deep}, instead of relying on the large-scale but noisier SAM-1B dataset. For training the matting task, we utilize a combination of Adobe Image Matting (AIM)~\cite{xu2017deep} and AIM-500 ~\cite{li2021deep}, which together provide diverse and representative foreground objects across a range of natural scenes.

\textbf{SAMA Training}
During training, we freeze all parameters of the pre-trained SAM backbone and update only the our proposed modules. When optimizing for the segmentation task, the matting prediction head remains frozen, and vice versa during matting training. Additional implementation details are provided in the supplementary material.

\textbf{Training Loss} 
We train SAMA model end-to-end using a multi-task loss to learn segmentation and matting tasks concurrently: 

$$
\mathcal{L} = \mathcal{L}_{\text{seg}} + \mathcal{L}_{\text{matting}}
$$

For segmentation training, we employ a composite loss function that integrates pixel, region, and boundary-aware supervision:

$$
\mathcal{L}_{\text{seg}} = \mathcal{L}_{\text{BCE}} + \mathcal{L}_{\text{IoU}} + \mathcal{L}_{\text{SSIM}}
$$

where  binary cross-entropy (BCE) loss provides pixel-level supervision to guide the generation of binary masks, intersection over union (IoU) loss introduces region-level constraints to enhance the overall segmentation quality, and structural similarity index measure (SSIM) loss encourages structural similarity, particularly improving mask accuracy near object boundaries.

For matting, we adopt a more fine-grained objective that accounts for subtle variations in transparency and edge quality:

\begin{equation}
\mathcal{L}_{\text{matting}} = \mathcal{L}_{l_1} + \mathcal{L}_{\text{SSIM}} + \mathcal{L}_{\text{Grad}} + \mathcal{L}_{\text{Laplacian}}
\end{equation} 

where $\ell_1$ loss ensures global consistency, SSIM preserves structural similarity, gradient loss~\cite{dai2022boosting} improves edge sharpness, and Laplacian loss ~\cite{hou2019context} captures high-frequency details.

\section{Experiments}
In this section, we present experiment settings and comparisons for both segmentation and matting across multiple benchmarks. 
We further evaluate SAMA’s performance under point-based interactive segmentation and zero-shot semantic image matting settings. To assess the contribution of the different modules, we conduct ablation studies. 

\subsection{Comparison Study on Segmentation task}
We conduct experiments on an extremely fine-grained segmentation datasets: DIS-5K \cite{qin2022highly} including a validation set with 470 images (DIS-VD), four subsets DIS-TE1 $\sim$ TE4 (500 images in each set) with increasing shape complexities, and DIS-TE (All 2000 testing images). We compare SAMA with SAM~\cite{kirillov2023segment}, HQ-SAM~\cite{ke2023segment}, Pi-SAM ~\cite{liu2024segment}, DIS-SAM~\cite{liu2023promoting} IS-Net~\cite{qin2022highly}, UDUN ~\cite{pei2023unite} and BiRefNet~\cite{zheng2024bilateral} tailored for the task of dichotomous image segmentation. For SAM, HQ-SAM, Pi-SAM, DIS-SAM and our SAMA, bounding boxes are provided as prompts for images, while ISNet, UDUN and BiRefNet only take images as the input. For a thorough performance evaluation of segmentation, we report maximum F-measure ($F^{\text{max}}_{\beta}$~\cite{achanta2009frequency}), weighted F-measure ($F^{w}_{\beta}$), mean absolute error ($MAE$), S-measure ($S_{\alpha}$~\cite{fan2017structure}), and average enhanced alignment measure ($E^{m}_{\phi}$~\cite{fan2018enhanced}) as evaluation metrics.

As shown in Table~\ref{dis_res}, our proposed SAMA consistently outperforms other models that are built based on the original SAM architecture, highlighting the effectiveness of our newly introduced modules and fine-tuning strategy. Furthermore, when compared with models specifically designed and extensively trained on the DIS dataset, our SAMA achieves competitive performance across multiple evaluation metrics. Even though the dataset is more complex on DIS TE3 and TE4, our SAMA still achieves close to SOTA performance.  It should be noted that these baseline models are often trained with significantly more epochs, allowing them to better fit the inherent distributional characteristics of the dataset. However, due to their reliance on fully automatic segmentation pipelines, such models lack the flexibility to support interactive segmentation, which remains a key strength of our approach.

\begin{table*}[t]
\centering
\scriptsize
\setlength{\tabcolsep}{2.5pt}
\renewcommand{\arraystretch}{1}

\begin{tabular}{c|ccccc|ccccc|ccccc}
\toprule
\multicolumn{1}{c|}{} & \multicolumn{5}{c|}{\textbf{DIS-VD}} & \multicolumn{5}{c|}{\textbf{DIS-TE1}} & \multicolumn{5}{c}{\textbf{DIS-TE2}} \\
Method & $F_\beta^{\text{max}}$ & $F_\beta^{w}$ & $M$ & $S_\alpha$ & $E_\phi^{m}$ &
$F_\beta^{\text{max}}$ & $F_\beta^{w}$ & $M$ & $S_\alpha$ & $E_\phi^{m}$ &
$F_\beta^{\text{max}}$ & $F_\beta^{w}$ & $M$ & $S_\alpha$ & $E_\phi^{m}$ \\
\midrule
IS-Net     & 0.791 & 0.717 & 0.074 & 0.813 & 0.856 & 0.740 & 0.662 & 0.074 & 0.787 & 0.820 & 0.799 & 0.728 & 0.070 & 0.823 & 0.858 \\
UDUN       & 0.823 & 0.763 & 0.059 & 0.838 & 0.892 & 0.784 & 0.720 & 0.059 & 0.817 & 0.860 & 0.829 & 0.768 & 0.058 & 0.843 & 0.886 \\
BiRefNet   & 0.891 & 0.854 & 0.038 & 0.898 & 0.931 & 0.860 & 0.819 & 0.037 & 0.885 & 0.911 & 0.894 & 0.857 & 0.036 & 0.900 & 0.930 \\
SAM        & 0.835 & 0.782 & 0.069 & 0.808 & 0.889 & 0.838 & 0.807 & 0.047 & 0.843 & 0.805 & 0.803 & 0.758 & 0.081 & 0.792 & 0.863 \\
HQ-SAM     & 0.851 & 0.829 & 0.045 & 0.848 & 0.919 & 0.903 & 0.888 & 0.019 & 0.907 & 0.959 & 0.895 & 0.874 & 0.029 & 0.883 & 0.950 \\
Pi-SAM     & 0.883 & 0.866 & 0.035 & 0.889 & 0.945 & 0.890 & 0.869 & 0.027 & 0.894 & 0.947 & 0.903 & 0.887 & 0.027 & 0.907 & 0.953 \\
DIS-SAM    & 0.920 & 0.877 & 0.031 & 0.909 & 0.948 & 0.929 & 0.897 & 0.019 & 0.929 & 0.960 & 0.924 & 0.889 & 0.025 & 0.921 & 0.955 \\
\textbf{SAMA} & \textbf{0.942} & \textbf{0.885} & \textbf{0.021} & \textbf{0.930} & \textbf{0.962} & 
\textbf{0.940} & \textbf{0.911} & \textbf{0.012} & \textbf{0.947} & \textbf{0.977} &
\textbf{0.932} & \textbf{0.904} & \textbf{0.019} & \textbf{0.934} & \textbf{0.962} \\
\bottomrule

\end{tabular}

\vspace{1mm}

\begin{tabular}{c|ccccc|ccccc|ccccc}
\toprule
\multicolumn{1}{c|}{} & \multicolumn{5}{c|}{\textbf{DIS-TE3}} & \multicolumn{5}{c|}{\textbf{DIS-TE4}} & \multicolumn{5}{c}{\textbf{DIS-TE (ALL)}} \\
Method & $F_\beta^{\text{max}}$ & $F_\beta^{w}$ & $M$ & $S_\alpha$ & $E_\phi^{m}$ &
$F_\beta^{\text{max}}$ & $F_\beta^{w}$ & $M$ & $S_\alpha$ & $E_\phi^{m}$ &
$F_\beta^{\text{max}}$ & $F_\beta^{w}$ & $M$ & $S_\alpha$ & $E_\phi^{m}$ \\
\midrule
IS-Net     & 0.830 & 0.758 & 0.064 & 0.836 & 0.883 & 0.827 & 0.753 & 0.072 & 0.830 & 0.870 & 0.799 & 0.725 & 0.070 & 0.819 & 0.858 \\
UDUN       & 0.865 & 0.809 & 0.050 & 0.865 & 0.917 & 0.846 & 0.792 & 0.059 & 0.849 & 0.901 & 0.831 & 0.772 & 0.057 & 0.844 & 0.891 \\
BiRefNet   & \textbf{0.925} & \textbf{0.893} & \textbf{0.028} & 0.919 & \textbf{0.955} & 0.904 & \textbf{0.864} & \textbf{0.039} & \textbf{0.869} & \textbf{0.939} & 0.896 & 0.858 & 0.035 & 0.901 & 0.934 \\
SAM        & 0.773 & 0.724 & 0.094 & 0.761 & 0.848 & 0.677 & 0.634 & 0.162 & 0.697 & 0.762 & 0.773 & 0.731 & 0.096 & 0.773 & 0.845 \\
HQ-SAM     & 0.860 & 0.853 & 0.045 & 0.851 & 0.926 & 0.786 & 0.748 & 0.088 & 0.799 & 0.863 & 0.859 & 0.835 & 0.045 & 0.860 & 0.924 \\
Pi-SAM     & 0.899 & 0.882 & 0.030 & 0.901 & 0.953 & 0.893 & 0.870 & 0.039 & 0.893 & 0.948 & 0.893 & 0.873 & 0.033 & 0.893 & 0.948 \\
DIS-SAM    & 0.918 & 0.877 & 0.030 & 0.908 & 0.948 & 0.899 & 0.849 & 0.043 & 0.888 & 0.932 & 0.917 & 0.872 & 0.029 & 0.911 & 0.949 \\
\textbf{SAMA} & 0.920 & 0.889 & 0.032 & \textbf{0.924} & 0.949 & 
\textbf{0.917} & 0.857 & 0.041 & 0.897 & 0.937 & 
\textbf{0.926} & \textbf{0.897} & \textbf{0.026} & \textbf{0.925} & \textbf{0.956} \\
\bottomrule
\end{tabular}
\caption{Comparison on DIS datasets. Higher $\uparrow$ is better except for $M$, where lower $\downarrow$ is better.}
\label{dis_res}
\end{table*}

\subsection{Comparison Study on Matting}
To assess the general matting performance of our proposed models, we conduct evaluations on two widely adopted benchmarks: Composition-1K~\cite{xu2017deep} and Distinctions-646~\cite{qiao2020attention}. We compare our SAMA with two categories of matting approaches: (1) trimap-free methods such as LFM ~\cite{zhang2019late}, MODNet ~\cite{ke2022modnet}, and MFC-Net ~\cite{zhao2024boosting}, and (2) trimap-based methods that leverage additional trimap guidance, including Information-Flow (I-F) ~\cite{aksoy2017designing}, DIM ~\cite{xu2017deep}, DCNN ~\cite{cho2016natural}, MGMatting ~\cite{yu2021mask}, and VITMatte ~\cite{yao2024vitmatte}. To quantify performance on alpha matting task, we use commonly adopted evaluation metrics including Sum of Absolute Difference (SAD) and Mean Squared Error (MSE).

As reported in Table ~\ref{matting_general}, our model achieves state-of-the-art performance among trimap-free methods on both benchmarks. Notably, SAMA, without relying on any trimap input, demonstrates substantial improvements across diverse visual scenes. When compared to leading trimap-based approaches, such as VITMatte, our SAMA achieves comparable results, highlighting its strong potential to generalize well in real-world matting scenarios while maintaining the advantages of a trimap-free framework.

\begin{table*}[t]
  \centering
  \begin{adjustbox}{max width=0.8\linewidth}
   \begin{tabular}{llccccccccccc}
\toprule
& & \multicolumn{5}{c}{\textbf{Trimap-based}} & \multicolumn{4}{c}{\textbf{Trimap-free}} \\
\cmidrule(lr){3-7} \cmidrule(lr){8-11}
    \textbf{Dataset} & \textbf{Metric}  & I-F & DIM &
    DCNN &  MGMatting  & VITMatte &
    LFM & MODNet & MFC-Net & SAMA \\
    \midrule
    \multirow{2}{*}{Composition‑1K} 
        & SAD\,$\downarrow$   & 70.3  &  50.4 & 115.8 & 32.1  & 21.5 & 58.4 & 47.1 & 35.6 & 22.8 \\
        & MSE\,$\downarrow$    & 13    &  14   & 23   & 7.0  &  3.3 & 11.8 & 12.3 & 8.7 &  2.9  \\
    \midrule
    \multirow{2}{*}{Distinction‑646}
        & SAD\,$\downarrow$      & 78.9    &  47.6  & 103.8   & 36.6    & 21.22   & 44.6   & 41.7  & 34.5   & 22.4   \\
        & MSE\,$\downarrow$    & 16   &  9   & 20   & 7.2   & 2.1   & 12.8   & 9.0   & 7.8   & 2.2   \\
    \bottomrule
  \end{tabular}
  \end{adjustbox}
  \caption{Quantitative results on Composition‑1K and Distinction‑646 test sets. Models before VITMatte are trimap-based, while those after are trimap-free. Lower values indicate better performance.}
  \label{matting_general}
\end{table*}

\subsection{Visual Results Comparison}
\begin{figure}[t]
  \centering
  \begin{subfigure}[b]{0.09\textwidth}
    \includegraphics[width=\linewidth]{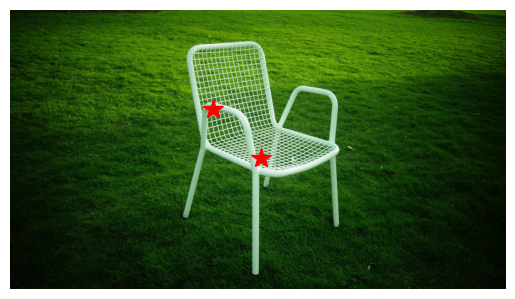}
  \end{subfigure}
  \begin{subfigure}[b]{0.09\textwidth}
    \includegraphics[width=\linewidth]{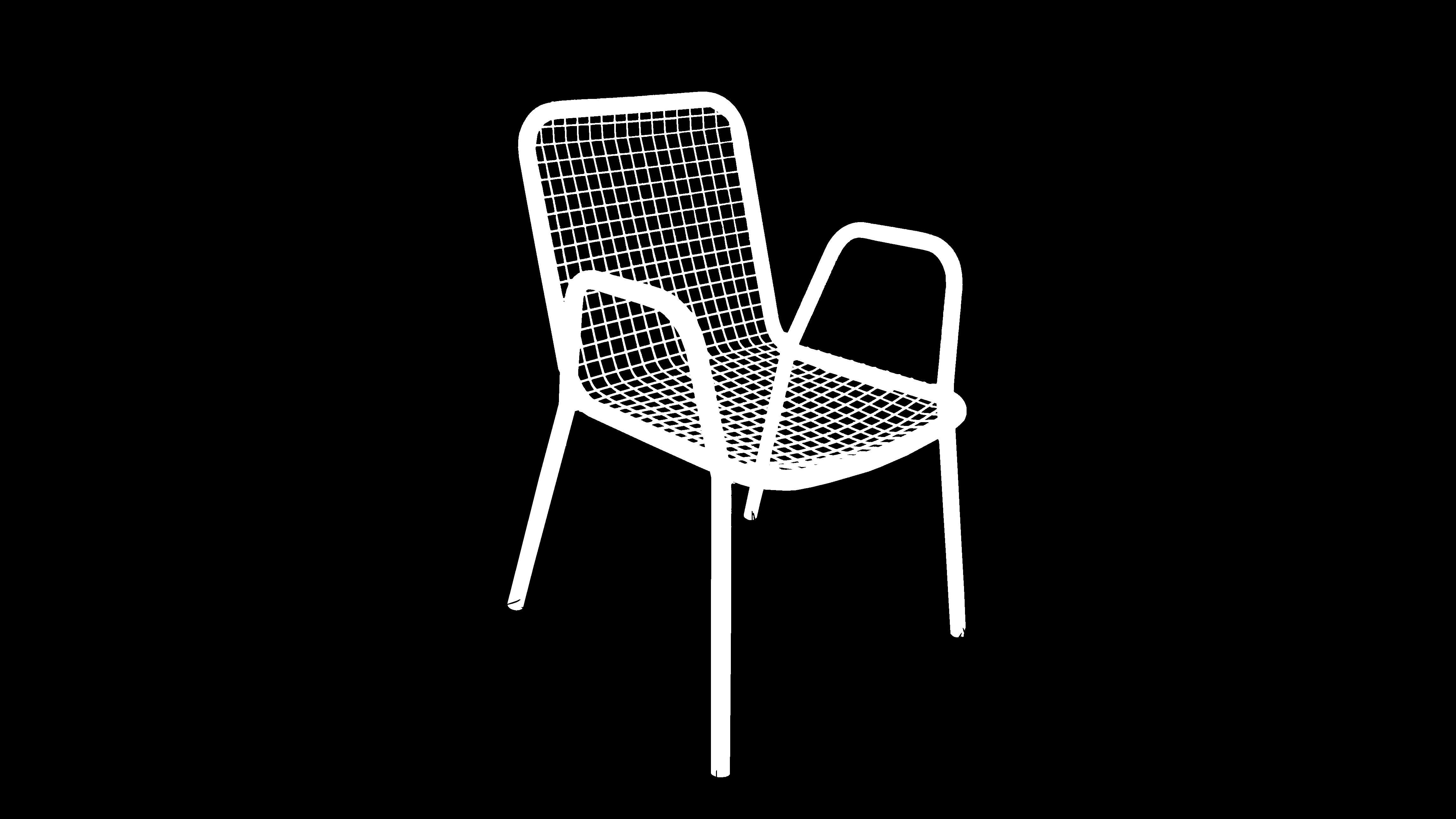}
  \end{subfigure}
  \begin{subfigure}[b]{0.09\textwidth}
    \includegraphics[width=\linewidth]{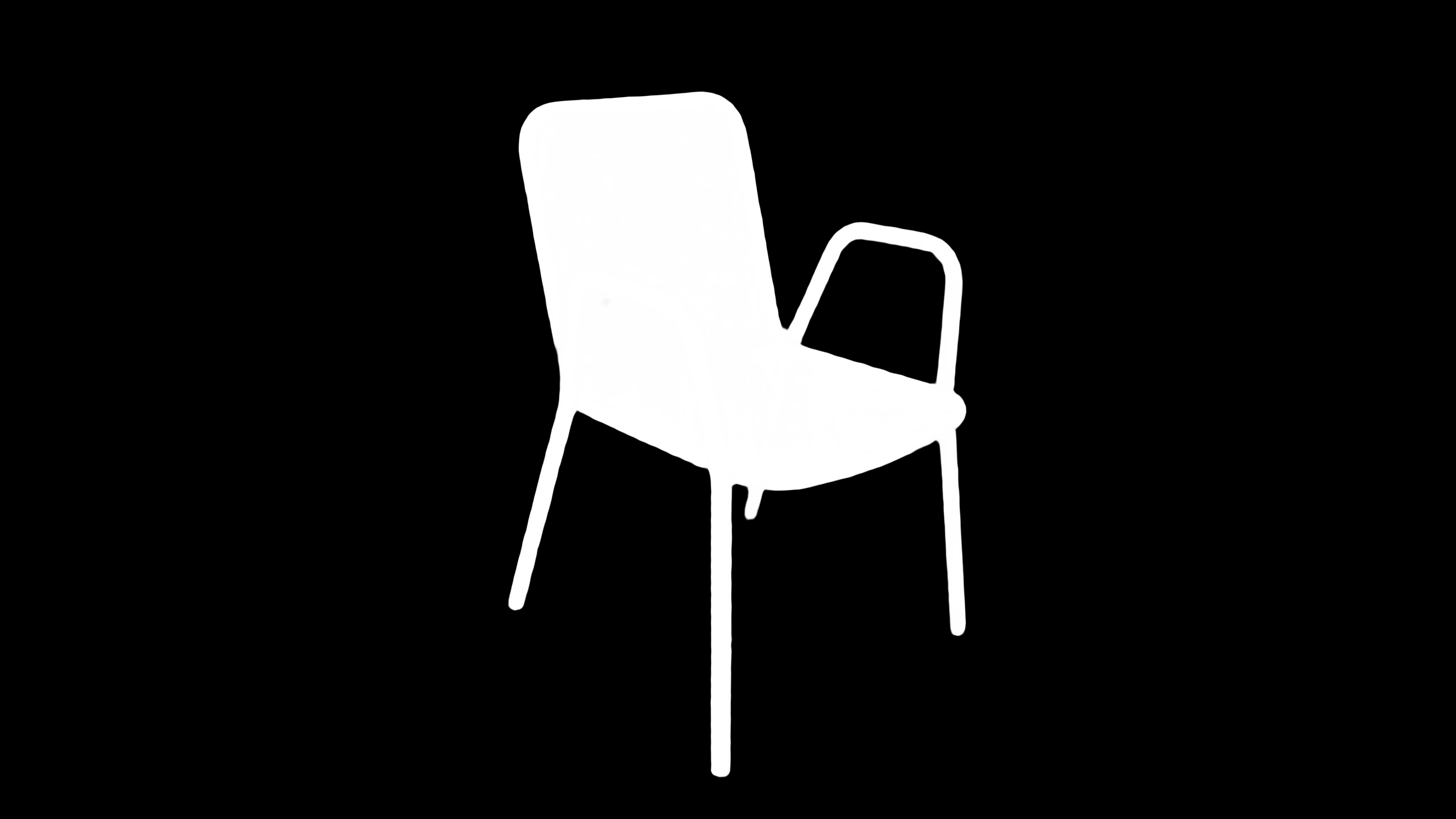}
  \end{subfigure}
  \begin{subfigure}[b]{0.09\textwidth}
    \includegraphics[width=\linewidth]{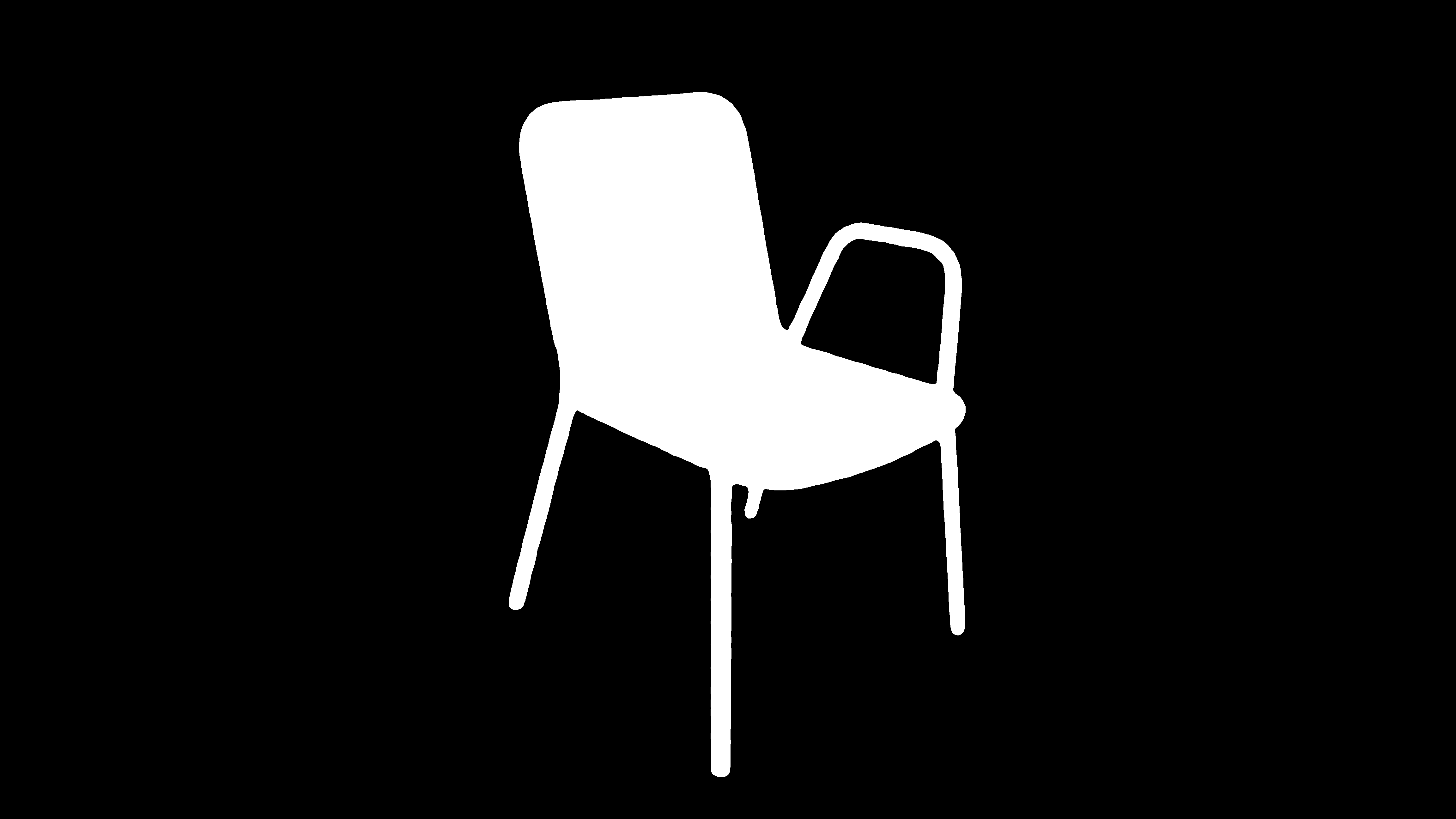}
  \end{subfigure}
  \begin{subfigure}[b]{0.09\textwidth}
    \includegraphics[width=\linewidth]{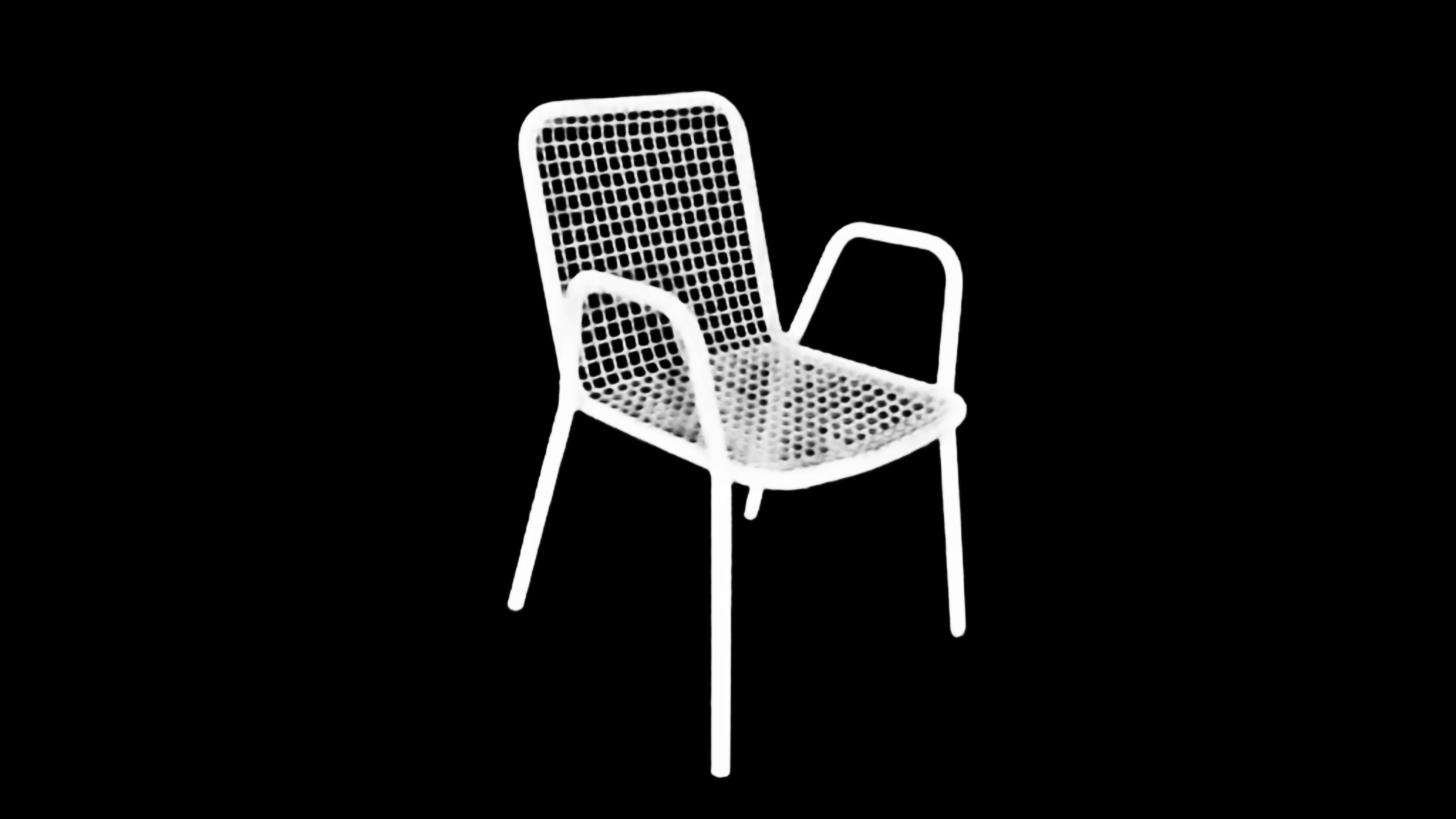}
  \end{subfigure}

  \begin{subfigure}[b]{0.09\textwidth}
    \includegraphics[width=\linewidth]{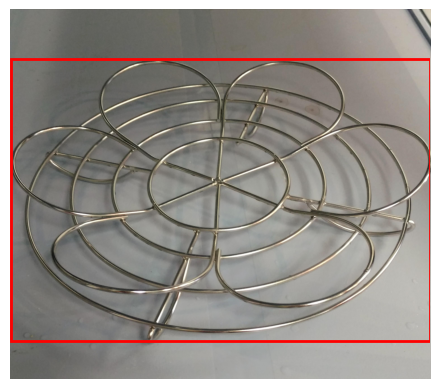}
  \end{subfigure}
  \begin{subfigure}[b]{0.09\textwidth}
    \includegraphics[width=\linewidth]{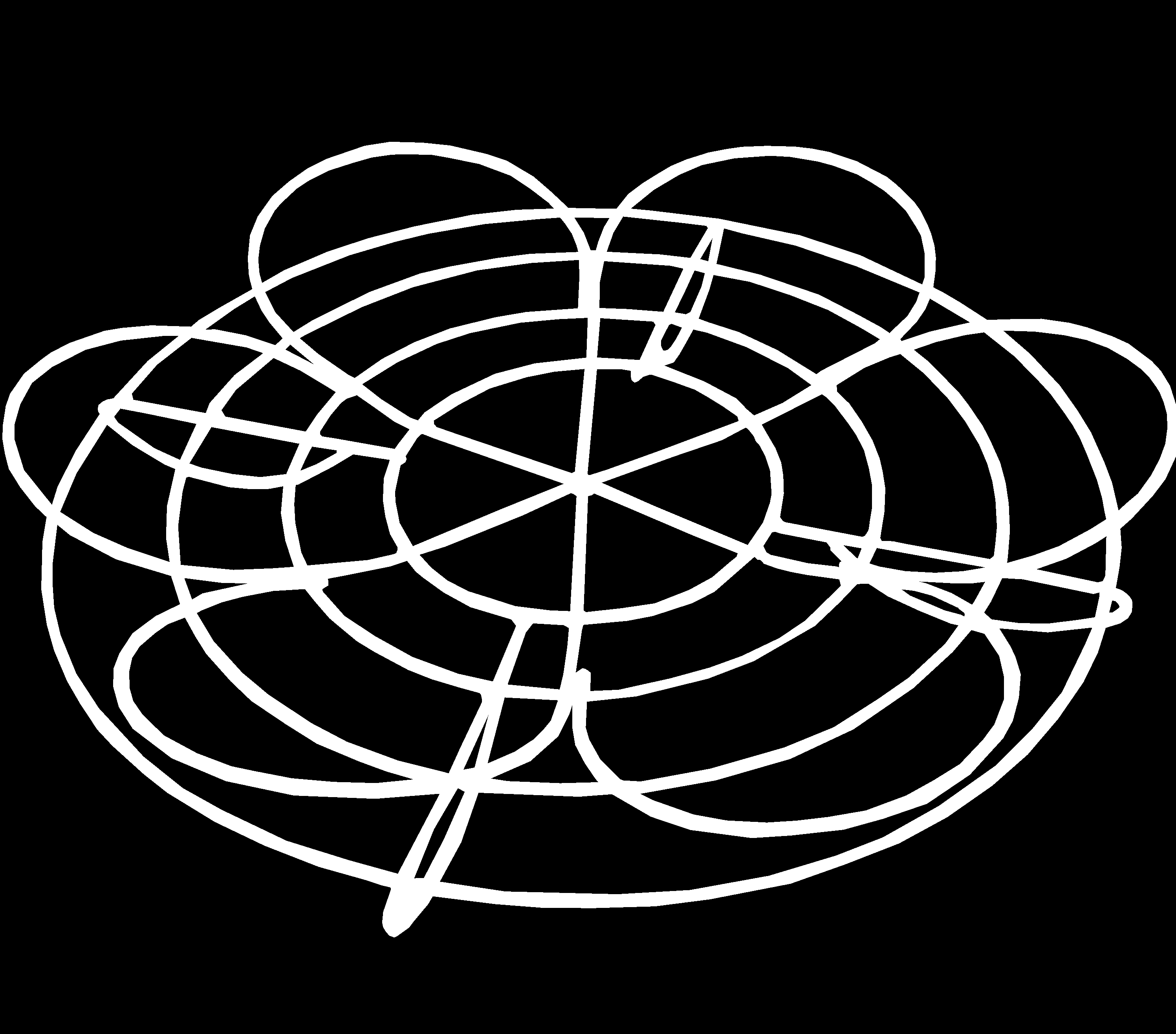}
  \end{subfigure}
  \begin{subfigure}[b]{0.09\textwidth}
    \includegraphics[width=\linewidth]{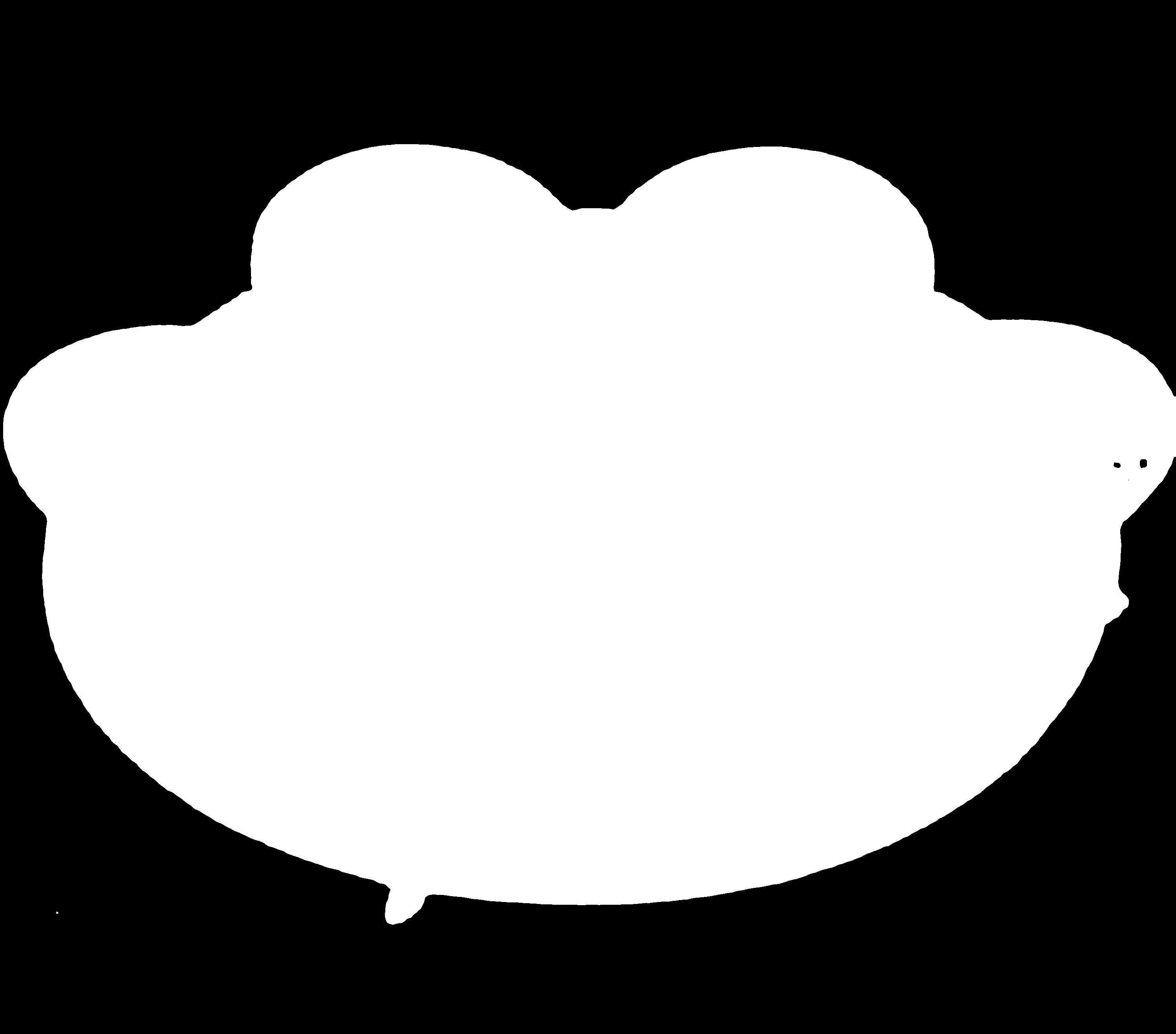}
  \end{subfigure}
  \begin{subfigure}[b]{0.09\textwidth}
    \includegraphics[width=\linewidth]{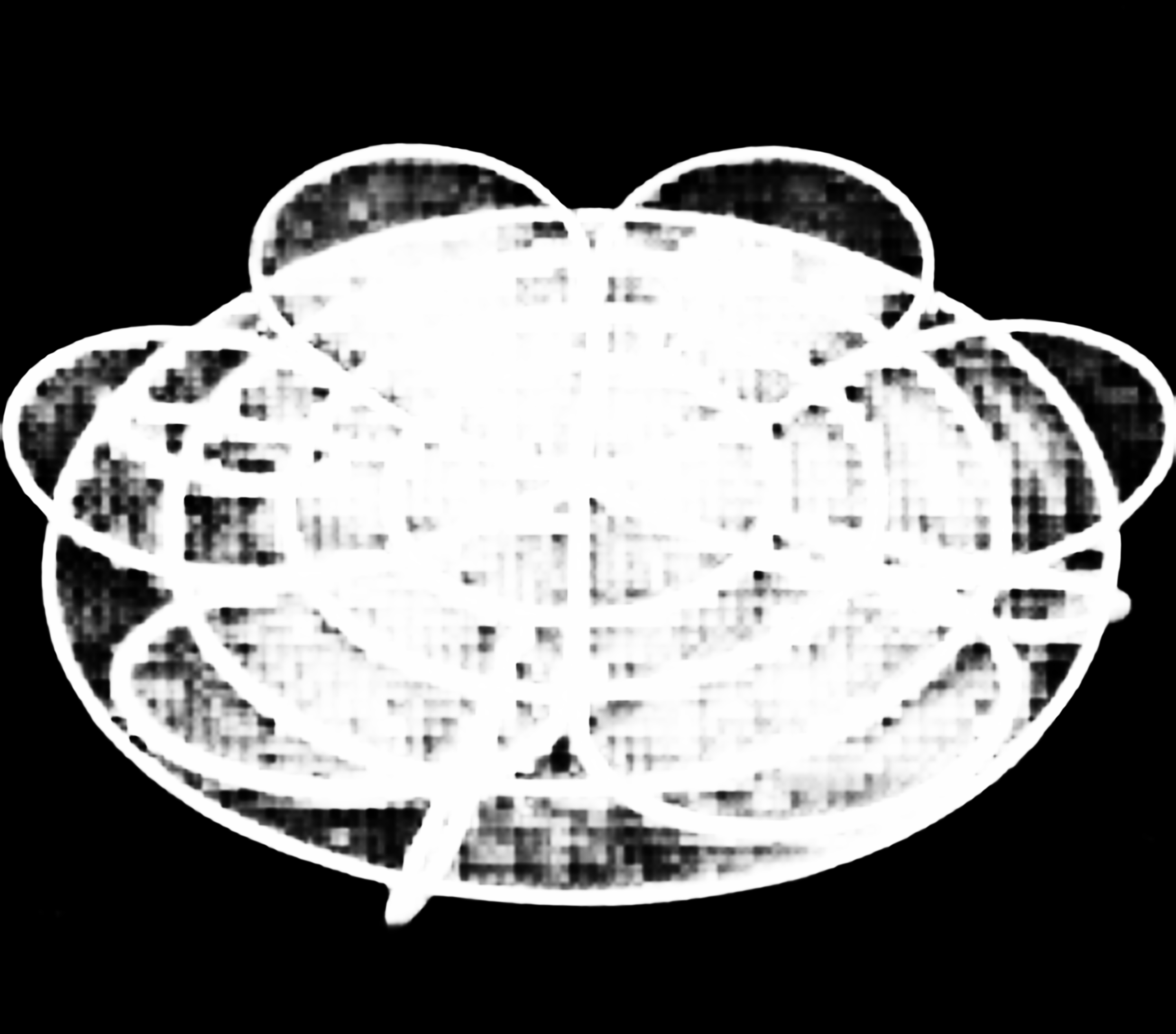}
  \end{subfigure}
  \begin{subfigure}[b]{0.09\textwidth}
    \includegraphics[width=\linewidth]{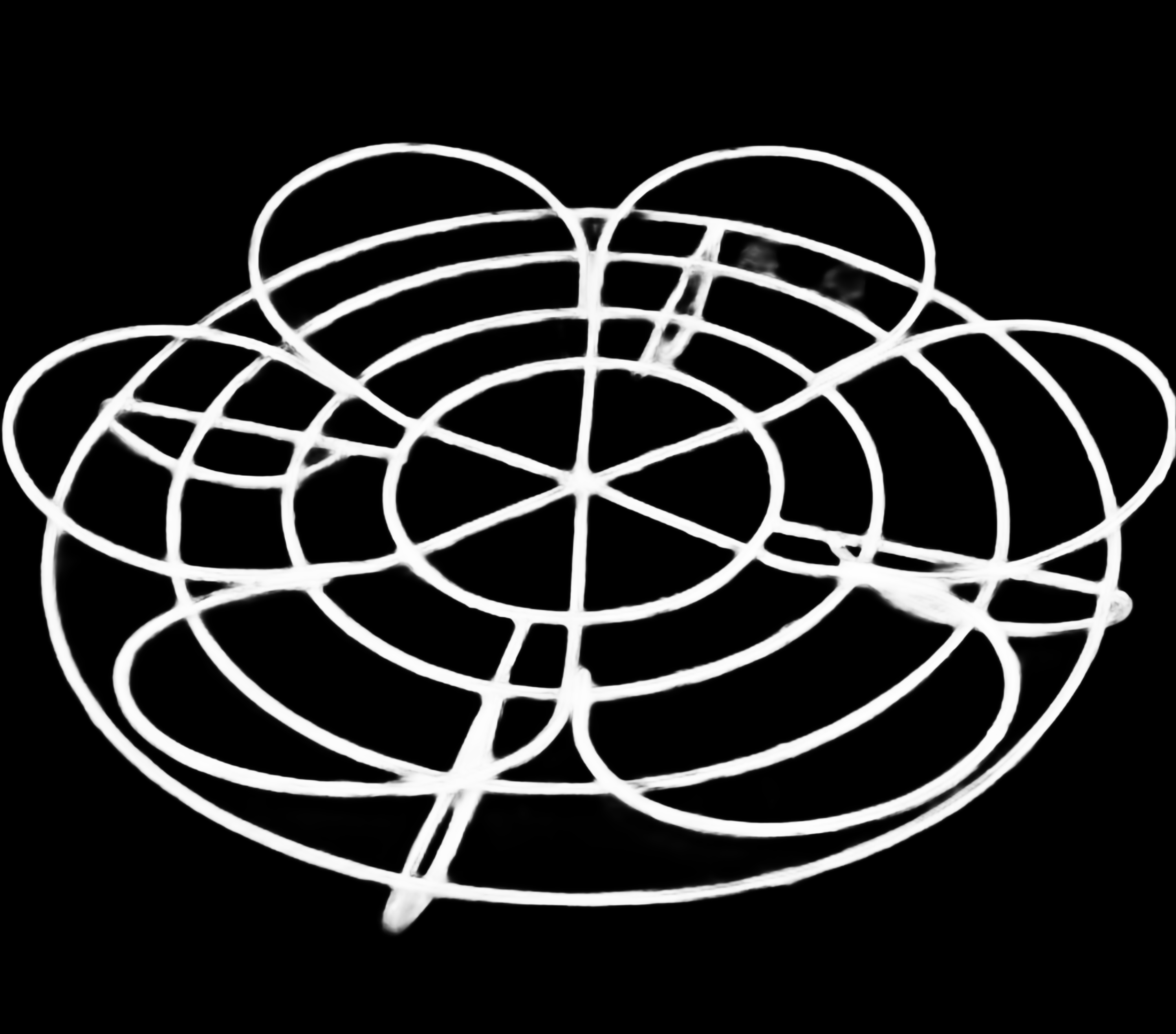}
  \end{subfigure}

  \begin{subfigure}[b]{0.09\textwidth}
    \includegraphics[width=\linewidth]{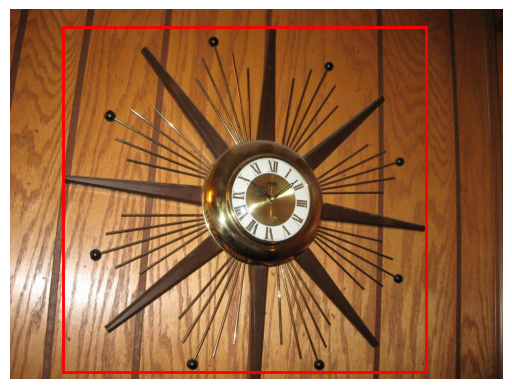}
  \end{subfigure}
  \begin{subfigure}[b]{0.09\textwidth}
    \includegraphics[width=\linewidth]{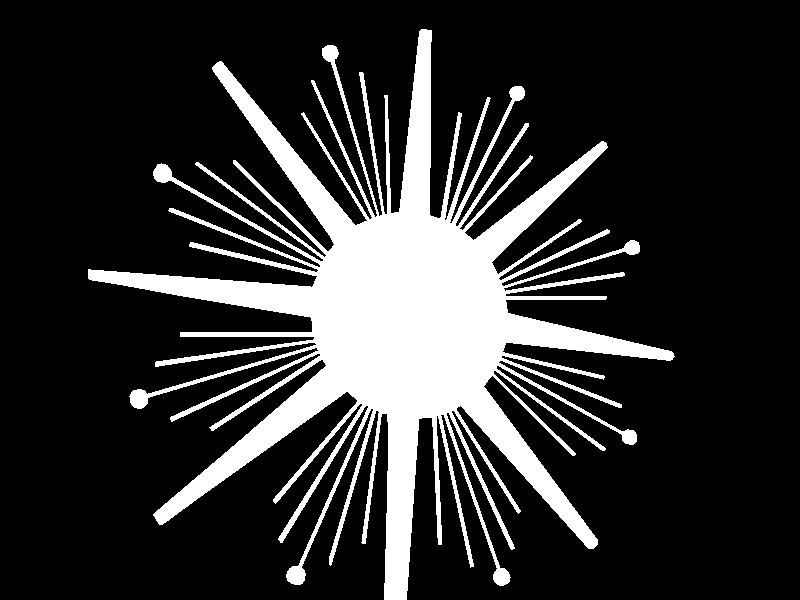}
  \end{subfigure}
  \begin{subfigure}[b]{0.09\textwidth}
    \includegraphics[width=\linewidth]{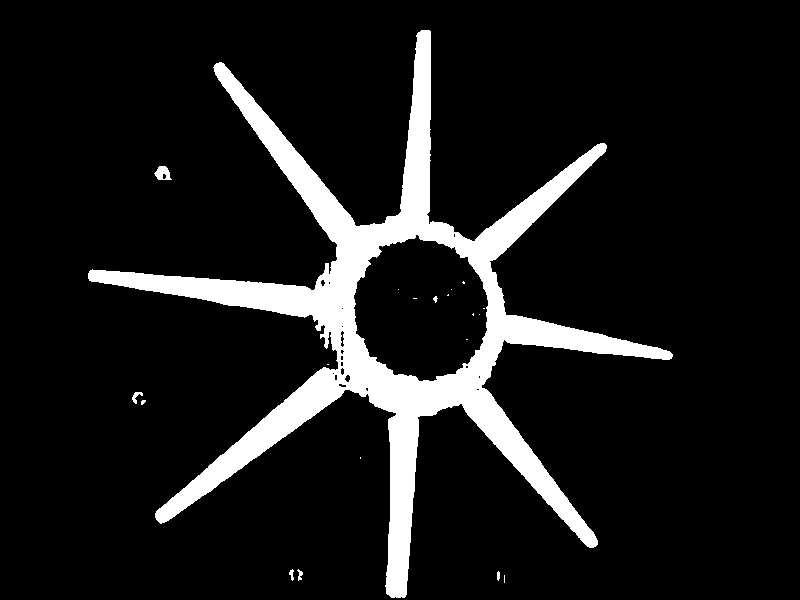}
  \end{subfigure}
  \begin{subfigure}[b]{0.09\textwidth}
    \includegraphics[width=\linewidth]{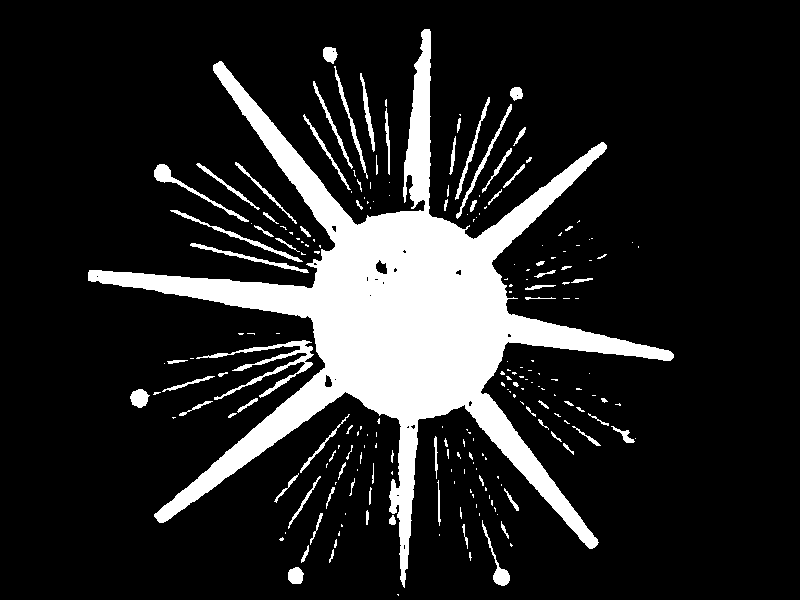}
  \end{subfigure}
  \begin{subfigure}[b]{0.09\textwidth}
    \includegraphics[width=\linewidth]{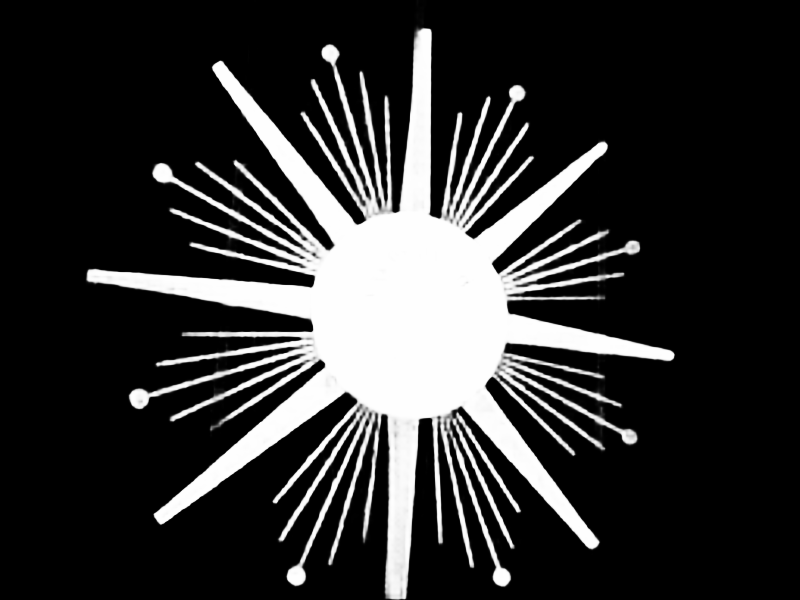}
  \end{subfigure}

  \begin{subfigure}[b]{0.09\textwidth}
    \includegraphics[width=\linewidth]{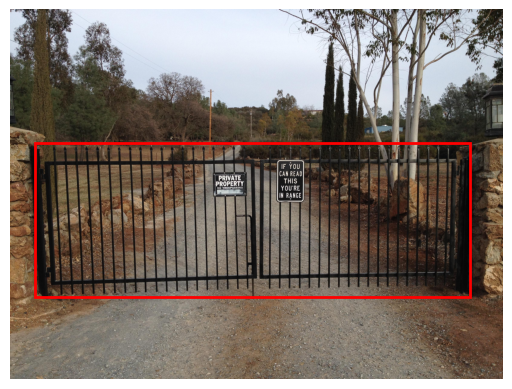}
  \end{subfigure}
  \begin{subfigure}[b]{0.09\textwidth}
    \includegraphics[width=\linewidth]{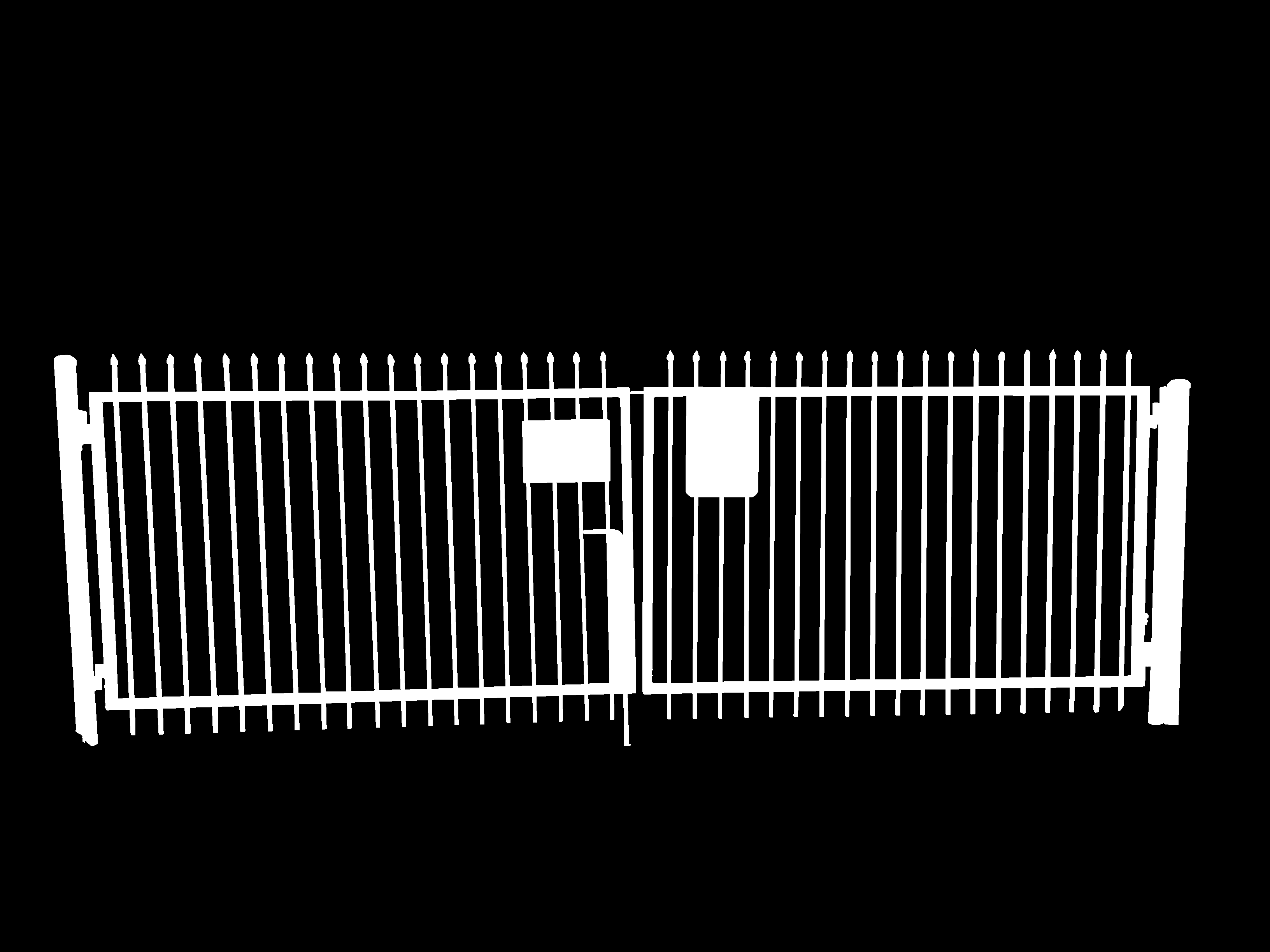}
  \end{subfigure}
  \begin{subfigure}[b]{0.09\textwidth}
    \includegraphics[width=\linewidth]{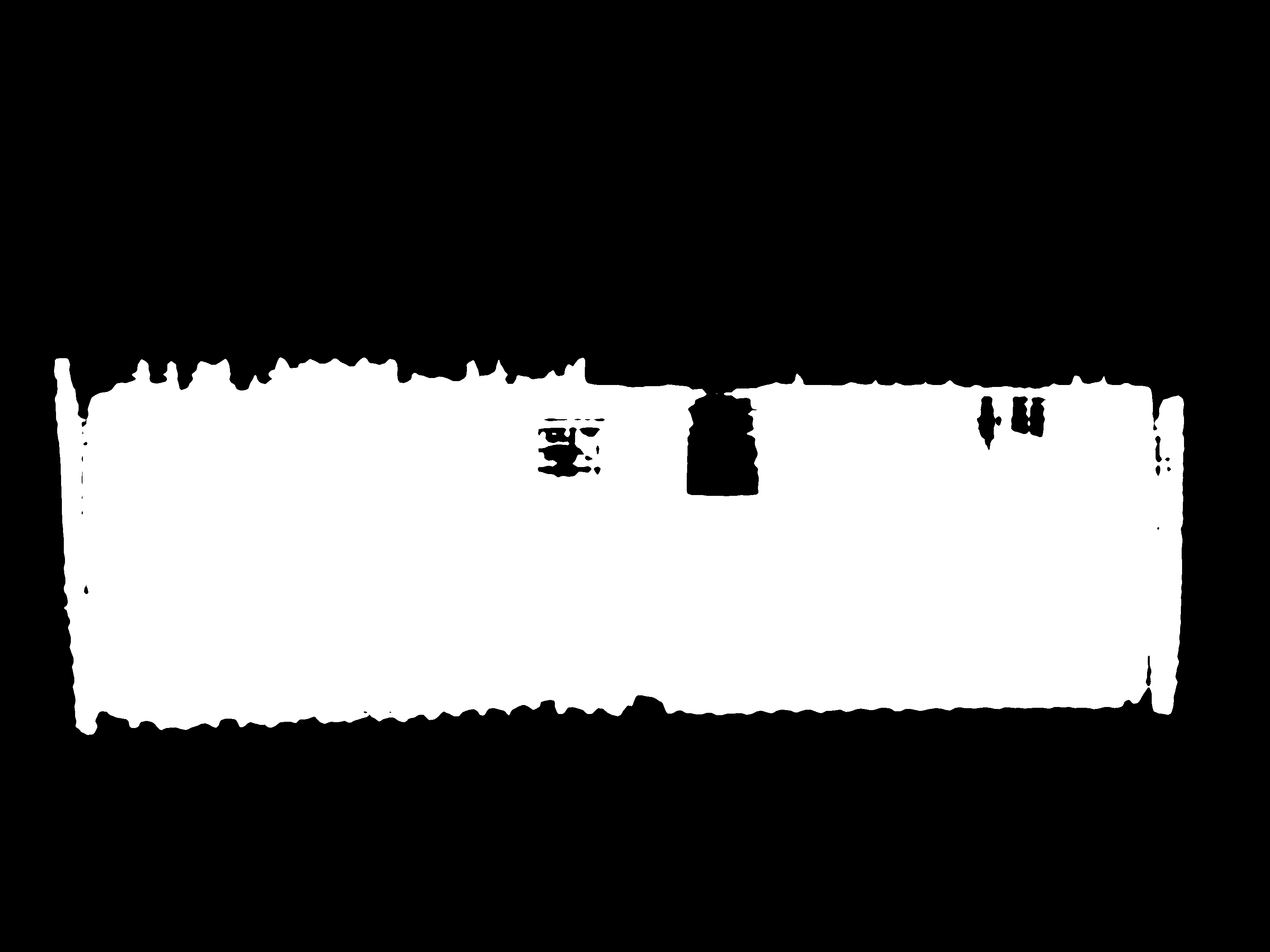}
  \end{subfigure}
  \begin{subfigure}[b]{0.09\textwidth}
    \includegraphics[width=\linewidth]{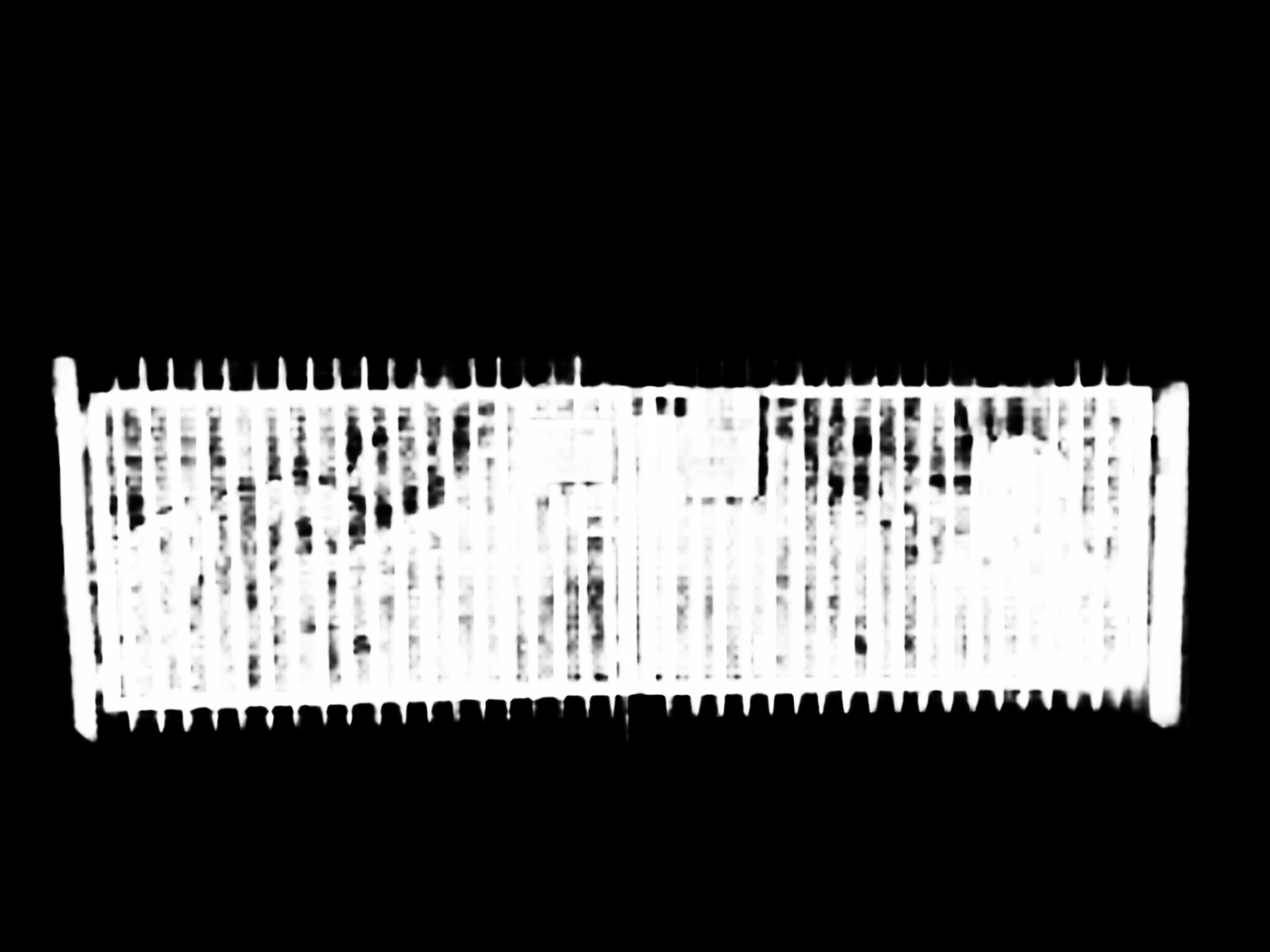}
  \end{subfigure}
  \begin{subfigure}[b]{0.09\textwidth}
    \includegraphics[width=\linewidth]{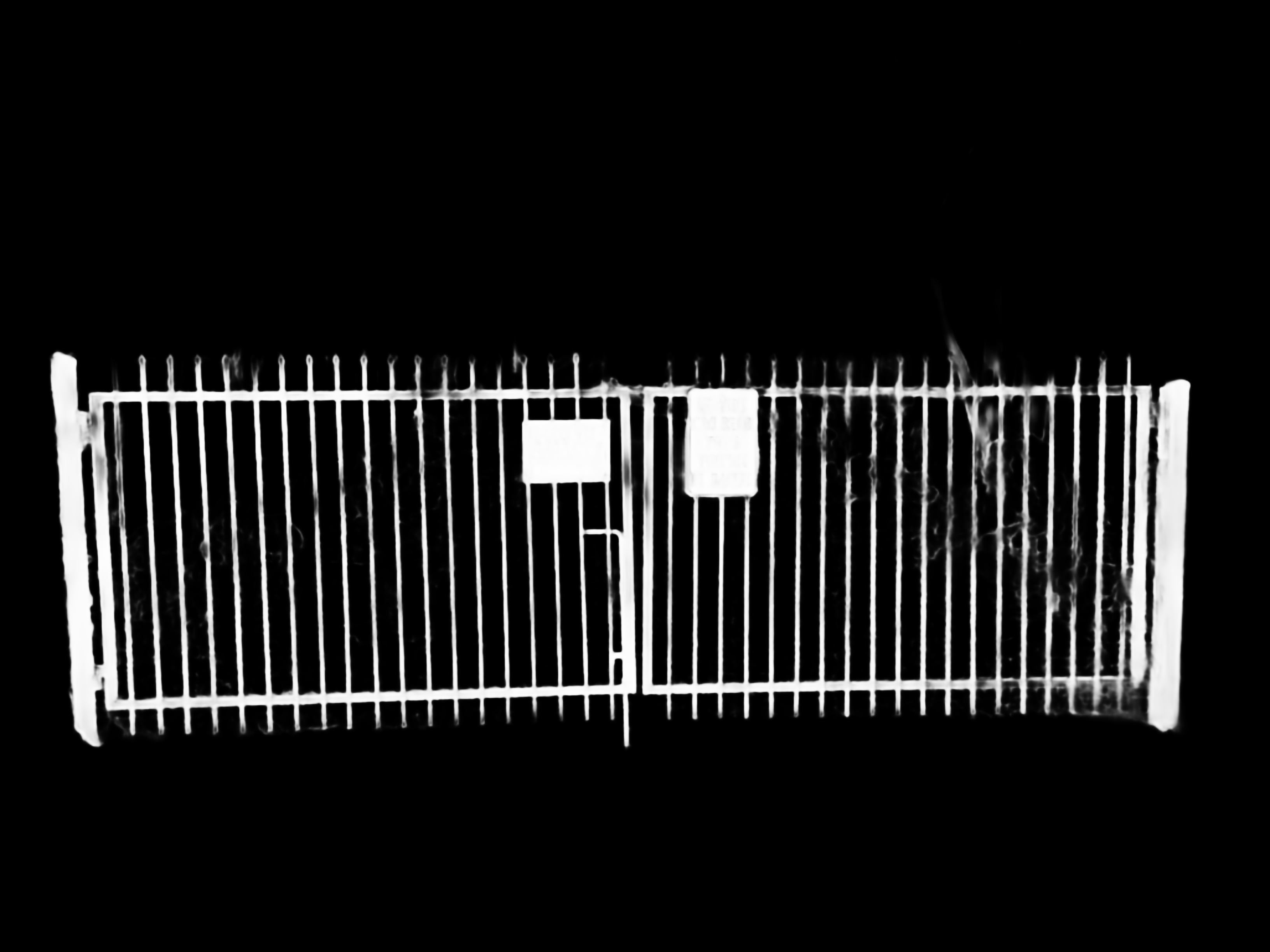}
  \end{subfigure}

  \begin{subfigure}[b]{0.09\textwidth}
    \includegraphics[width=\linewidth]{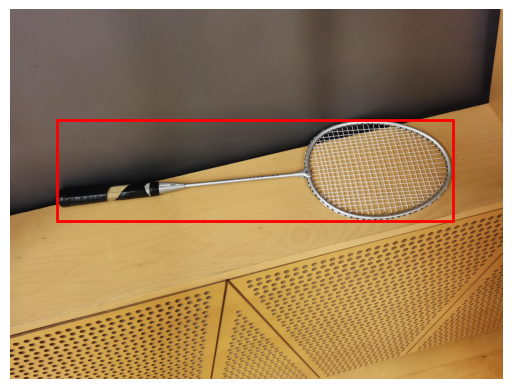}
  \end{subfigure}
  \begin{subfigure}[b]{0.09\textwidth}
    \includegraphics[width=\linewidth]{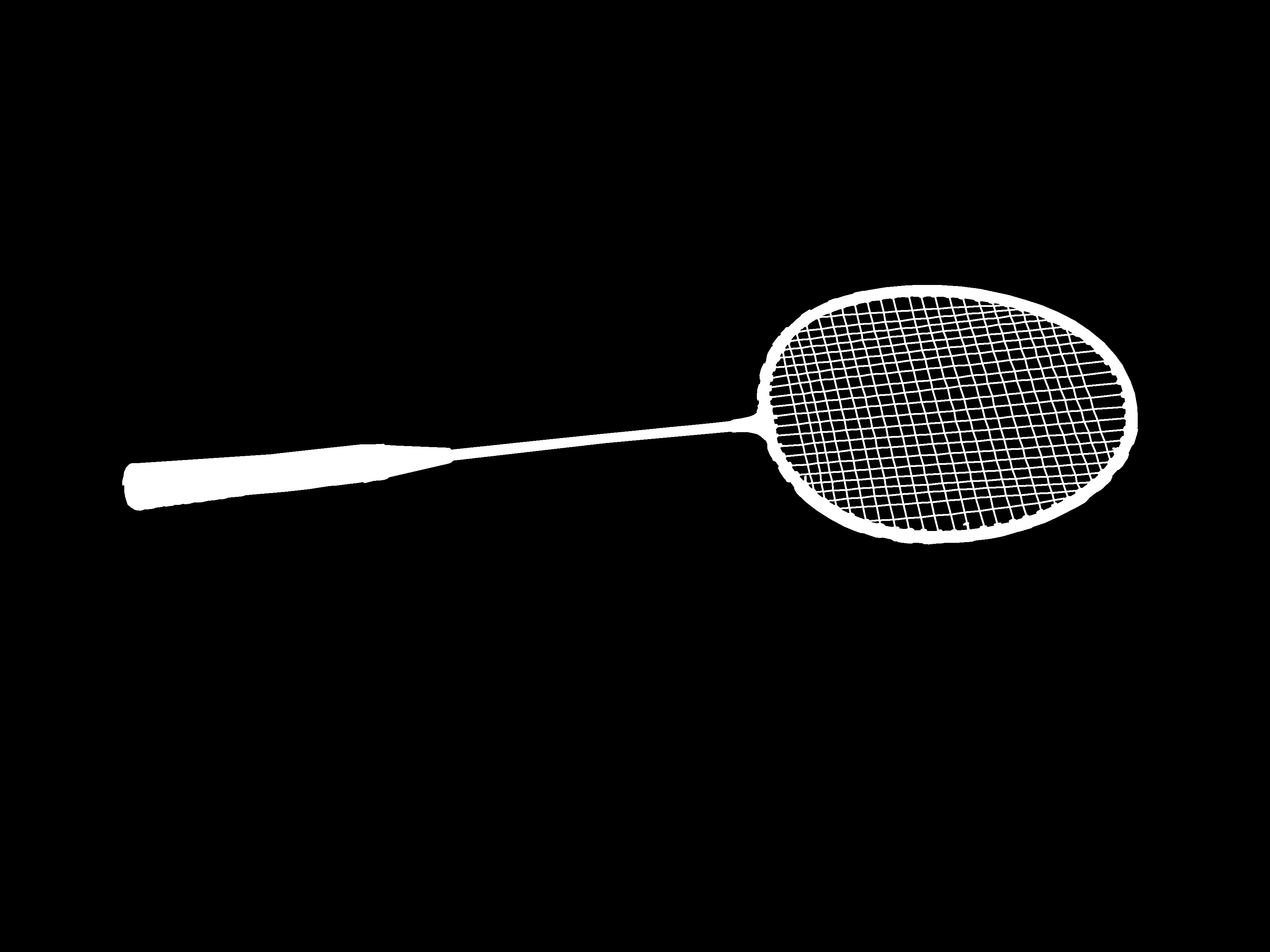}
  \end{subfigure}
  \begin{subfigure}[b]{0.09\textwidth}
    \includegraphics[width=\linewidth]{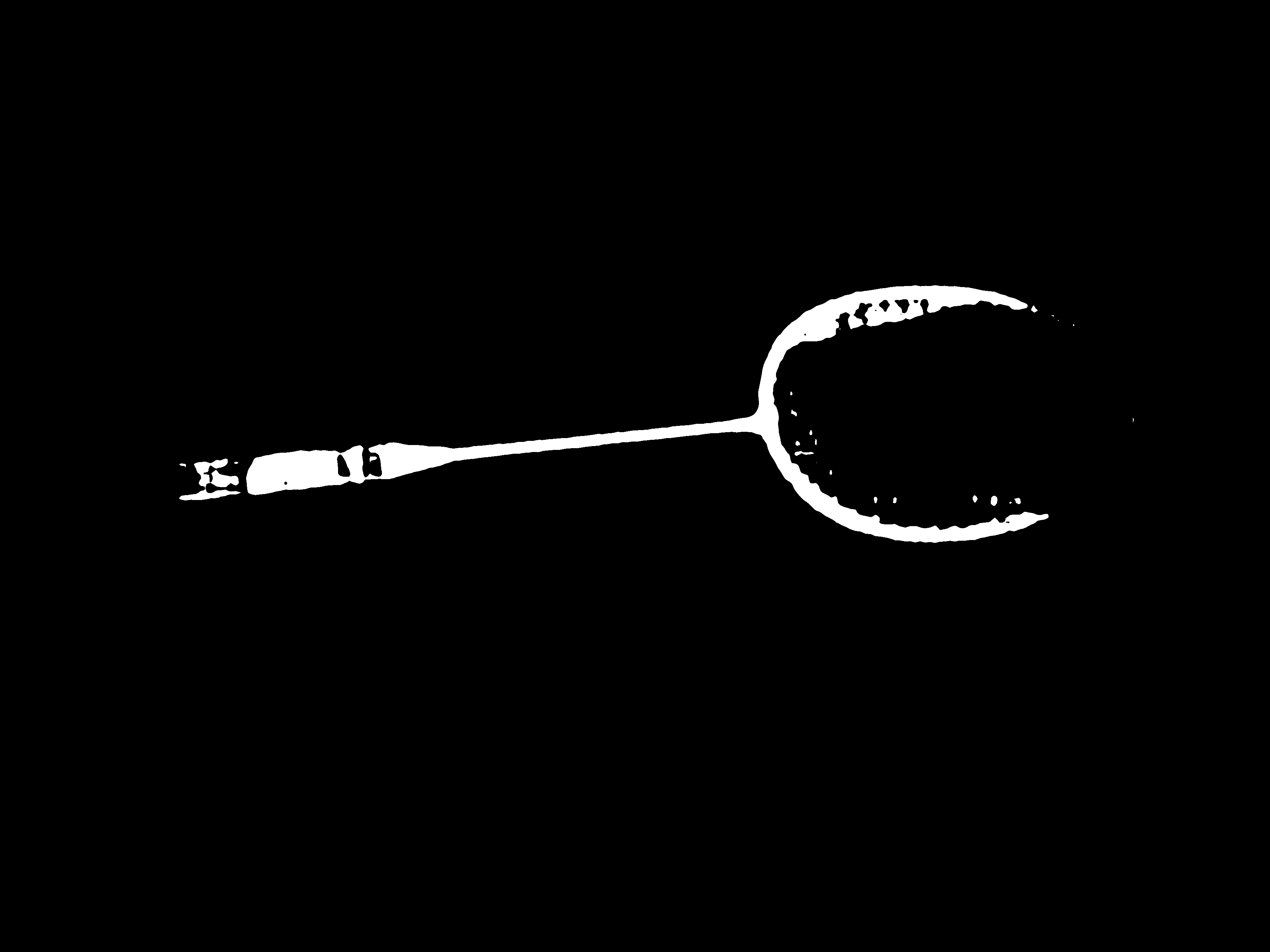}
  \end{subfigure}
  \begin{subfigure}[b]{0.09\textwidth}
    \includegraphics[width=\linewidth]{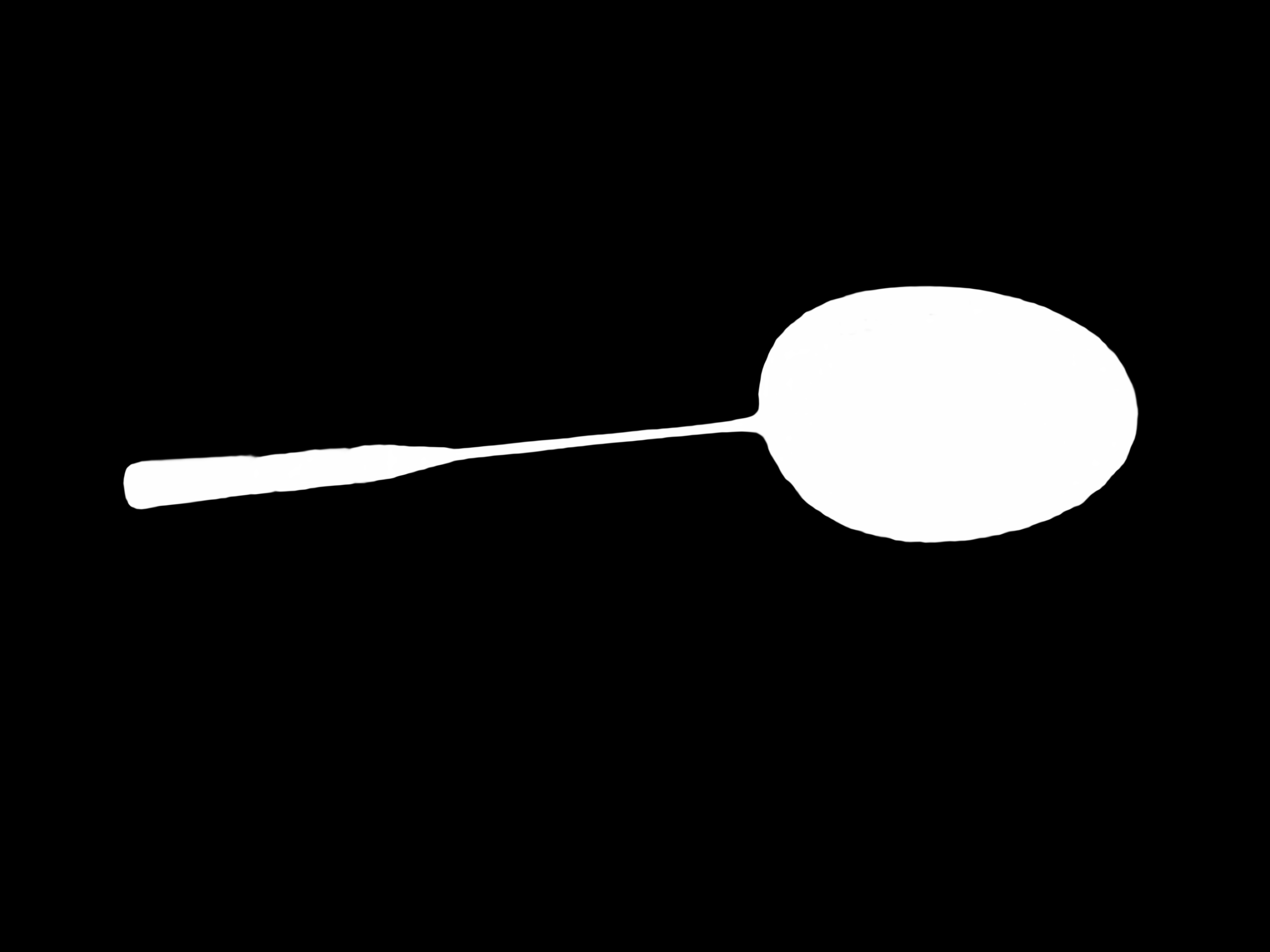}
  \end{subfigure}
  \begin{subfigure}[b]{0.09\textwidth}
    \includegraphics[width=\linewidth]{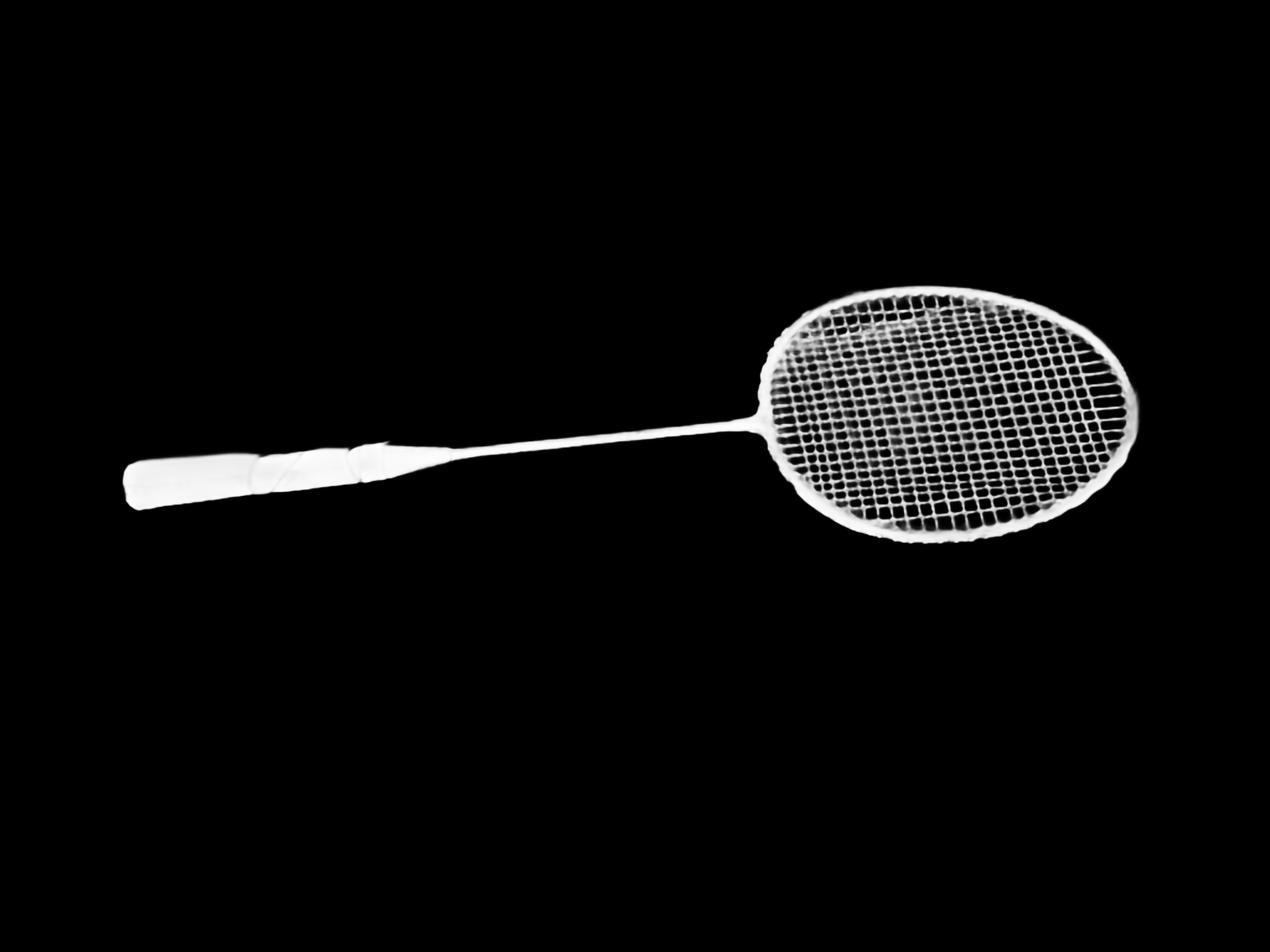}
  \end{subfigure}

  \begin{subfigure}[b]{0.09\textwidth}
    \includegraphics[width=\linewidth]{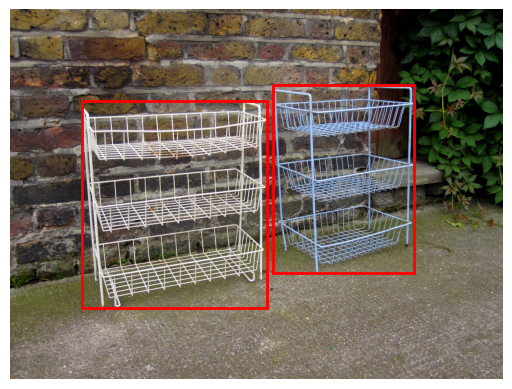}
    \caption*{Image}
  \end{subfigure}
  \begin{subfigure}[b]{0.09\textwidth}
    \includegraphics[width=\linewidth]{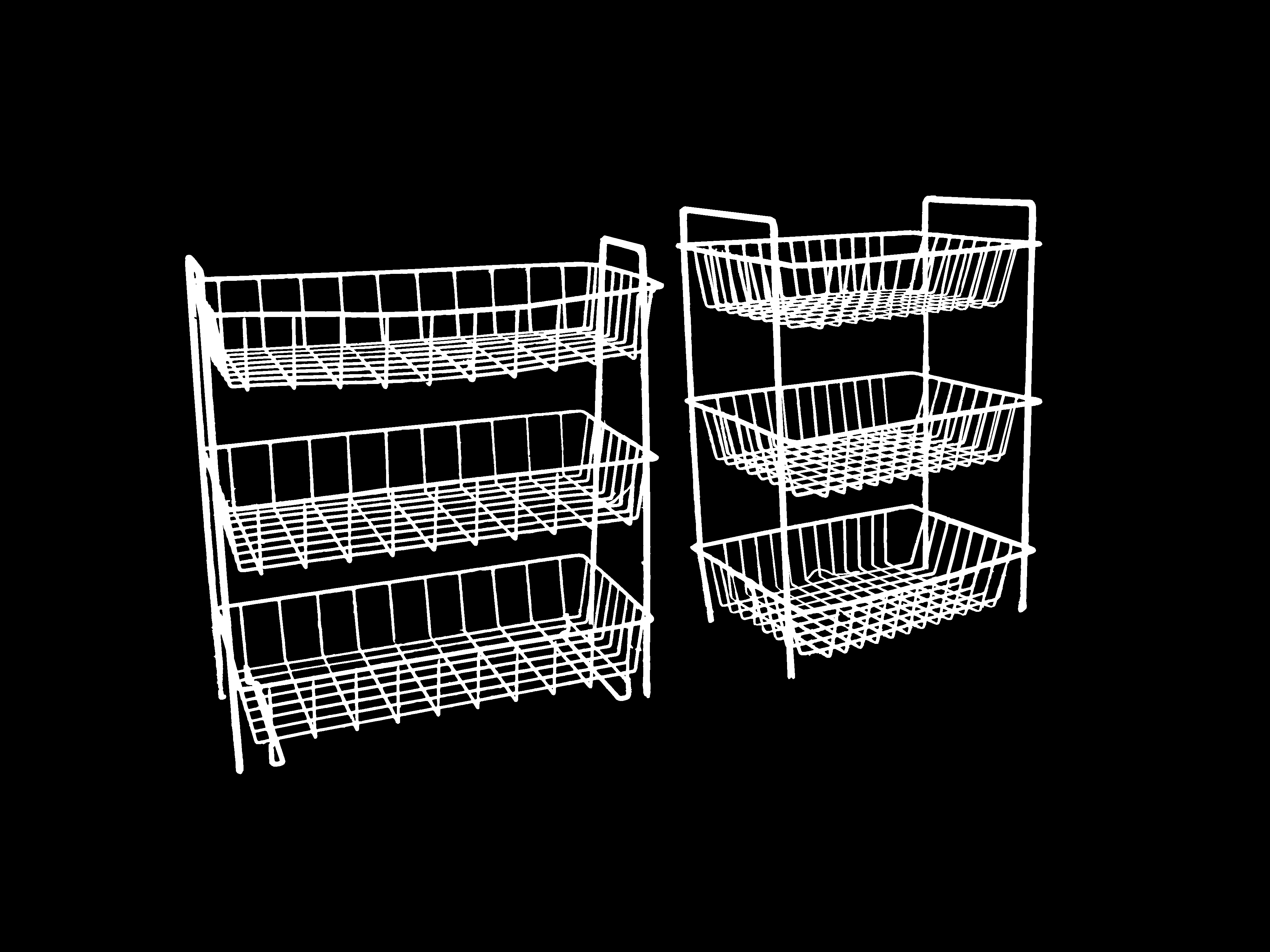}
    \caption*{GT}
  \end{subfigure}
  \begin{subfigure}[b]{0.09\textwidth}
    \includegraphics[width=\linewidth]{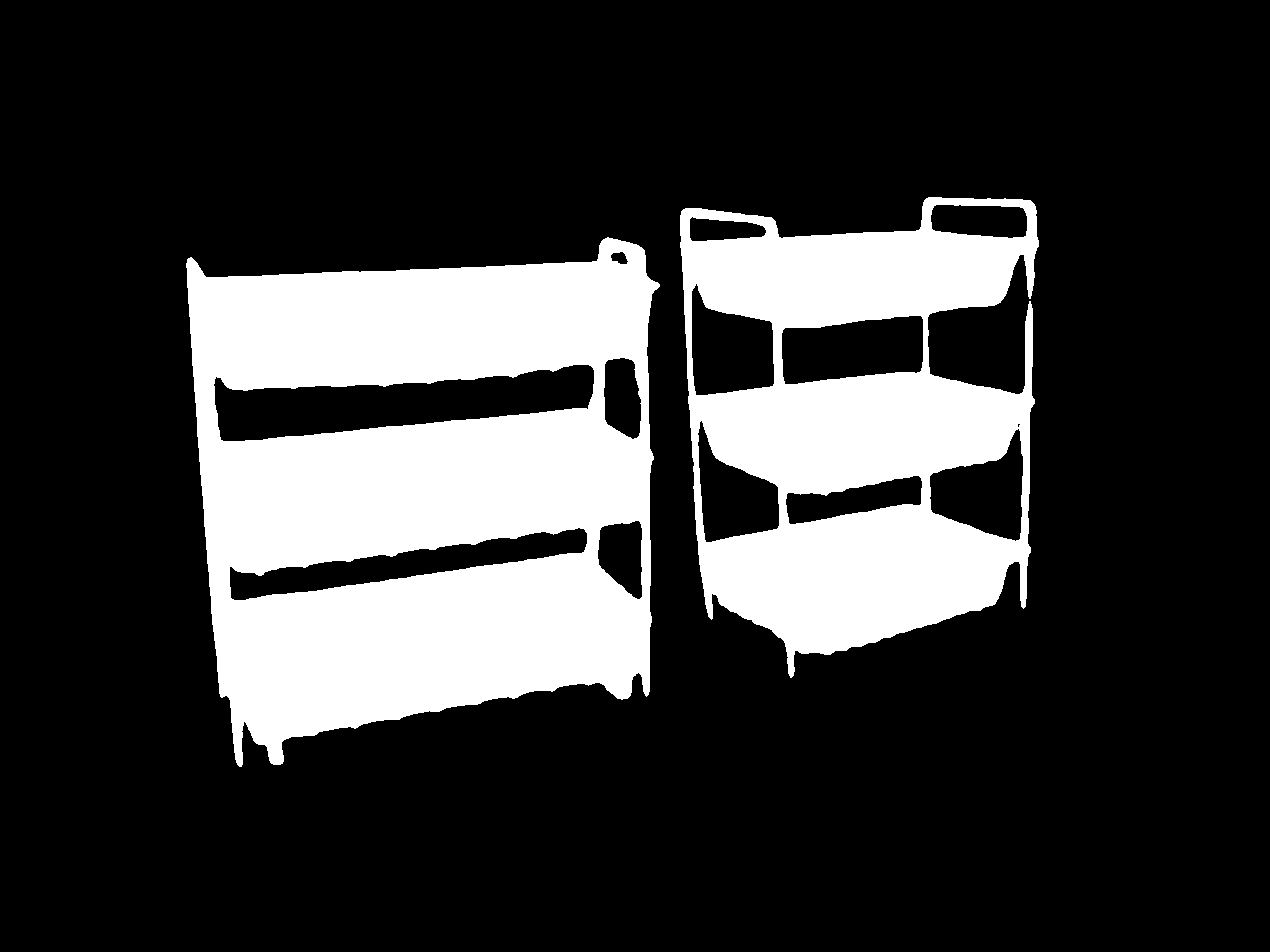}
    \caption*{SAM}
  \end{subfigure}
  \begin{subfigure}[b]{0.09\textwidth}
    \includegraphics[width=\linewidth]{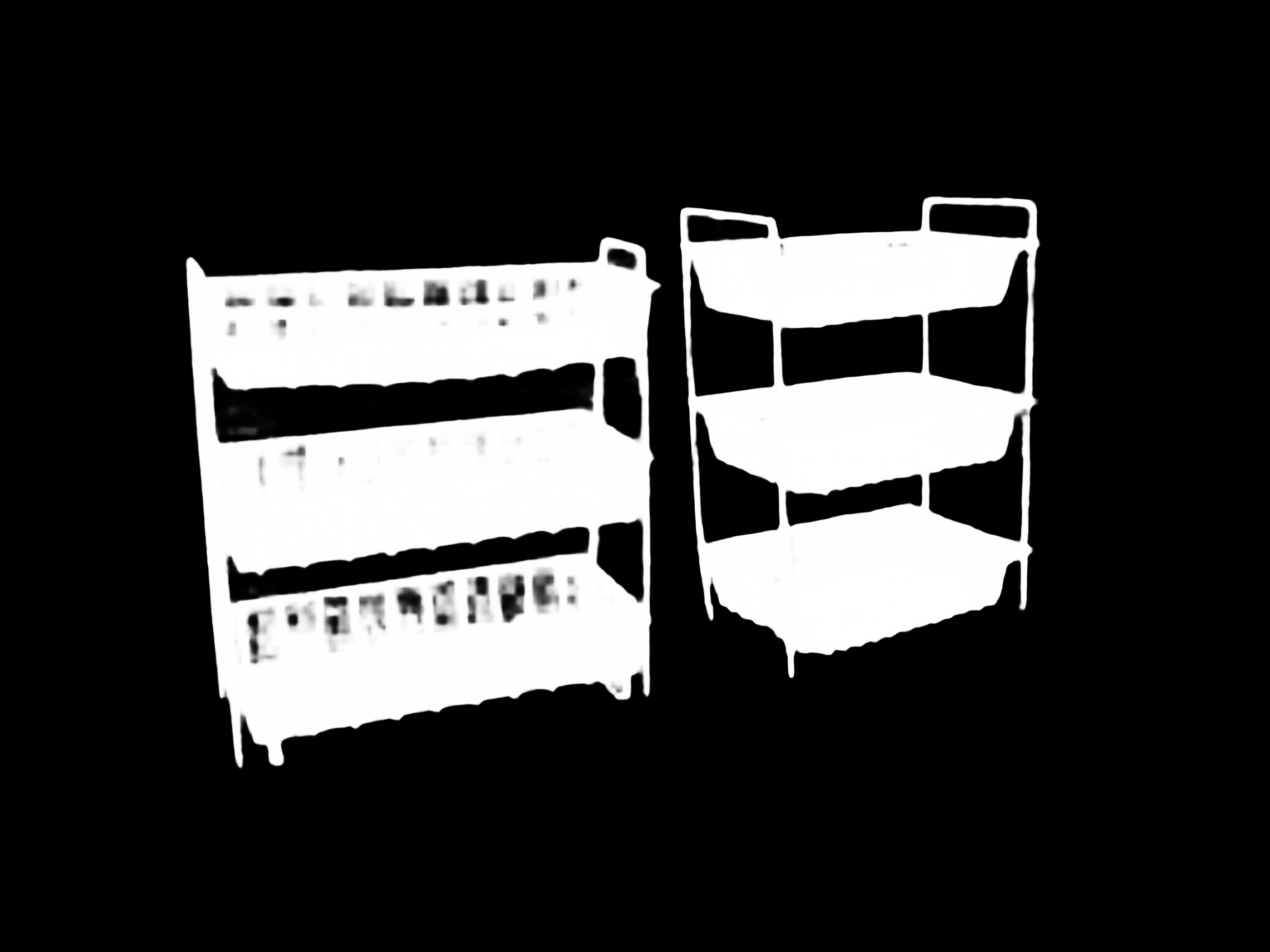}
    \caption*{HQ-SAM}
  \end{subfigure}
  \begin{subfigure}[b]{0.09\textwidth}
    \includegraphics[width=\linewidth]{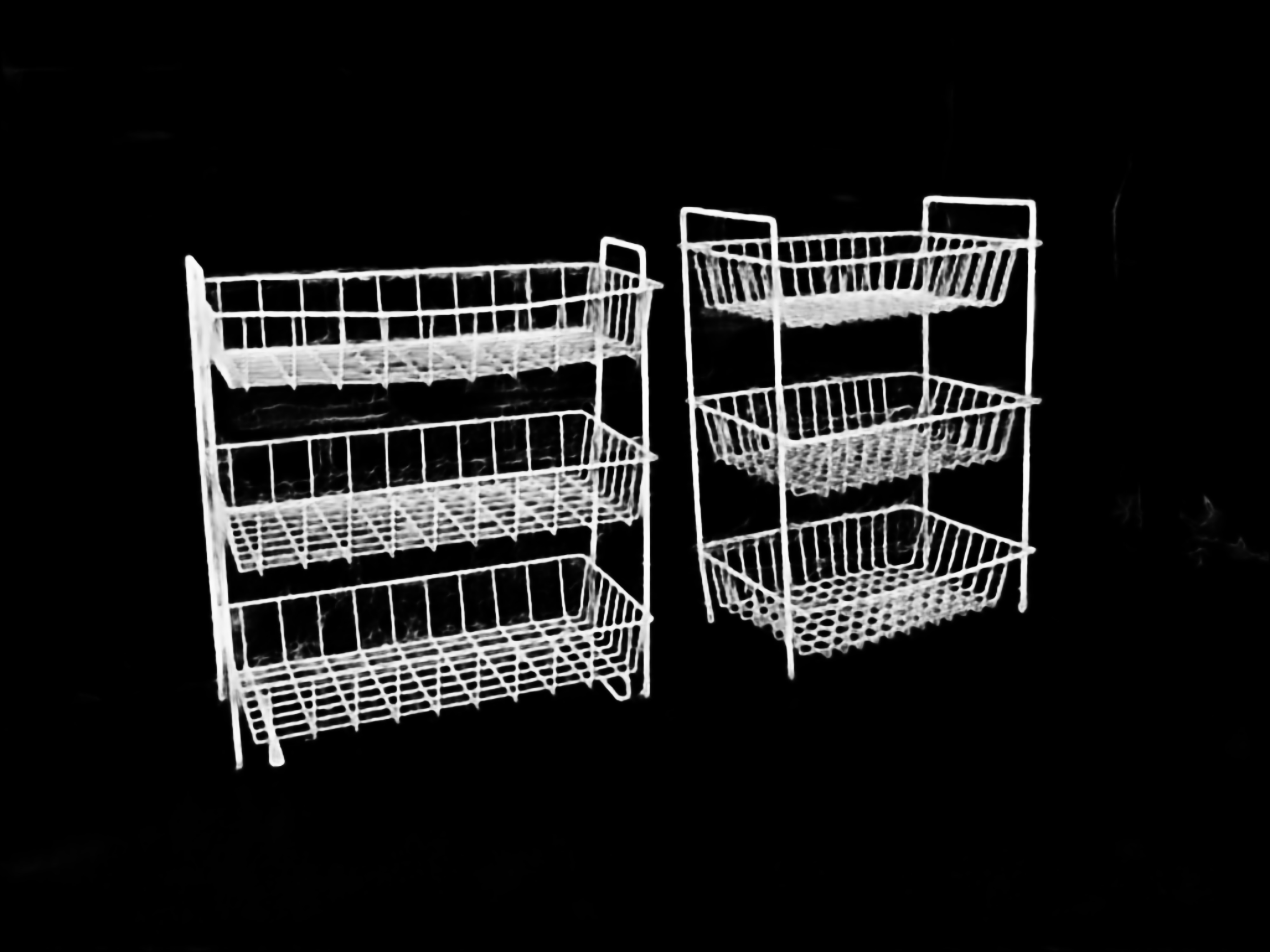}
    \caption*{SAMA}
  \end{subfigure}

  \caption{Comparison of segmentation results.}
  \label{fig:seg_comparison}
  
\end{figure}

From Figure~\ref{fig:seg_comparison},  we show the visual results comparison between SAM, HQ-SAM, and our SAMA, given the same red box prompt. Our SAMA produces significantly more detailed results. For example, SAMA's result for the chair on the first row exhibits clear mesh in the mask. On the fourth row, the thin lines on the gate are also segmented out from the input image. These examples demonstrate SAMA's ability in segmenting detailed features from images.

\begin{figure}[ht]
  \centering
  \begin{subfigure}[b]{0.09\textwidth}
    \includegraphics[width=\linewidth]{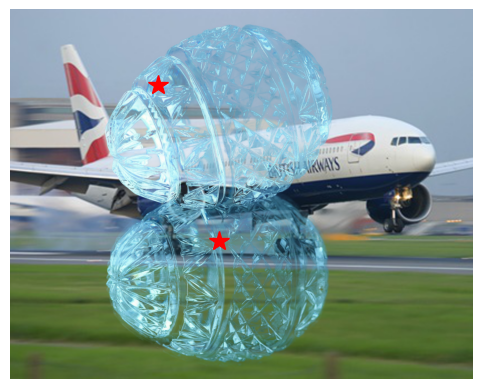}
  \end{subfigure}
  \begin{subfigure}[b]{0.09\textwidth}
    \includegraphics[width=\linewidth]{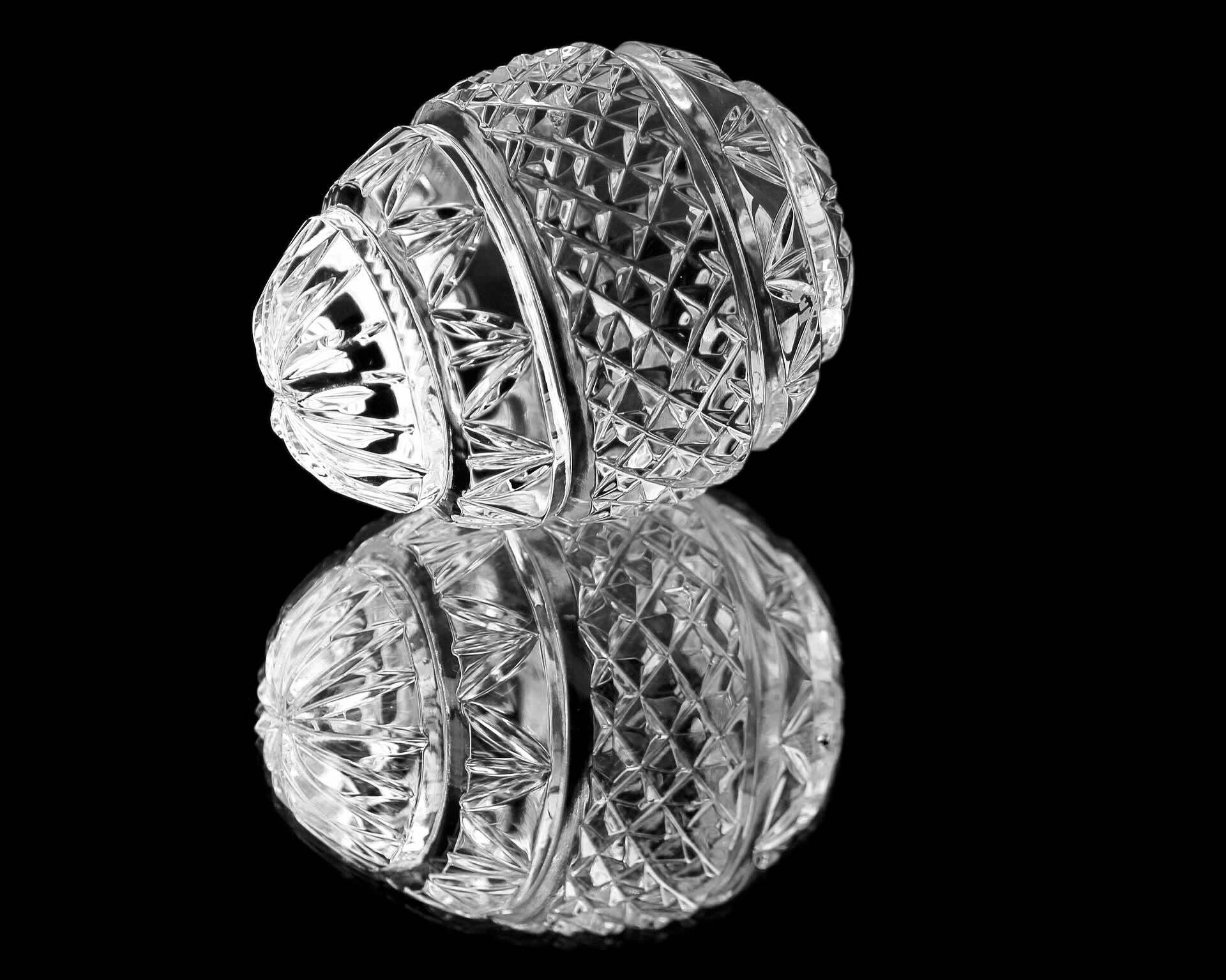}
  \end{subfigure}
  \begin{subfigure}[b]{0.09\textwidth}
    \includegraphics[width=\linewidth]{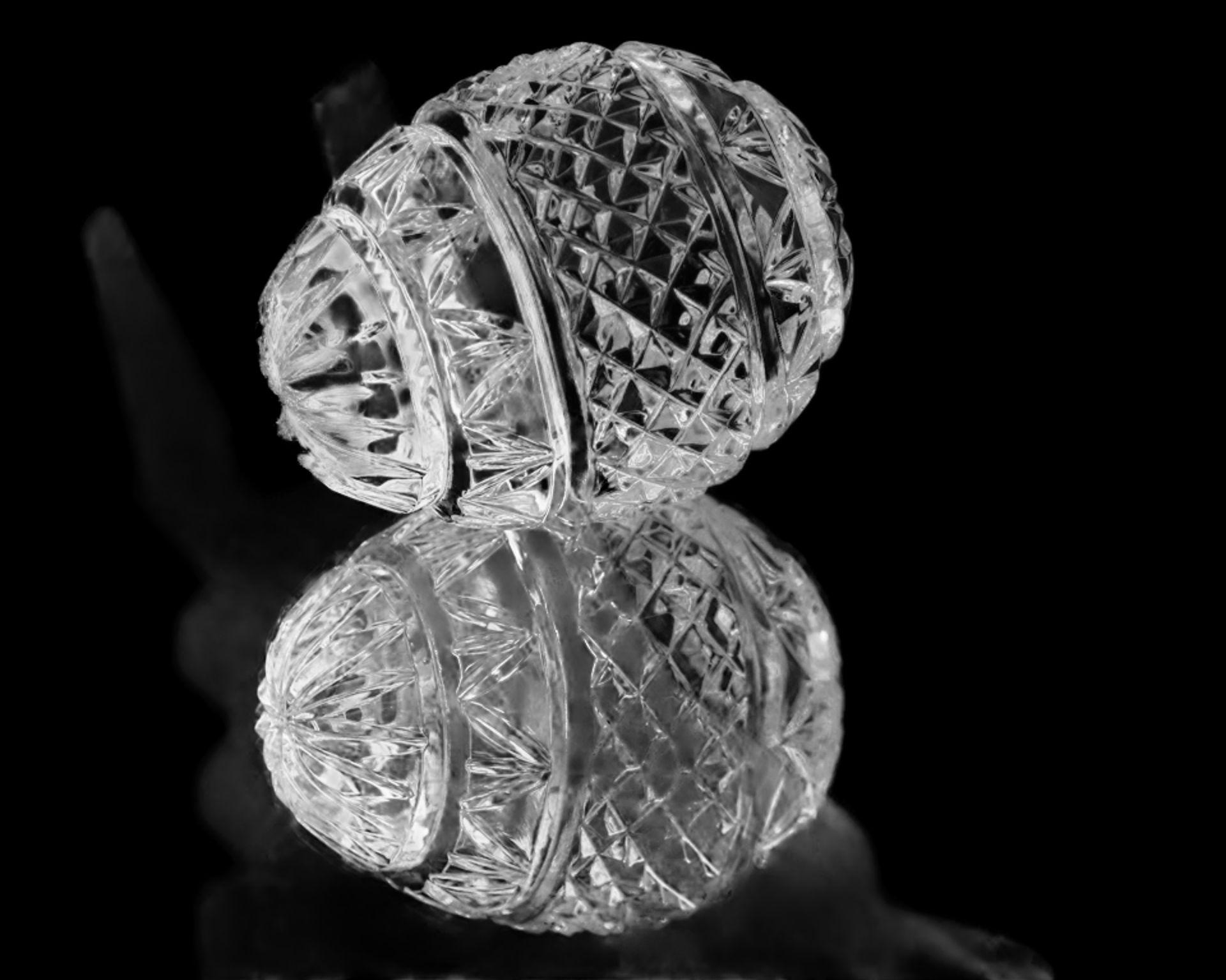}
  \end{subfigure}
  \begin{subfigure}[b]{0.09\textwidth}
    \includegraphics[width=\linewidth]{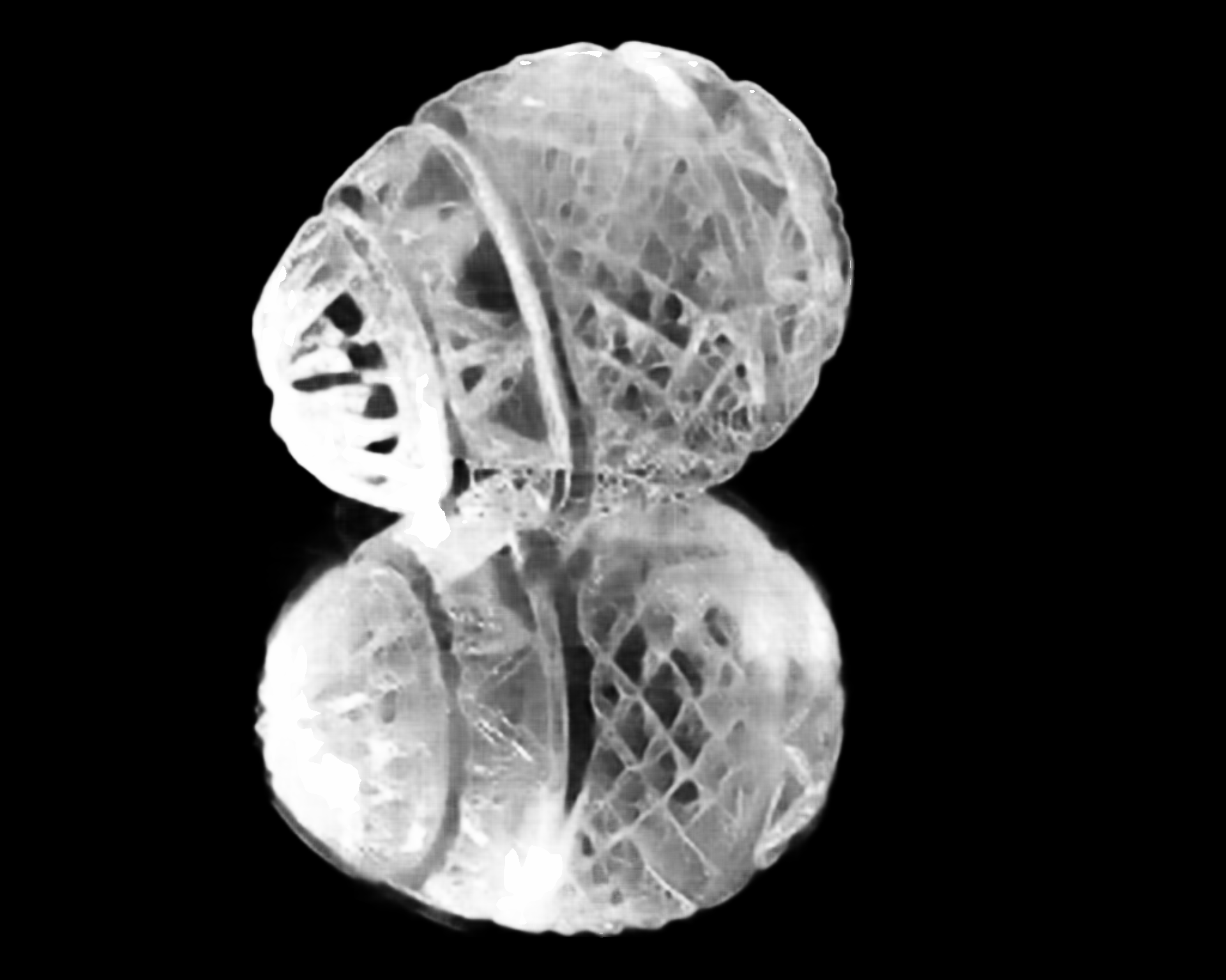}
  \end{subfigure}
  \begin{subfigure}[b]{0.09\textwidth}
    \includegraphics[width=\linewidth]{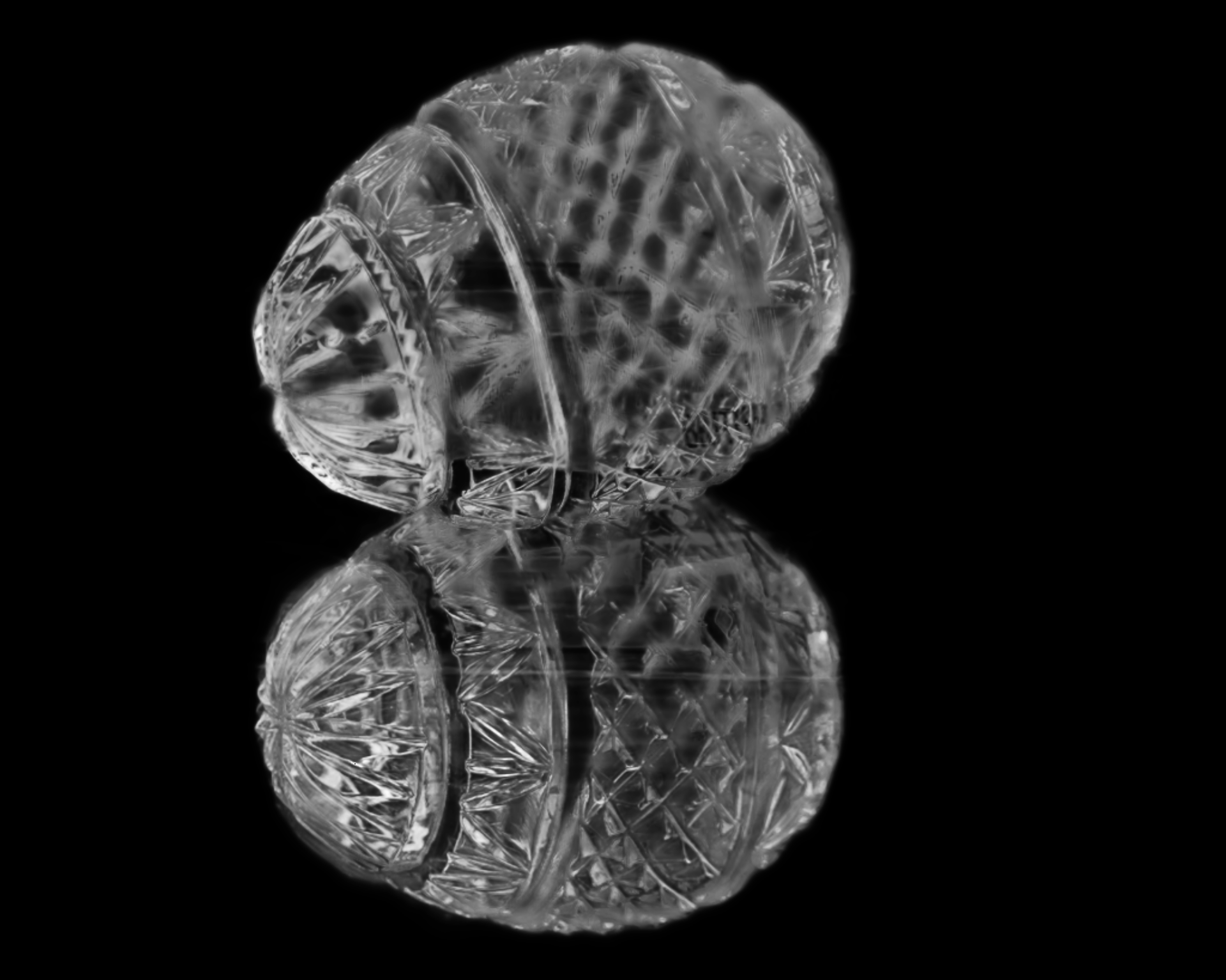}
  \end{subfigure}

  \begin{subfigure}[b]{0.09\textwidth}
    \includegraphics[width=\linewidth]{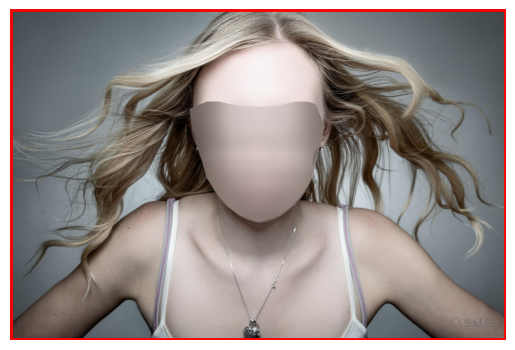}
  \end{subfigure}
  \begin{subfigure}[b]{0.09\textwidth}
    \includegraphics[width=\linewidth]{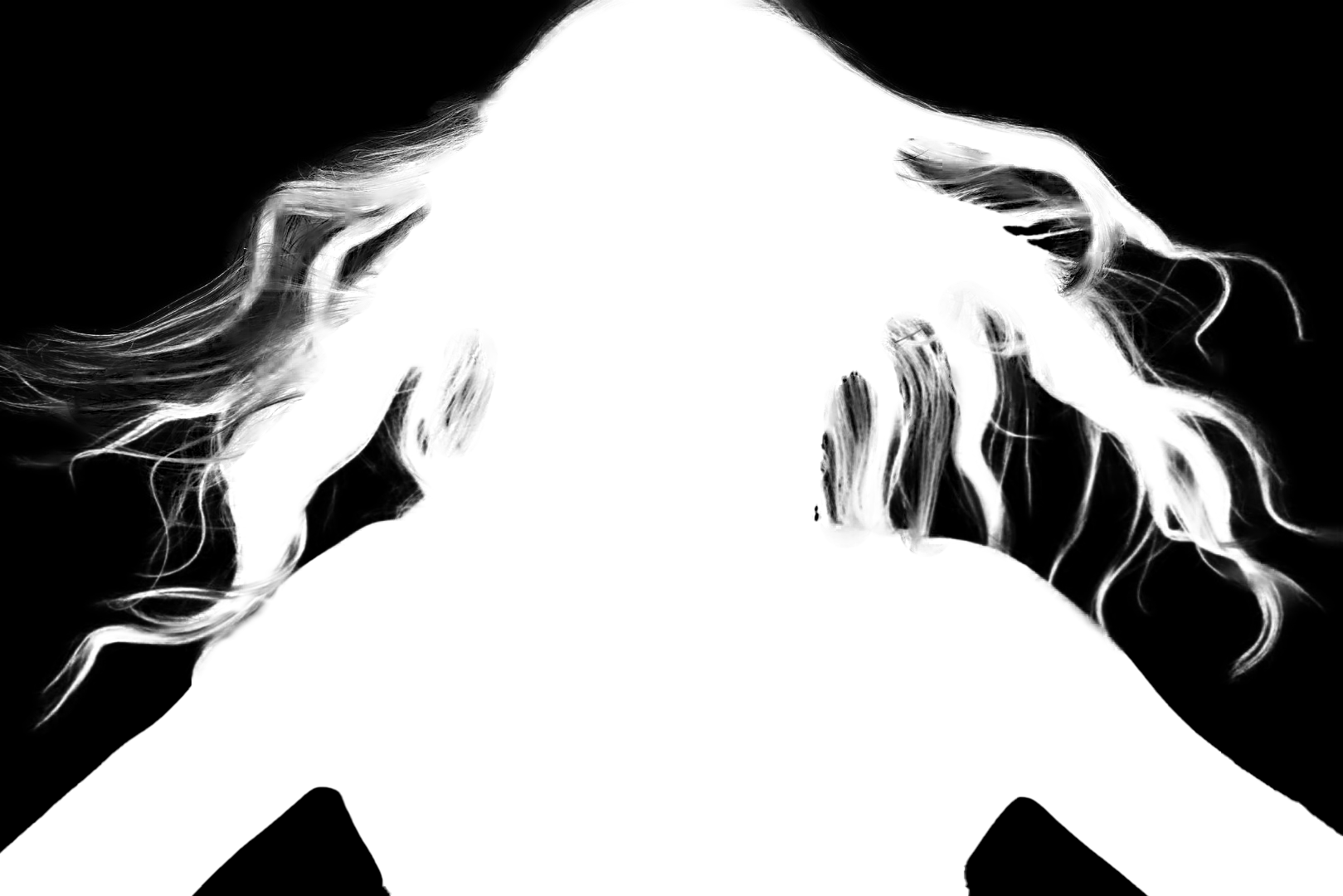}
  \end{subfigure}
  \begin{subfigure}[b]{0.09\textwidth}
    \includegraphics[width=\linewidth]{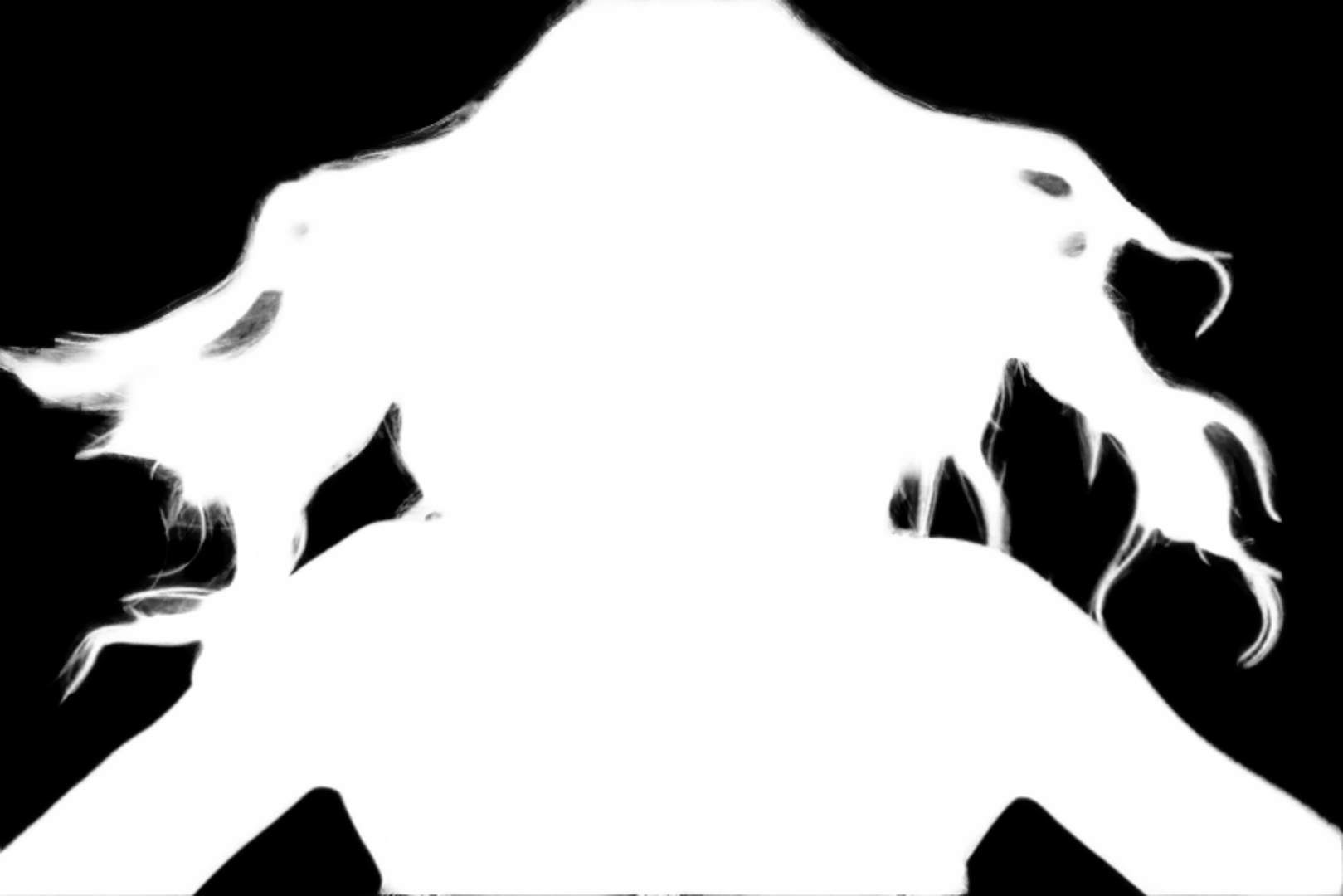}
  \end{subfigure}
  \begin{subfigure}[b]{0.09\textwidth}
    \includegraphics[width=\linewidth]{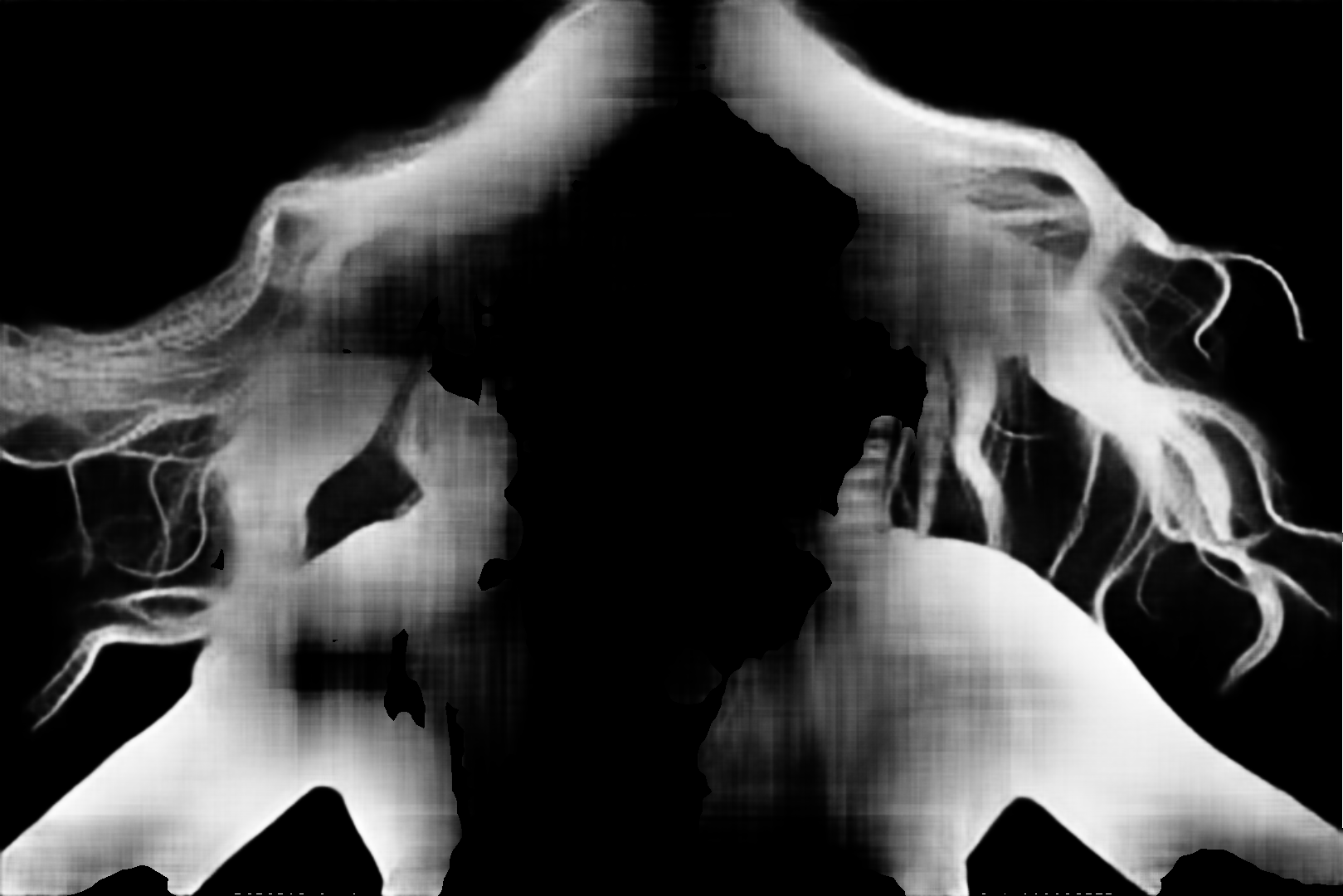}
  \end{subfigure}
  \begin{subfigure}[b]{0.09\textwidth}
    \includegraphics[width=\linewidth]{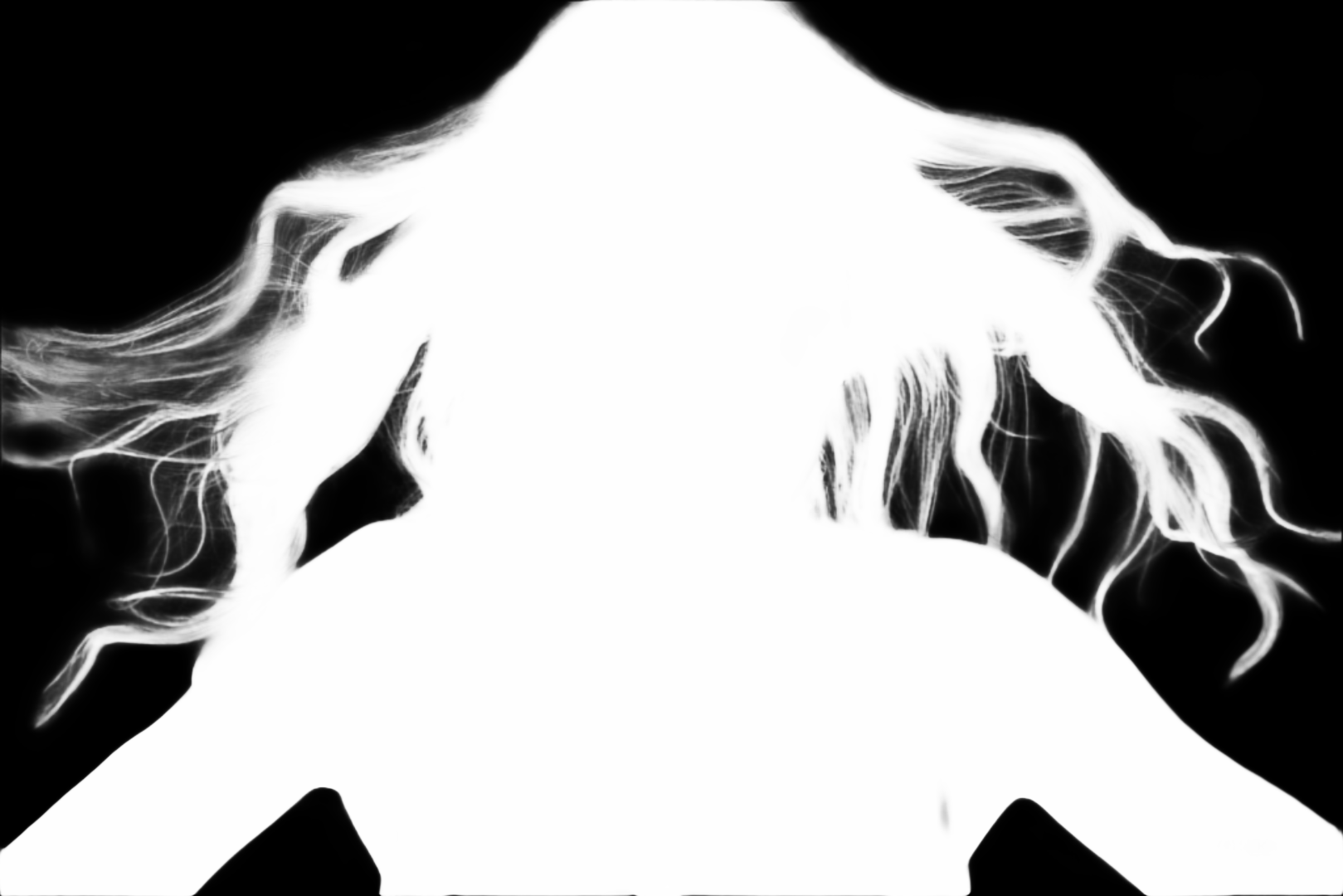}
  \end{subfigure}

  \begin{subfigure}[b]{0.09\textwidth}
    \includegraphics[width=\linewidth]{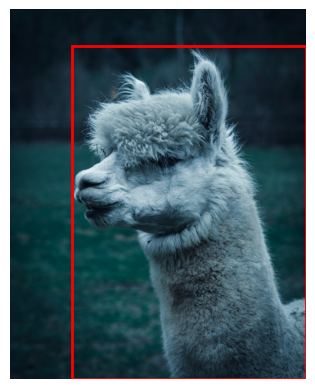}
  \end{subfigure}
  \begin{subfigure}[b]{0.09\textwidth}
    \includegraphics[width=\linewidth]{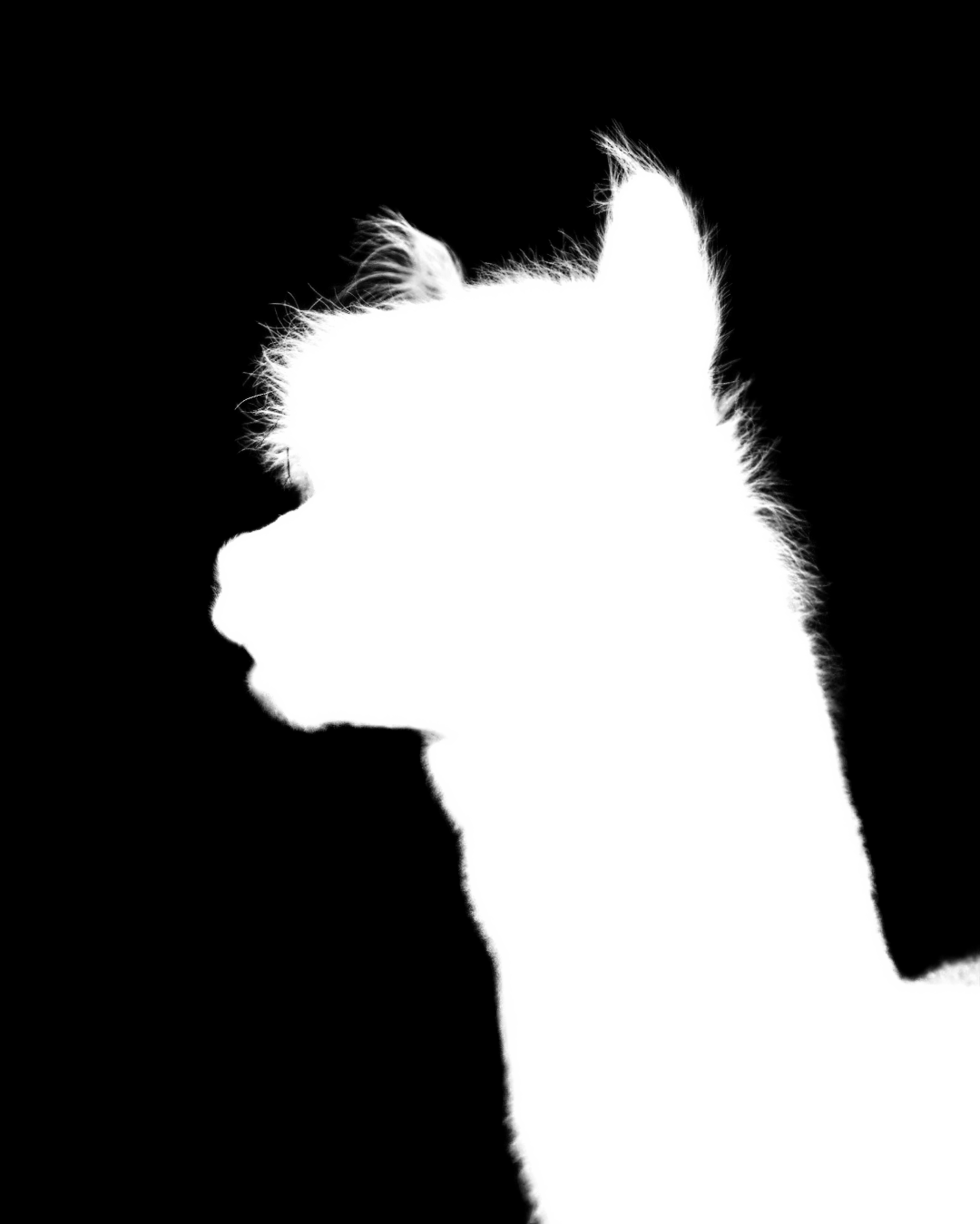}
  \end{subfigure}
  \begin{subfigure}[b]{0.09\textwidth}
    \includegraphics[width=\linewidth]{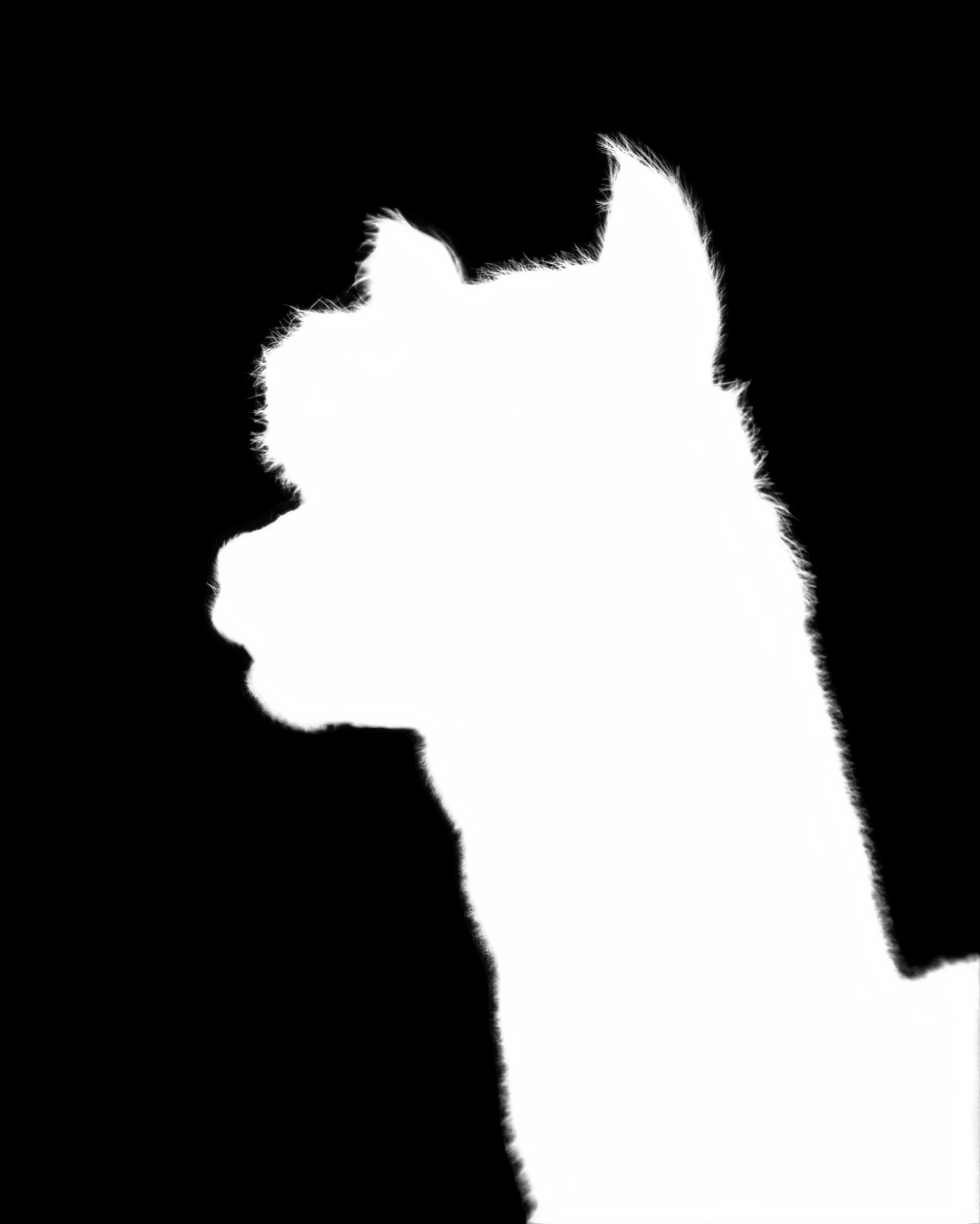}
  \end{subfigure}
  \begin{subfigure}[b]{0.09\textwidth}
    \includegraphics[width=\linewidth]{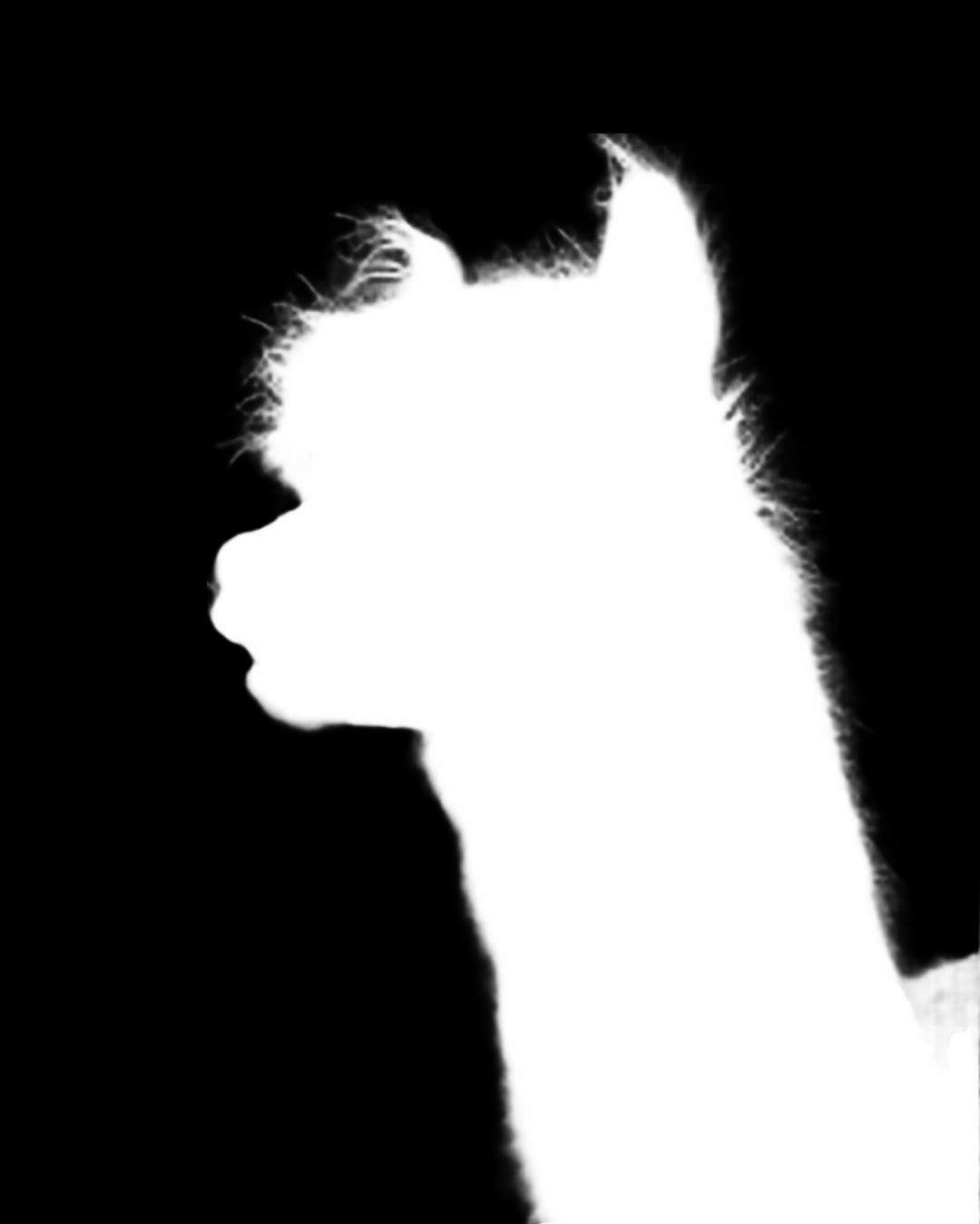}
  \end{subfigure}
  \begin{subfigure}[b]{0.09\textwidth}
    \includegraphics[width=\linewidth]{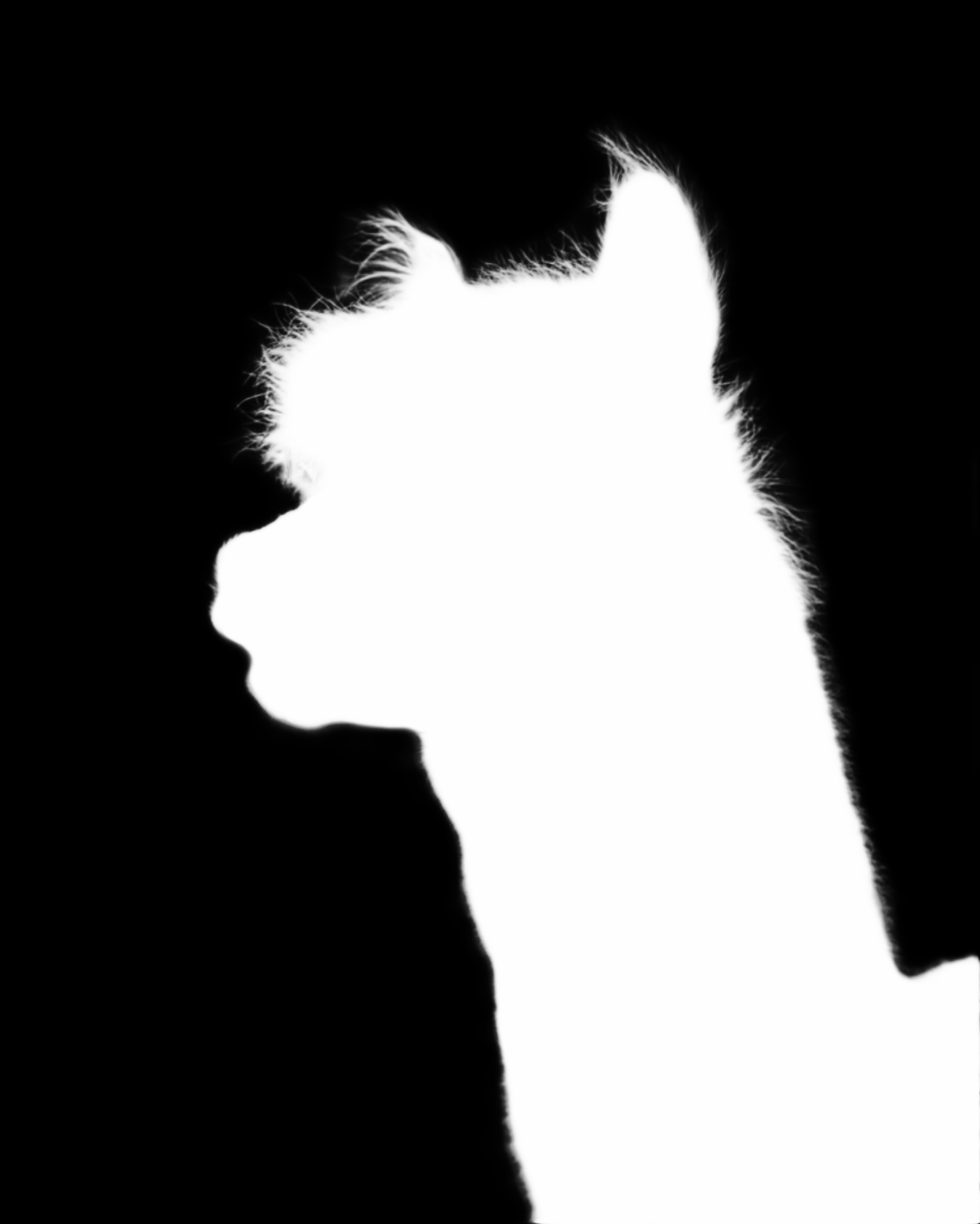}
  \end{subfigure}

  \begin{subfigure}[b]{0.09\textwidth}
    \includegraphics[width=\linewidth]{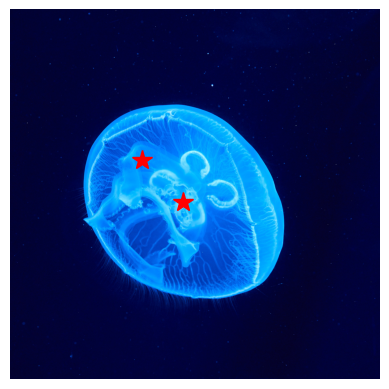}
  \end{subfigure}
  \begin{subfigure}[b]{0.09\textwidth}
    \includegraphics[width=\linewidth]{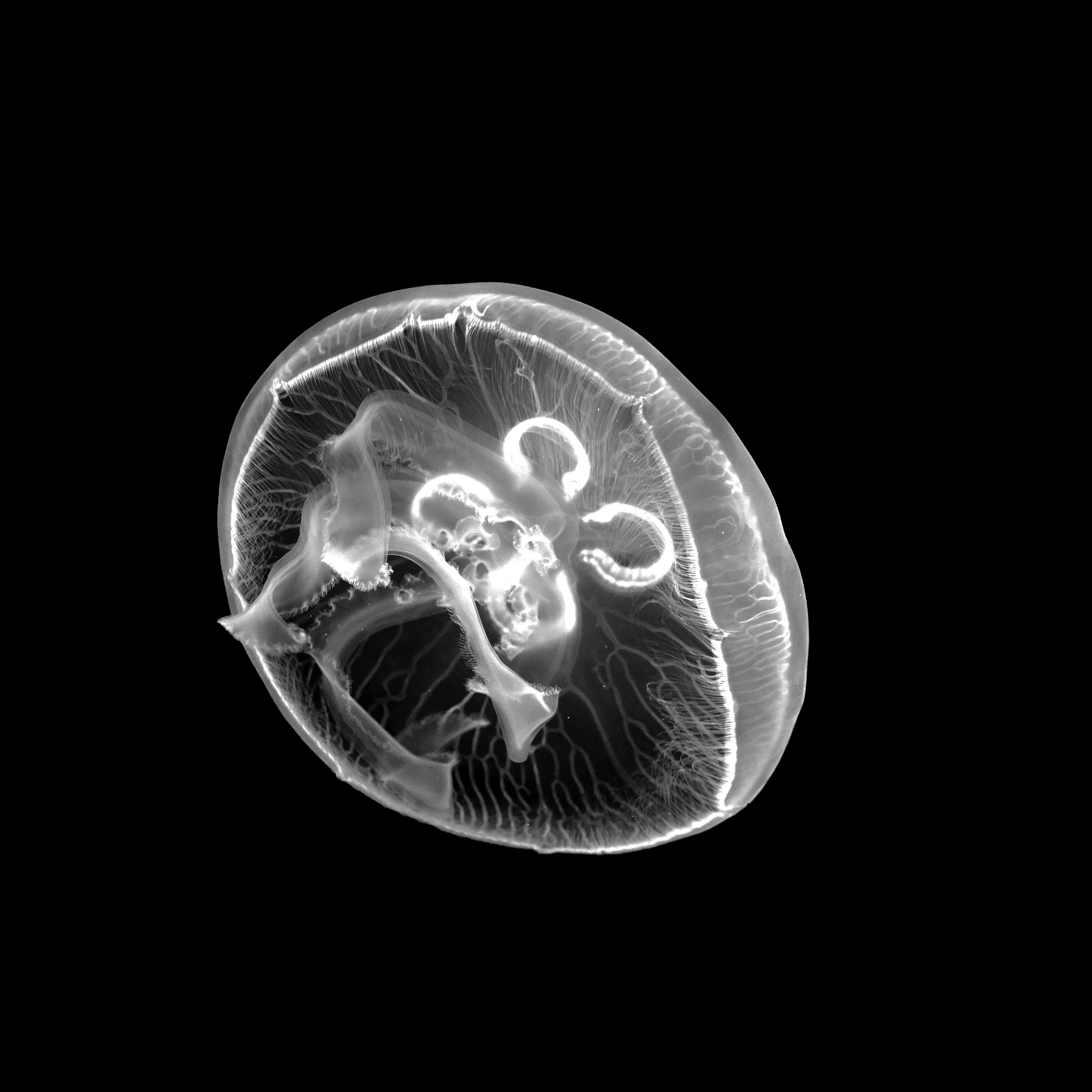}
  \end{subfigure}
  \begin{subfigure}[b]{0.09\textwidth}
    \includegraphics[width=\linewidth]{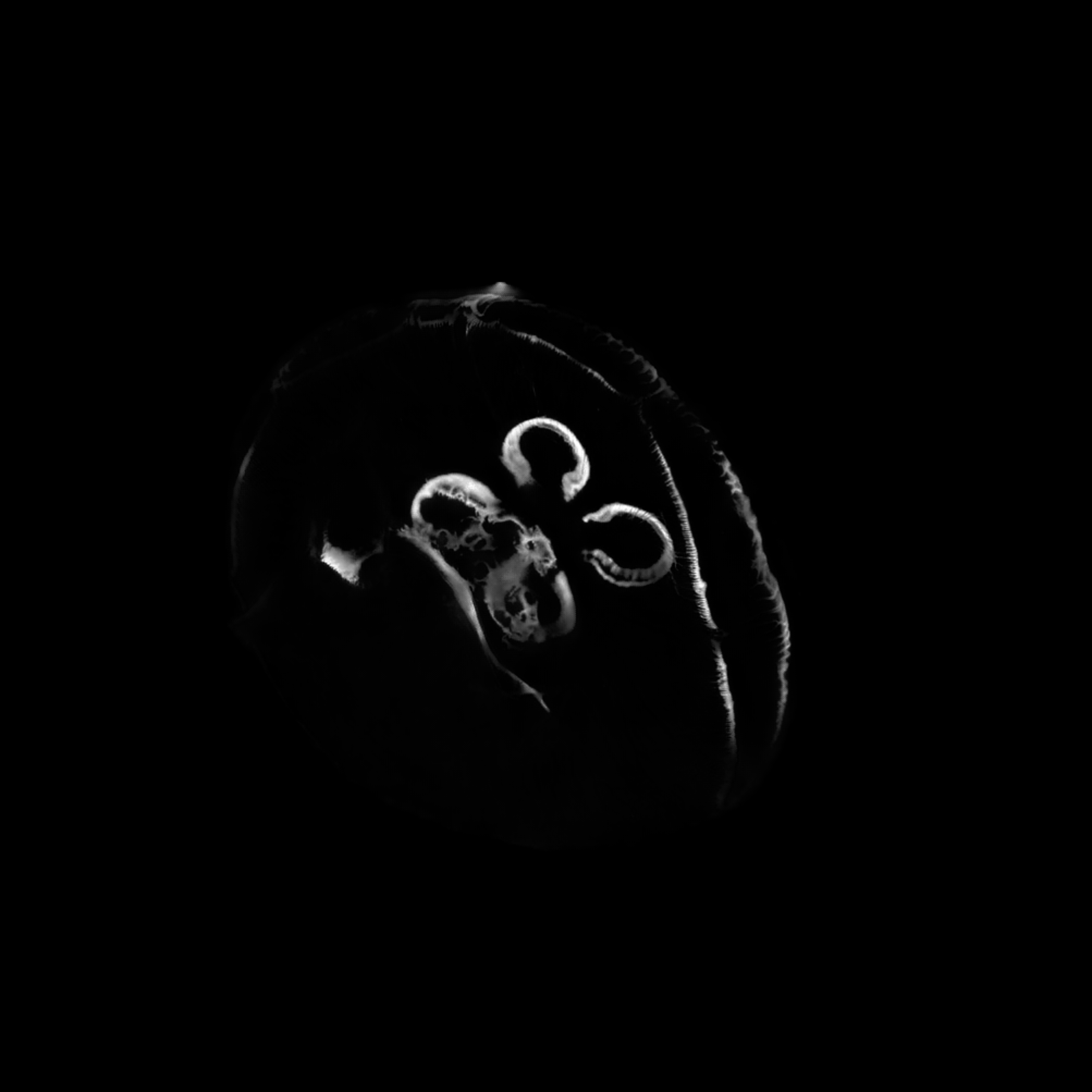}
  \end{subfigure}
  \begin{subfigure}[b]{0.09\textwidth}
    \includegraphics[width=\linewidth]{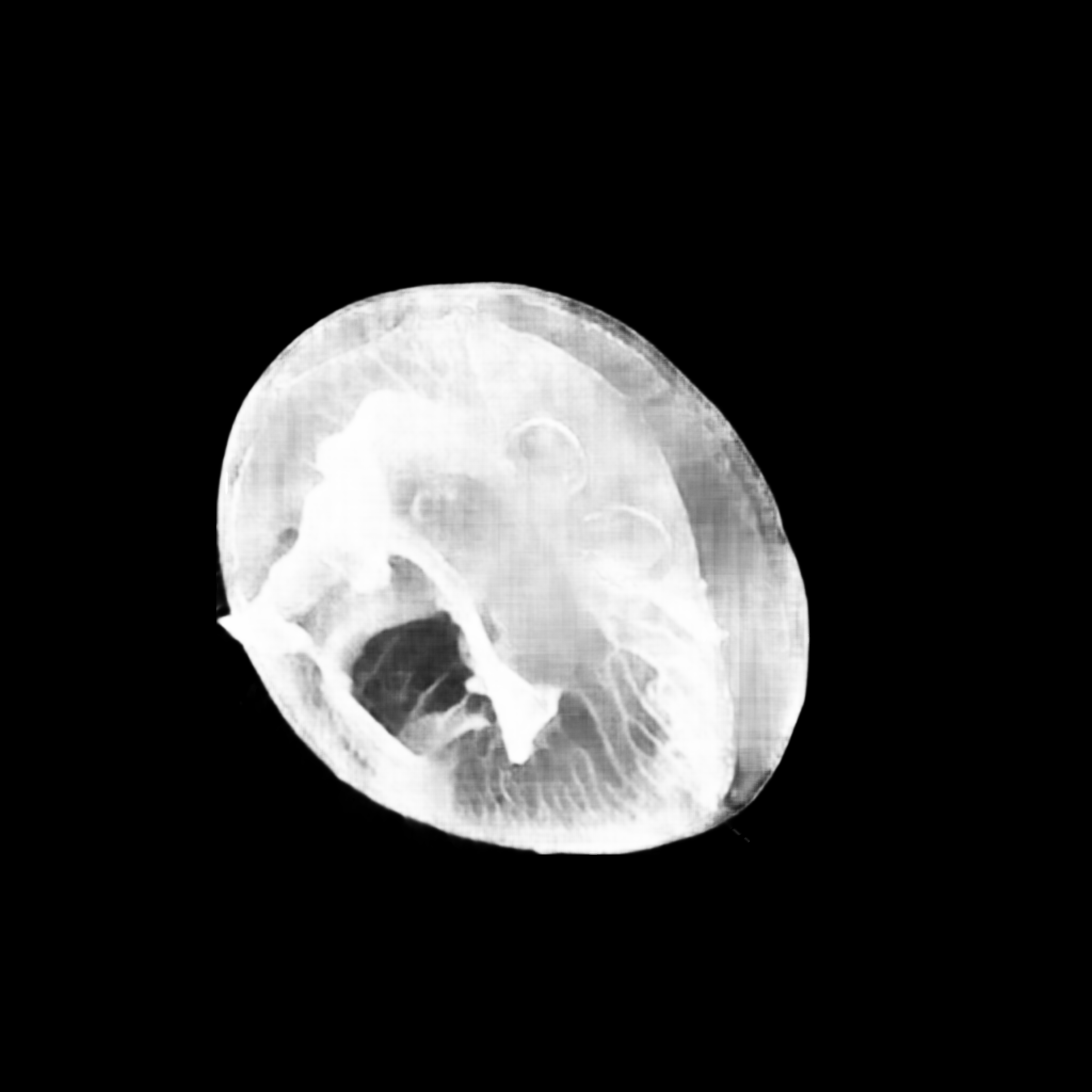}
  \end{subfigure}
  \begin{subfigure}[b]{0.09\textwidth}
    \includegraphics[width=\linewidth]{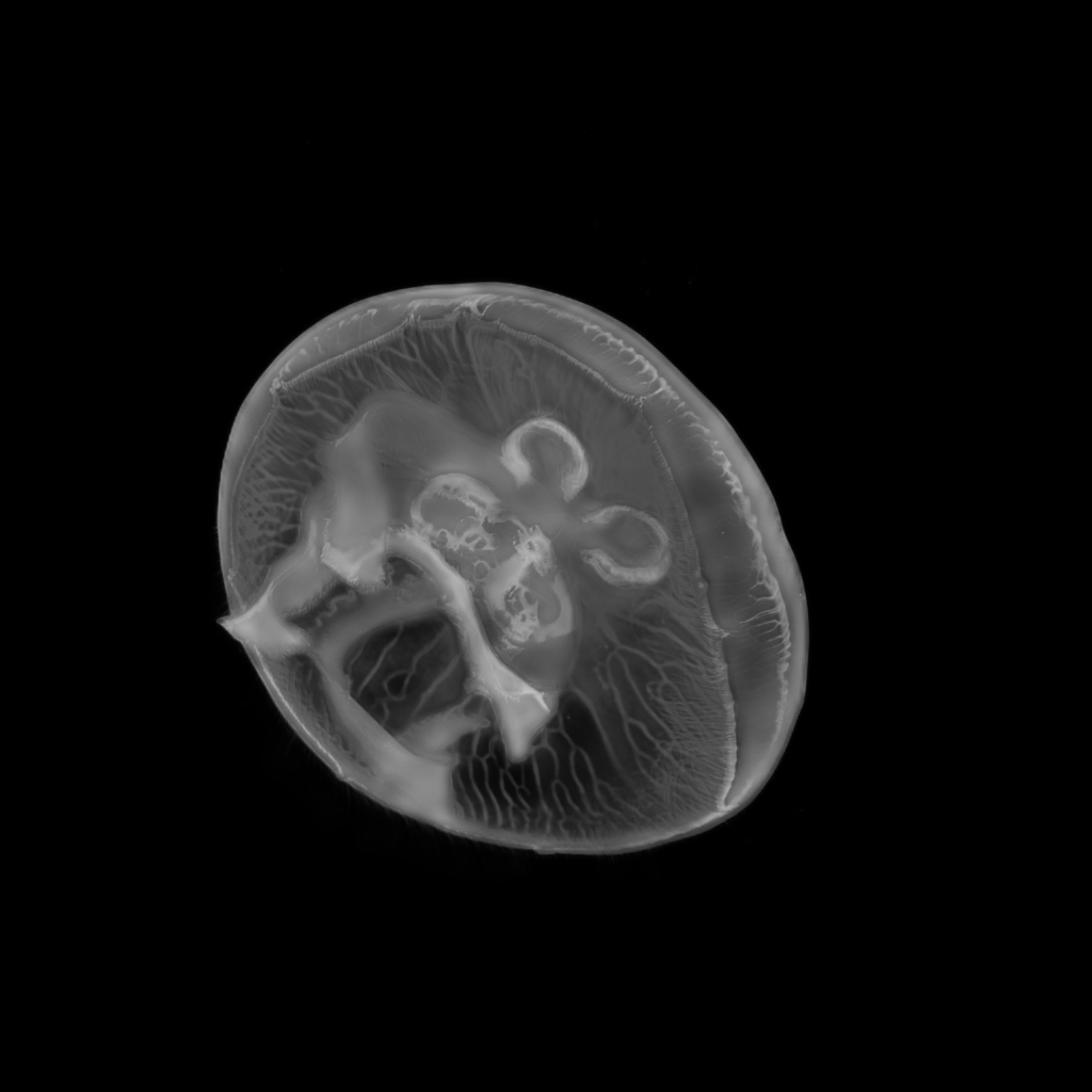}
  \end{subfigure}

  \begin{subfigure}[b]{0.09\textwidth}
    \includegraphics[width=\linewidth]{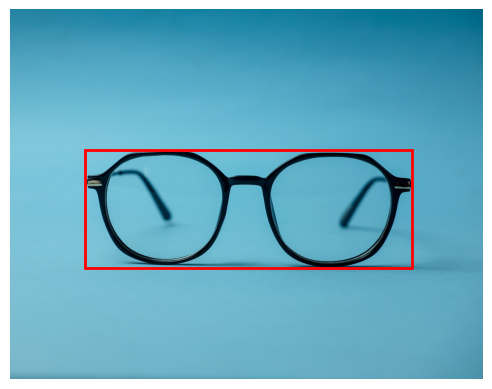}
    \caption*{Image}
  \end{subfigure}
  \begin{subfigure}[b]{0.09\textwidth}
    \includegraphics[width=\linewidth]{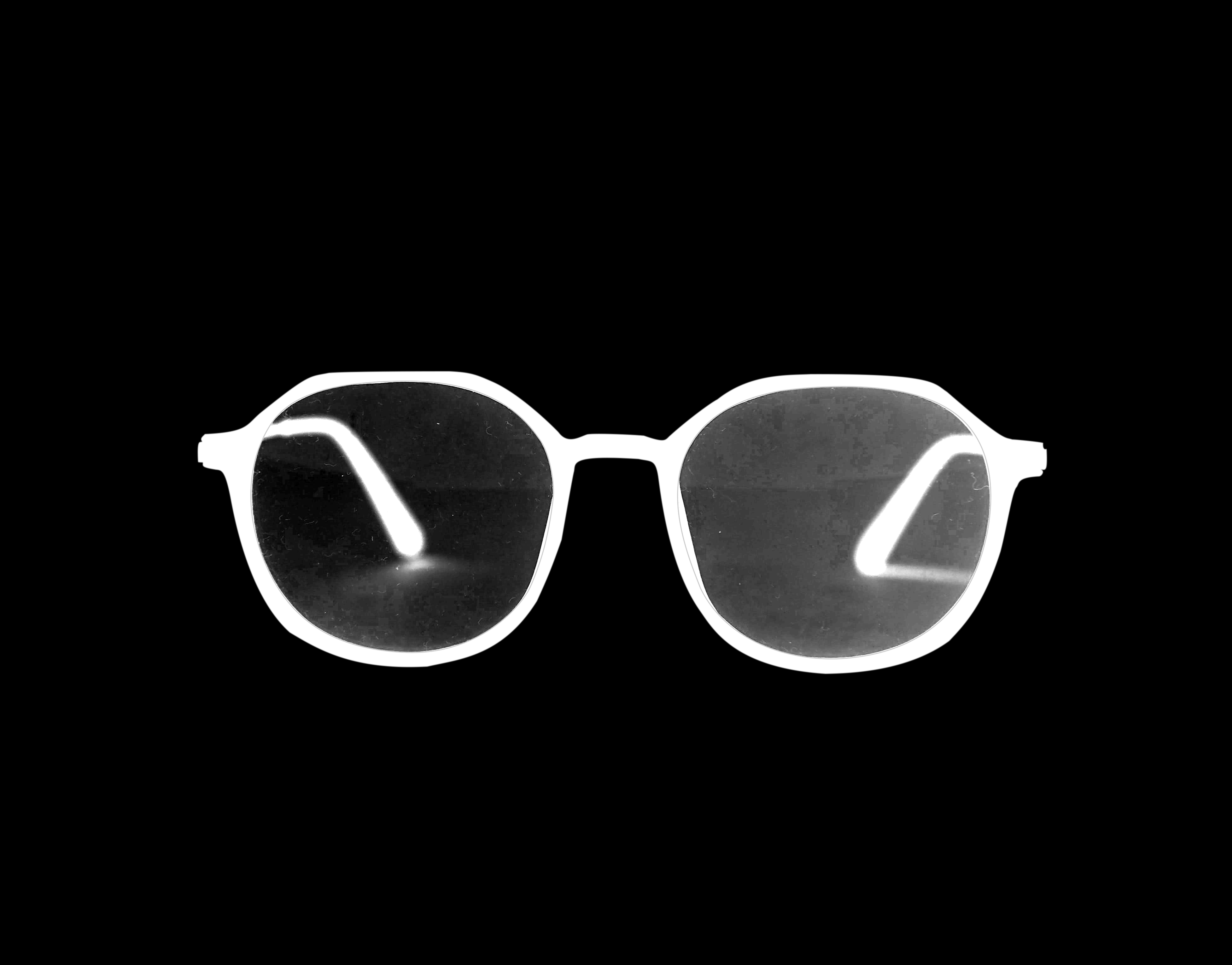}
    \caption*{GT}
  \end{subfigure}
  \begin{subfigure}[b]{0.09\textwidth}
    \includegraphics[width=\linewidth]{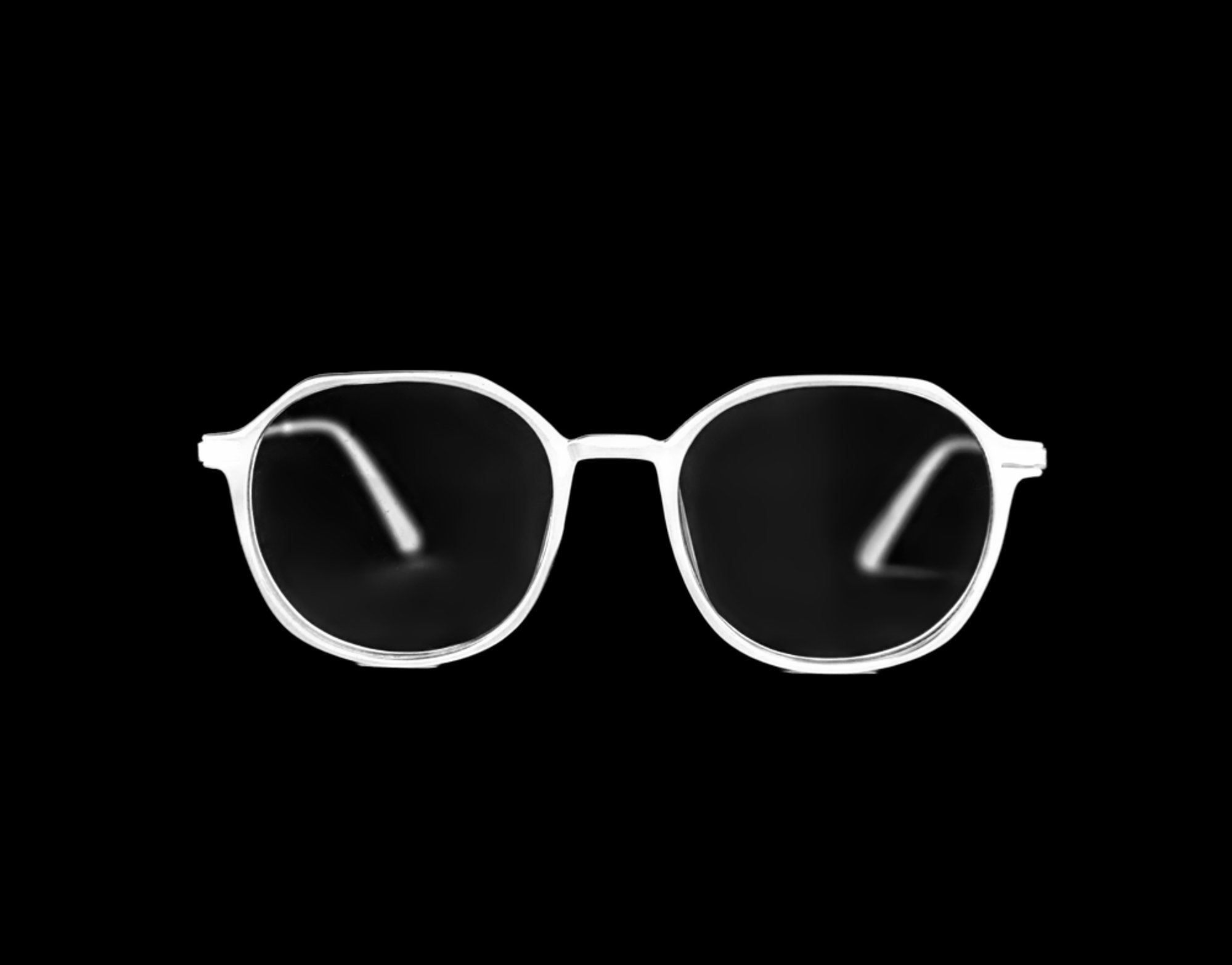}
    \caption*{MatAny}
  \end{subfigure}
  \begin{subfigure}[b]{0.09\textwidth}
    \includegraphics[width=\linewidth]{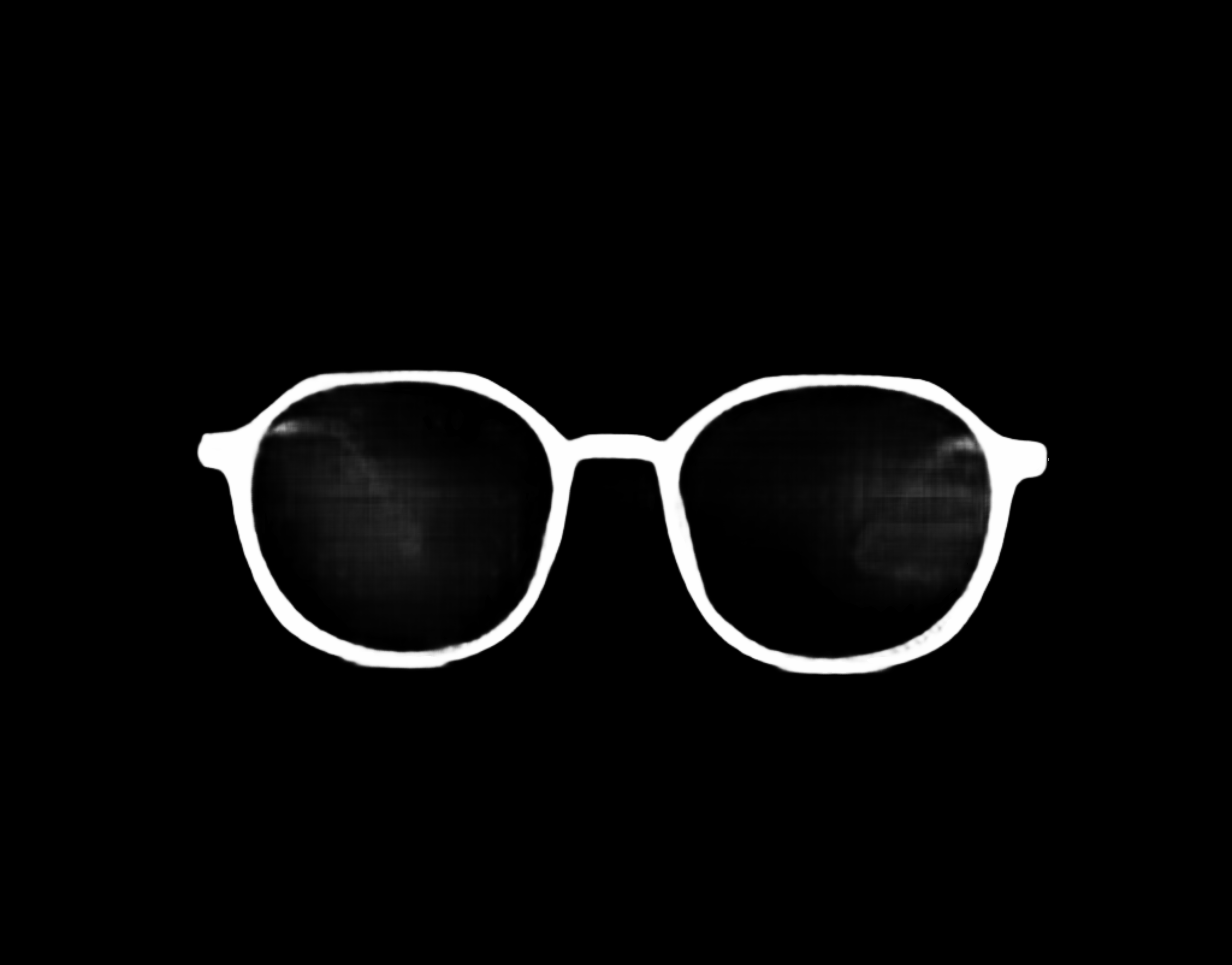}
    \caption*{MAM}
  \end{subfigure}
  \begin{subfigure}[b]{0.09\textwidth}
    \includegraphics[width=\linewidth]{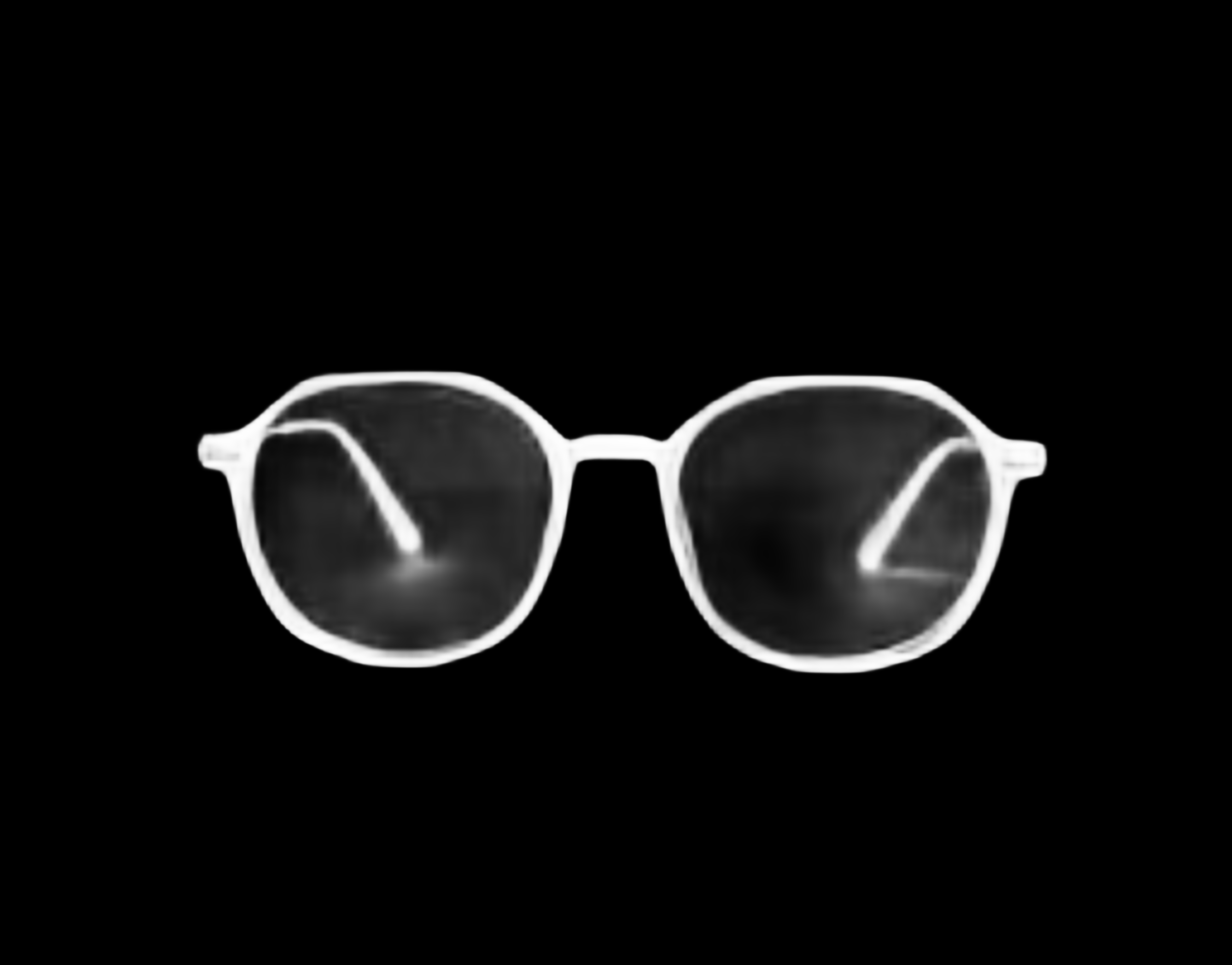}
    \caption*{SAMA}
  \end{subfigure}

  \caption{Comparison of matting results.}
  \label{fig:matting_comparison}
\end{figure}

Additionally, Figure~\ref{fig:matting_comparison} shows the visual results comparison between MatAny, MAM and our SAMA. SAMA also presents visible improvements on the matting task. For instance, the matting mask from SAMA on the second row clearly shows details of the woman's hair. Besides, on the fifth row, the transparency of the glasses is also obvious on SAMA's output mask. These results indicate that SAMA can handle the transparency and hair/fur in the matting mask.


\subsection{Point Prompts Effect Comparison}

\begin{figure}[htbp]
  \centering
  \vspace{-0.5em}

  \includegraphics[width=\linewidth]{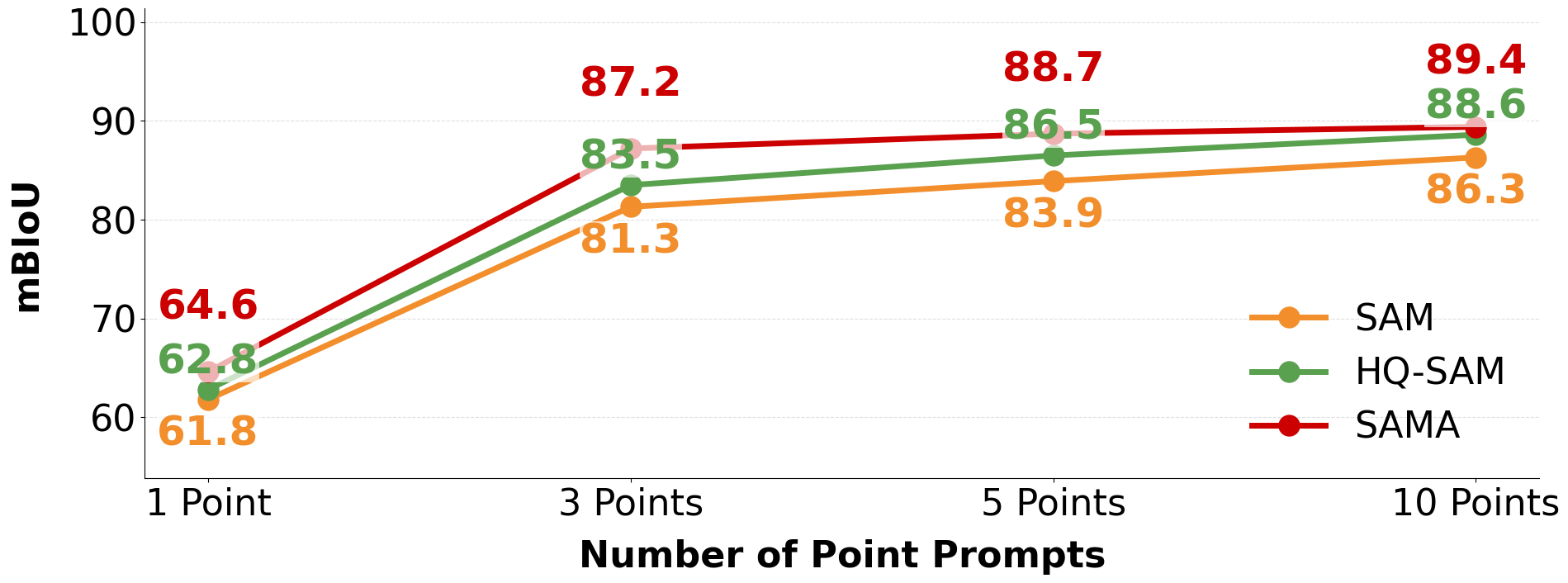}

  \vspace{-0.2em}

  \includegraphics[width=\linewidth]{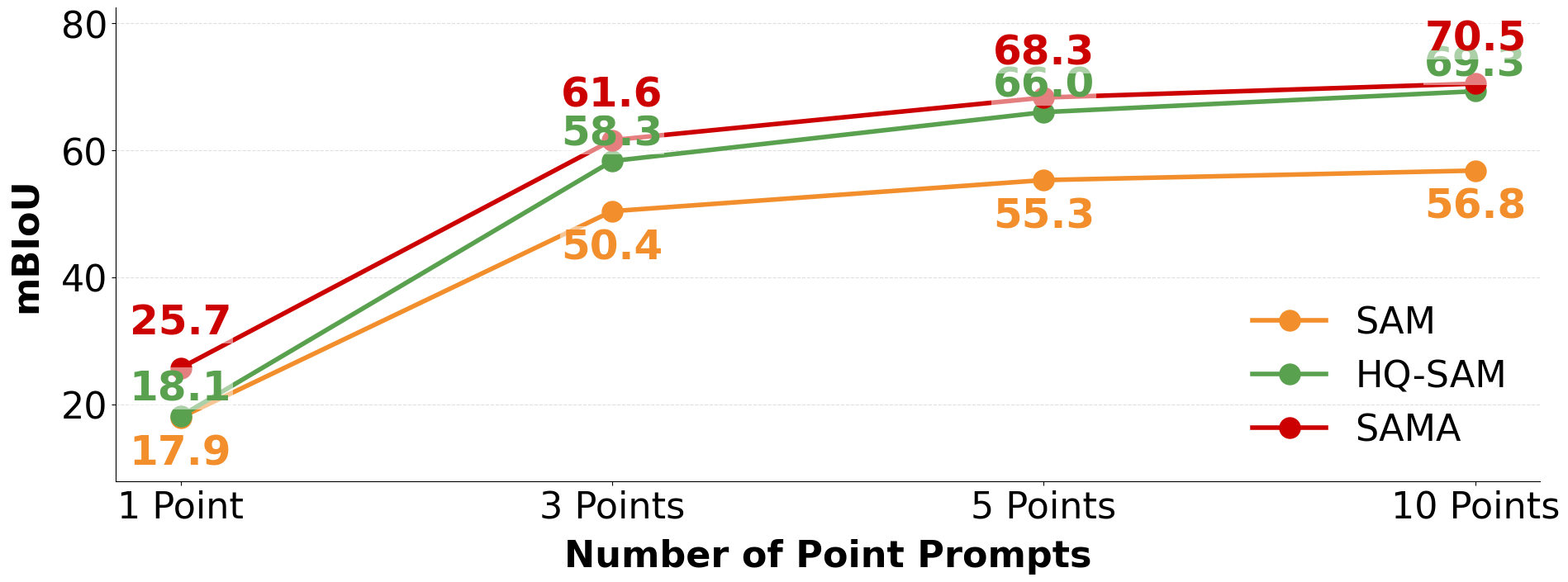}

  \vspace{-0.5em}
  \caption{Results of Interactive Segmentation with varying point prompts}
  \label{fig:two_images}
\end{figure}

Followed~\cite{ke2023segment}, to analyze how interactive point prompts affect the performance of segmentation, we evaluate SAMA with varying numbers of input points on COIFT~\cite{liew2021deep}, which is a zero-shot interactive segmentation dataset about thin objects. As illustrated in Figure~\ref{fig:two_images}, SAMA consistently achieves higher mean Intersection over Union (mIoU) scores than both HQ-SAM and the original SAM across all prompt configurations. Notably, in comparison to HQ-SAM, which is also trained on the DIS-5K dataset~\cite{qin2022highly}—our SAMA exhibits more significant improvements on COIFT under zero-shot conditions, particularly when fewer point prompts (1, 3, or 5) are provided. These results highlight the superior generalization ability of SAMA in interactive segmentation scenarios with limited user input.


\begin{table}[htbp]  
  \centering
  \small
  \setlength{\tabcolsep}{4pt}
  \renewcommand{\arraystretch}{1.15}

  \begin{tabular}{cc|ccccc}
    \toprule
    \textbf{MVLE} & \textbf{L-A} &
      $\mathbf{F_{\!\beta}^{\text{max}}\!\uparrow}$ &
      $\mathbf{F_{\!\beta}^{w}\!\uparrow}$ &
      $\mathbf{M\!\downarrow}$ &
      $\mathbf{S_{\alpha}\!\uparrow}$ &
      $\mathbf{E_{\!\phi}^{m}\!\uparrow}$ \\
    \midrule
    --            & --             & 0.872 & 0.849 & 0.038 & 0.868 & 0.932  \\
    --            & \checkmark     & 0.893 & 0.876 & 0.029 & 0.912 & 0.944 \\
    \checkmark    & --             & 0.882 & 0.869 & 0.027 & 0.903 & 0.942  \\
    \checkmark    & \checkmark     & \textbf{0.942} & \textbf{0.885} & \textbf{0.021} & \textbf{0.930} & \textbf{0.962} \\
    \bottomrule
  \end{tabular}
  \caption{Ablation study on MVLE and Local-Adapter (L-A) on segmentation dataset DIS‑VD. Higher$\uparrow$ is better except for $M$, where lower$\downarrow$ is better.}
  \label{tab:ablation_two_ds}
\end{table}

\begin{table}[htbp]
\centering
\begin{adjustbox}{max width=0.8\linewidth}
\setlength{\tabcolsep}{5pt} 
\renewcommand{\arraystretch}{1}
\resizebox{\columnwidth}{!}{%
\begin{tabular}{cc|cc|cc}
\toprule
\textbf{MVLE} & \textbf{L-A} & \multicolumn{2}{c|}{\textbf{AM2K}} & \multicolumn{2}{c}{\textbf{P3M-500}} \\
\cmidrule(lr){3-4} \cmidrule(lr){5-6}
 & & \textbf{SAD}\,$\downarrow$ & \textbf{MSE}\,$\downarrow$ & \textbf{SAD}\,$\downarrow$ & \textbf{MSE}\,$\downarrow$ \\
\midrule
-- & --           & 19.79 & 0.028  & 16.06 & 0.0346 \\
-- & \checkmark   & 12.74 & 0.007  & 12.15 & 0.0057 \\
\checkmark & --      & 13.12 & 0.011  & 13.09 & 0.0063 \\
\checkmark & \checkmark & \textbf{8.04} & \textbf{0.003} & \textbf{9.08} & \textbf{0.0028} \\
\bottomrule
\end{tabular}%
}
\end{adjustbox}
\caption{Ablation study on MVLE and Local-Adapter (L-A) on matting datasets AM2K and P3M-500. Lower values indicate better performance for SAD and MSE.}
\label{comparematting}
\end{table}

\begin{table}[htbp]
  \centering
  \scriptsize
  \renewcommand{\arraystretch}{1}
  \setlength{\tabcolsep}{5pt}  

  \resizebox{\columnwidth}{!}{  
    \begin{tabular}{cc ccccc cc}
      \toprule
      & & \multicolumn{5}{c}{\textbf{DIS‑VD}} & \multicolumn{2}{c}{\textbf{RefMatte-RW100}} \\
      \cmidrule(lr){3-7} \cmidrule(lr){8-9}
      Seg & Matting &
        $F_{\!\beta}^{\text{max}}\!\uparrow$ &
        $F_{\!\beta}^{w}\!\uparrow$ &
        $M\!\downarrow$ &
        $S_{\alpha}\!\uparrow$ &
        $E_{\!\phi}^{m}\!\uparrow$ &
        \textbf{SAD}\,$\downarrow$ & 
        \textbf{MSE}\,$\downarrow$ \\
      \midrule
      \checkmark & --  & 0.917 & 0.876 & 0.027 & 0.916 & 0.946 & 62.70 & 0.054 \\
      -- & \checkmark  & 0.682 & 0.709 & 0.071 & 0.807 & 0.855 & 34.25 & 0.021 \\
      \checkmark & \checkmark & \textbf{0.942} & \textbf{0.885} & \textbf{0.021} & \textbf{0.930} & \textbf{0.962} & \textbf{25.69} & \textbf{0.0100} \\
      \bottomrule
    \end{tabular}
  }
  \caption{Multi-task learning study on datasets DIS‑VD and RefMatte-RW100. Higher$\uparrow$ is better except for $M$, where lower$\downarrow$ is better.}
\label{multi_task}
\end{table}

\subsection{Ablation Study}

\textbf{Ablation on MVLE and Local-Adapter Module}
We conduct ablation experiments on our SAMA to evaluate the effectiveness of our proposed modules in SAMA. Specifically, for the segmentation task, we use two segmentation datasets:  DIS-VD for highly accurate segmentation and COIFT for interactive segmentation to conduct our experiments. We show our results in Table~\ref{tab:ablation_two_ds}.

From Table~\ref{tab:ablation_two_ds}, we find that the MVLE module boosts performance across all metrics on the baseline model, showing its role in offering complementary information from precise local features for fine-grained segmentation. Besides, Local-Adapter improves significantly on DIS-VD including more complex structures in images, which means it is useful to provide information about local details of objects. 
We also analyze the contribution of MVLE and Local-Adapter to the matting task. We adopt the same ablation protocol as used in the segmentation task to evaluate the contributions of MVLE and LA to matting performance. As reported in Table~\ref{comparematting}, the exclusion of either MVLE or LA leads to a substantial decline in matting accuracy, highlighting the critical role of both components. These findings demonstrate that the integration of MVLE and LA significantly enhances the model’s ability to capture fine-grained details essential for high-quality alpha matte prediction. 

\textbf{Ablation on Multi-task Learning }
We evaluate the effectiveness of our SAMA framework in jointly training both segmentation and matting tasks. To assess the benefits of multi-task learning, we compare the proposed joint training approach with models trained exclusively on either the segmentation or the matting task. As shown in Table~\ref{multi_task}, models trained jointly outperform those trained on either task alone. Incorporating matting benefits segmentation by providing fine-grained boundary details, while joint training with segmentation significantly boosts matting accuracy even without trimaps. This demonstrates that large-scale interactive segmentation data effectively supports interactive matting, especially when matting annotations are limited.

\section{Conclusion}
We propose SAMA, a lightweight extension of SAM that jointly performs image segmentation and matting. Using a Multi-View Localization Encoder, Local-Adapter, and task-specific prediction heads, SAMA improves performance on both segmentation and matting with minimal overhead. Future work includes runtime optimization and extending SAMA to video segmentation.

\FloatBarrier

\bibliography{aaai2026}

\clearpage  

\section{APPENDIX}

In the supplementary material, we first present additional experiments of SAMA, including extended zero-shot evaluations for both segmentation and matting, as well as efficiency comparisons. We then provide detailed implementation specifications of SAMA. Finally, we discuss the limitations of SAMA in the concluding section.


\section{Zero-shot Segmentation Matting}


In additional experiments, we use COIFT~\cite{liew2021deep}~\footnote{http://www.vision.ime.usp.br/~lucyacm/thesis/coift.html} and DIS5K~\cite{qin2022highly}~\footnote{https://github.com/xuebinqin/DIS} for segmentation, and AM2K~\cite{li2022bridging}, P3M-500~\cite{li2021privacy} and Distinctions-646~\cite{qiao2020attention}, RefMatte-RW100~\cite{li2023referring} for matting.

\subsection{More Zero-shot Segmentation Evaluations}

\subsubsection{Zero-Shot Interactive Segmentation}
Table~\ref{coift} presents a comparative evaluation of the zero-shot segmentation performance of our proposed SAMA against existing interactive segmentation models on the COIFT~\cite{liew2021deep} benchmark. The results demonstrate that SAMA consistently outperforms other SAM-based approaches, exhibiting a clear advantage in accurately interpreting user prompts under the interactive segmentation setting.

\begin{table}[H] 
  \centering
  \small
  \setlength{\tabcolsep}{6pt}
  \renewcommand{\arraystretch}{1.15}
  \caption{Results on the \textbf{COIFT} test set (280 samples).  
           Higher$\uparrow$ is better except for $M$, where lower$\downarrow$ is better.}
  \begin{tabular}{lcccccc}
    \toprule
    Method &
      $F_{\!\beta}^{\text{max}}\!\uparrow$ &
      $F_{\!\beta}^{w}\!\uparrow$ &
      $M \downarrow$ &
      $S_{\alpha}\!\uparrow$ &
      $E_{\!\phi}^{m}\!\uparrow$  \\
    \midrule
    SAM            & .966 & .967 & .007 & .964 & .988  \\
    HQ‑SAM         & .974 & .976 & .005 & .971 & .991  \\
    DIS‑SAM        & .982 & .969 & .005 & .978 & .988  \\
    \textbf{SAMA(ours)}  & \textbf{.990} & \textbf{.982} & \textbf{.004} & \textbf{.984} & \textbf{.993}  \\
    \bottomrule
  \end{tabular}
   \label{coift}
\end{table}

\subsubsection{Zero-Shot Salient Object Segmentation} 
In this section, we evaluate the performance of our proposed SAMA on the Salient Object Detection (SOD) task using the HRSOD benchmark, which focuses on segmenting the most visually prominent object in a scene. We compare SAMA against both SAM-based methods and several established baselines, including LDF ~\cite{wei2020label}, HRSOD ~\cite{zeng2019towards}, PGNet ~\cite{xie2022pyramid}, DHQ ~\cite{tang2021disentangled}, and BiRefNet ~\cite{zheng2024bilateral}. As presented in Table ~\ref{hrsod}, SAMA consistently outperforms all competing methods across multiple evaluation metrics. These results demonstrate the robustness of SAMA in zero-shot settings, particularly in accurately detecting salient objects across diverse object categories in high-resolution imagery.
 
\begin{table*}[t]
  \centering
  \caption{Performance on HRSOD for salient object segmentation (higher is better except $\mathcal{M}$). 
           Bold denotes the best value per row.}
  \begin{tabular}{lccccccccc}
    \toprule
        & LDF & HRSOD & DHQ & BiRefNet & SAM 
        & HQ‑SAM & DIS‑SAM & Pi‑SAM & SAM‑UQ \\
    \midrule
    $S_m \uparrow$
        & .904 & .896 & .920 & .960 & .932 & .958 & .969 & .972 & \textbf{.977} \\
    $F_{\!\beta}^{x} \uparrow$
        & .904 & .905 & .922 & .962 & .955 & .973 & .971 & .974 & \textbf{.986} \\
    $E_{\!\phi}^{m} \uparrow$
        & .919 & .934 & .947 & .979 & .963 & .985 & .984 & \textbf{.991} &  .988 \\
    $\mathcal{M} \downarrow$
        & .032 & .030 & .022 & .011 & .022 & .012 & .008 & .006 & \textbf{.005} \\
    \bottomrule
  \end{tabular}
  \label{hrsod}
\end{table*}

\subsection{More Zero-shot Matting Evaluations}
\subsubsection{Zero-shot Semantic Image Matting}
Table~\ref{semantic} presents the zero-shot performance of SAMA on two semantic image matting benchmarks: AM2K~\cite{li2022bridging}, an animal-specific dataset, and P3M-500~\cite{li2021privacy}, which focuses on human portraits. On both benchmarks, SAMA achieves significant improvements in MSE and SAD metrics compared to existing interactive matting methods, including MAM, SMat, and MatAny.

When compared to domain-specific models, SAMA outperforms task-specific approaches such as GFM~\cite{li2022bridging} on the AM2K dataset, despite not being trained on animal categories. Similarly, SAMA demonstrates competitive performance on the P3M-500 dataset, rivaling portrait-specific models like PPM and PPM-VITAE~\cite{li2021privacy}.
These results highlight the strong generalization capability of SAMA in zero-shot settings. It consistently delivers robust performance across diverse object categories, especially under the interactive matting paradigm, outperforming prior interactive methods.

\begin{table*}[t]
\centering
\caption{Comparisons on Semantic matting datasets.}
\begin{tabular}{ll c c c c c c c c}
\toprule
\textbf{Dataset} & \textbf{Metric} & GFM & PPM & PPM-ViTAE & SMat & MatAny & MAM & SAMA \\
\midrule

\multirow{2}{*}{\textbf{AM2K}} 
& SAD & 11.11 & 23.06 & 37.84  & 16.84 & 11.9 & 17.30 & \textbf{8.04}\\
& MSE & 0.0031 & 0.0096 & 0.0189  & 0.0047 & 0.0033 & 0.0035  & \textbf{0.0030} \\

\midrule

\multirow{2}{*}{\textbf{P3M-500}} 
& SAD & 111.98 & 13.38 & \textbf{7.80}  & 12.43  & 17.82 & 21.20 & 9.08\\
& MSE & 0.0613 & 0.0042 & \textbf{0.0017} & 0.0036 & 0.0057 & 0.0082 & 0.0028 \\

\bottomrule
\label{semantic}
\end{tabular}
\end{table*}

\subsubsection{Zero-shot Prompt Robustness in Matting}



In Table~\ref{prompt_matting}, we evaluate the performance of SAMA on the RefMatte-RW100~\cite{li2023referring} benchmark. To assess the robustness of SAMA under different types of interactive prompts, including those with added noise, we conduct evaluations using three prompt types: point prompts (10 randomly selected points), bounding boxes, and bounding boxes with added noise.

As shown in Table~\ref{prompt_matting}, SAMA achieves the best performance with box and noisy box prompts and performs second-best with point prompts, closely matching the results of SMat. Although SMat~\cite{ye2024unifying} demonstrates strong performance in point prompts, it unexpectedly performs only third-best in both box and noisy box prompts. This behavior suggests a limitation in SMat’s handling of bounding boxes as prompt inputs. While bounding box is commonly used with other detection models, such as GroundingDINO~\cite{liu2024grounding}, our SAMA's advantage in bounding box prompts might be more useful in practice.

\begin{table}[h]
\centering
\small
\renewcommand{\arraystretch}{1.3}
\setlength{\tabcolsep}{4pt} 
\caption{Prompt robustness Comparison in matting dataset RefMatte-RW100.}
\resizebox{\columnwidth}{!}{%
\begin{tabular}{llccccc} 
\toprule
\textbf{Prompt} & \textbf{Metric} 
& SAM 
& MatAny 
& MAM 
& SMat 
& SAMA (ours) \\
\midrule
\multirow{2}{*}{point} 
& SAD & 122.76 & 63.99 & 614.34  & \textbf{25.60}  & 39.2 \\
& MSE &  67.9  & 0.0340 & 0.3450 & \textbf{0.0120} & 0.0121 \\
\midrule
\multirow{2}{*}{box} 
& SAD & 120.10 & 52.91 & 29.23  & 34.86  & \textbf{25.69} \\
& MSE &  65.9  & 0.0270   & 0.0151 & 0.0172 & \textbf{0.0100} \\
\midrule
\multirow{2}{*}{noisy box} 
& SAD & 168.82 & 85.51 & 32.74  & 34.73  & \textbf{27.57} \\
& MSE &  89.6  & 0.0456  & 0.0139  & 0.0146  & \textbf{0.0111} \\
\bottomrule
\end{tabular}%
}
\label{prompt_matting}
\end{table}

\subsubsection{Zero-shot Instance Image Matting}

Table \ref{tab:imq_synthetic} and \ref{tab:imq_natural} present a comparative evaluation of our proposed SAMA model on the widely-used instance-level image matting benchmark, HIM2K~\cite{sun2022human}. We compare SAMA against established instance matting baselines, including Mask R-CNN ~\cite{he2017mask}, GCA ~\cite{li2020natural}, SIM ~\cite{sun2021semantic}, FBA ~\cite{forte2020f}, MGMatting ~\cite{yu2021mask}, and InstMatt ~\cite{sun2022human}, as well as SAM-based interactive matting models such as SAM and MAM. 

As shown in Table~\ref{tab:imq_natural}, SAMA achieves a substantial performance improvement compared to other SAM-based methods. Notably, even without leveraging external mask guidance like InstMatt, which is specifically trained for instance matting, SAMA delivers the second-best performance and closely approaches InstMatt's results. On the natural subset—crucial due to the dominance of natural scenes in real-world settings—SAMA outperforms all other instance-level models. These findings underscore the effectiveness and robustness of SAMA in practical, unconstrained environments.

\begin{table}[t]
  \centering
  \scriptsize  
  \setlength{\tabcolsep}{2pt}  
  \renewcommand{\arraystretch}{1.15}
  \caption{IMQ scores on \textbf{Synthetic Subset} ($\uparrow$ is better).}
  \label{tab:imq_synthetic}
  \begin{tabular}{l|ccccccccc}
    \toprule
    Metric & MRCNN & GCA & SIM & FBA & MGM & InstM & SAM & MAM & SAMA \\
    \midrule
    IMQ\textsubscript{mad}  & 18.37 & 37.76 & 43.02 & 36.01 & 51.67 & \textbf{63.59} & 49.69 & 56.32 & 57.41 \\
    IMQ\textsubscript{mse}  & 25.65 & 51.56 & 52.90 & 51.44 & 67.08 & \textbf{78.14} & 61.44 & 69.47 & 77.25 \\
    IMQ\textsubscript{grad} &  0.45 & 38.33 & 40.63 & 37.86 & 53.03 & \textbf{64.50} &  4.34 & 31.36 & 47.26 \\
    IMQ\textsubscript{conn} & 19.07 & 39.90 & 44.29 & 38.81 & 55.38 & \textbf{67.71} & 51.84 & 56.82 & 59.29 \\
    \bottomrule
  \end{tabular}
\end{table}

\begin{table}[t]
  \centering
  \scriptsize  
  \setlength{\tabcolsep}{2pt}  
  \renewcommand{\arraystretch}{1.15}
  \caption{IMQ scores on \textbf{Natural Subset} ($\uparrow$ is better).}
  \label{tab:imq_natural}
  \begin{tabular}{l|ccccccccc}
    \toprule
    Metric & MRCNN & GCA & SIM & FBA & MGM & InstM & SAM & MAM & SAMA \\
    \midrule
    IMQ\textsubscript{mad}  & 24.22 & 45.72 & 54.43 & 34.81 & 57.98 & 70.26 & 61.15 & 69.83 & \textbf{71.06} \\
    IMQ\textsubscript{mse}  & 33.74 & 61.40 & 66.67 & 48.32 & 71.12 & 81.34 & 74.01 & 82.52 & \textbf{86.77} \\
    IMQ\textsubscript{grad} &  2.27 & 44.77 & 49.56 & 36.29 & 66.53 & \textbf{74.90} & 13.64 & 52.26 & 69.92 \\
    IMQ\textsubscript{conn} & 26.65 & 48.81 & 58.12 & 37.23 & 60.86 & 72.60 & 65.85 & 73.54 & \textbf{74.32} \\
    \bottomrule
  \end{tabular}
\end{table}

\begin{table*}[t]
\centering
\begin{tabular}{lccccccccccc}
\toprule
\multirow{2}{*}{Model}  & \multicolumn{5}{c}{DIS-ALL} & \multicolumn{2}{c}{P3M-500} & \multicolumn{2}{c}{Model Params} & FPS \\
   \cmidrule(lr){2-6} \cmidrule(lr){7-8} \cmidrule(lr){9-10}
 & $F_{\!\beta}^{\text{max}}\!\uparrow$ &
      $F_{\!\beta}^{w}\!\uparrow$ &
      $M \downarrow$ &
      $S_{\alpha}\!\uparrow$ &
      $E_{\!\phi}^{m}\!\uparrow$ &
      SAD\,$\downarrow$ &
      MSE\,$\downarrow$
      & Total & Learn. &  &  \\
\midrule
HQ-SAM-B       & 0.841 & 0.771 & 0.061 & 0.867 & 0.889 & - & - & 362.1  & 4.1  & 9.8 \\
SAMA-B          & 0.901 & 0.793 & 0.054 & 0.886 & 0.914 & 11.80 & 0.0042 & 389.8 & 31.8 & 8.6 \\
\hline
HQ-SAM-L       & 0.902 & 0.801 & 0.066 & 0.879 & 0.905 & - & - & 1196.1 & 5.1 & 4.8  \\
SAMA-L          & 0.917 & 0.865 & 0.037 & 0.908 & 0.941 & 10.84 & 0.0037  & 1228.3 &  37.3 & 4.2 \\
\hline
HQ-SAM-H       & 0.859 & 0.835 & 0.045 & 0.860 & 0.924 & - & - & 2452.1  & 6.1  & 3.4  \\
SAMA-H          & 0.926 & 0.897 & 0.026 & 0.925 & 0.956 & 9.08 & 0.0028  & 2488.8  & 42.8 & 3.3 \\

\bottomrule
\end{tabular}
\caption{Comparison of SAMA and HQ-SAM across segmentation benchmarks, including model size and FPS. The model parameters are in MB. 'Learn.' means learnable parameters.}
\label{tab:comparison}
\end{table*}

\section{Comparisons with different backbones}

In Table~\ref{tab:comparison}, we compare SAMA with  HQ-SAM on different backbones, including ViT-B, ViT-L, and ViT-H.
We conduct a comprehensive comparison of model performance on the DIS-TE (ALL) dataset (consisting of all 2,000 samples from the DIS test set), evaluating quantitative results, model size, and inference speed. 

As shown in Table~\ref{tab:comparison}, SAMA consistently outperforms HQ-SAM across all metrics and backbone configurations. Although SAMA introduces additional learnable parameters, the proportion relative to the frozen pretrained SAM remains small. Consequently, the frames per second (FPS) metric shows minimal degradation compared to HQ-SAM. Notably, as the backbone size increases, the FPS gap between the models becomes less pronounced. We further evaluate SAMA on the matting task using different backbones. Results indicate that larger backbones yield improved matting performance. Since HQ-SAM is not able to produce matting masks, it is omitted from the corresponding comparison in Table~\ref{tab:comparison}. Importantly, SAMA is capable of generating high-quality matting masks during inference with only a marginal reduction in FPS, demonstrating a favorable trade-off between performance and efficiency.

\section{Implementation Details}
We implement SAMA and conduct all experiments in PyTorch. During the traning, all parameters of the pretrained SAM model are frozen, and only the proposed modules are updated, using two NVIDIA A100~80GB GPUs. SAMA is jointly trained on segmentation and matting datasets for a total of 120K iterations. Optimization is performed with the Adam optimizer, using an initial learning rate of $5 \times 10^{-4}$ and a batch size of~6. The maximum number of training epochs is set to~100. To ensure compatibility with diverse prompt types and maintain SAM’s flexibility in interactive settings, we adopt the prompt sampling strategy from~\cite{ke2023segment}, which incorporates a mixture of bounding boxes, randomly sampled points, and coarse masks. In this process, the SAMA tokens of size $2\times256$ are concatenated with SAM’s mask tokens (size $4\times256$), IoU token (size $1\times256$), and prompt tokens (size $N_{\text{prompt}}\times256$) as the input to the proposed mask decoder. In the prediction heads, the output features are upsampled to $1024\times1024$ to produce high-resolution masks.

\subsection{Efficiency Analysis}

To evaluate SAMA’s computational trade-offs, Table~\ref{efficient} reports an efficiency comparison across SAM-based models. Although SAMA introduces more tunable parameters than lightweight fine-tuning methods, its inference speed (FPS) drops only slightly, and the trainable parameter ratio relative to the full SAM remains small. This minor cost is justified by SAMA’s improved fine-grained segmentation performance and added matting capability. Compared with matting-specific models such as MatAny, which depend on large pretrained ViT backbones, SAMA is more efficient at inference. Relative to MAM, SAMA also uses fewer learnable parameters while achieving higher FPS, demonstrating a favorable balance between accuracy and efficiency.

\begin{table}[H]
\centering
\small
\renewcommand{\arraystretch}{1.2}
\setlength{\tabcolsep}{6pt} 
\resizebox{\columnwidth}{!}{%
\begin{tabular}{l|cccccc}
\toprule
\textbf{Metric} & \textbf{SAM} & \textbf{HQ-SAM} & \textbf{PI-SAM} & \textbf{MAM} & \textbf{MatAny} & \textbf{SAMA} \\
\midrule
LP  & 2446 & 10.5 & 11.7 & 71 & 0   & 44  \\
FPS & 3.5  & 3.4  & 3.4  & 3.2 & 2.6 & 3.3 \\
HR  & No   & No   & Yes  & No  & Yes & No  \\
\bottomrule
\end{tabular}%
}
\caption{Learnable parameters (LP, MB), inference speed (FPS), and human refinement (HR). HR means human refinement, indicating if the model needs humans to
refine the mask during inference.}
\label{efficient}
\end{table}



\section{Limitations}
Although our model delivers strong performance on both segmentation and matting, it still has two notable limitations:

\textbf{Video segmentation.} The recent SAM2 framework~\cite{ravi2024sam} extends segmentation to the video domain, while our method is limited to static images. We will therefore explore integrating temporal cues to enable accurate and efficient video segmentation in future work.

\textbf{Computational efficiency.} Leveraging SAM as the backbone provides robust interactive capabilities, but at the expense of speed and memory. Inference remains comparatively slow and demands high-end GPUs ($\ge$ 10 GB). Reducing both time and memory footprints will be a key focus moving forward.

\textbf{Open-world Vocabulary.} Since SAM3~\cite{carion2025sam} introduces Promptable Concept Segmentation (PCS), which accepts textual prompts and produces corresponding segmentation masks, we will adopt the same mechanism in the future to replace the bounding-box–based prompts from GroundingDINO with direct text-based prompts.



\end{document}